\documentclass[reqno,twoside]{amsart}

\usepackage[utf8]{inputenc}
\usepackage[T1]{fontenc}
\usepackage[english]{babel}
\usepackage{amsmath,amssymb,amsfonts,amsthm}
\usepackage{microtype}
\usepackage{fullpage}
\usepackage{floatrow}
\usepackage{lmodern}
\usepackage{textgreek}
\usepackage{mathtools}
\usepackage{mathrsfs}
\usepackage{graphicx}
\usepackage{xcolor}
\usepackage{microtype}
\usepackage{multirow}
\usepackage{booktabs}
\usepackage{sidecap}

\usepackage{tabularx}
\usepackage{xltabular}
\usepackage{stmaryrd}
\usepackage{makecell}
\usepackage{multirow}
\usepackage{enumerate} 
\usepackage{placeins}
\usepackage[colorlinks=true,linkcolor=red,citecolor=red]{hyperref}

\usepackage{algorithm}
\usepackage{algorithmicx}
\usepackage{algpseudocode}

\usepackage{tikz}
\usetikzlibrary{arrows.meta,backgrounds,calc,fit,positioning,shapes.geometric}

\definecolor{stateblue}{HTML}{E5EDF4}
\definecolor{adjointred}{HTML}{EEE3E7}
\definecolor{simplexgreen}{HTML}{E2ECE8}
\definecolor{gradientyellow}{HTML}{F1EBDD}
\definecolor{diagramink}{HTML}{28343D}
\definecolor{diagramline}{HTML}{65747D}

\tikzset{
	stage/.style={draw=diagramline, text=diagramink, rounded corners=3pt, thick, align=center,
		minimum height=18mm, text width=24mm, inner sep=3pt, font=\scriptsize},
	flow/.style={-{Latex[length=2.7mm,width=1.8mm]}, thick, draw=diagramline, text=diagramink}
}

\def \*#1{\boldsymbol{#1}}
\newcommand{\NN}{\mathbb{N}}
\newcommand{\RR}{\mathbb{R}}
\newcommand{\MC}[1]{\mathcal{#1}}
\newcommand{\MF}[1]{\mathfrak{#1}}
\newcommand{\uad}{\mathcal{U}_{\text{ad}}}
\newcommand{\inter}[1]{\llbracket #1 \rrbracket}
\newcommand{\interinv}[1]{\llbracket #1 \rrbracket^{\leftarrow}}
\newcommand{\Dcal}{\Delta^b}
\newcommand{\norm}[1]{\left\|#1\right\|}
\newcommand{\E}{\mathbb{E}}
\newcommand{\ip}[2]{\left\langle #1,#2\right\rangle}
\newcommand{\Tset}{\left\{0,\ldots,T-1\right\}}

\DeclareMathOperator{\diag}{diag}

\DeclareMathOperator*{\argmin}{arg\,min}

\DisableLigatures{encoding = *, family = * }

\newtheorem{assumptionInternal}{\hspace{-0.4em}}  

\newenvironment{assumption}[1][]{%
	\begin{assumptionInternal}[#1]%
	}{%
	\end{assumptionInternal}%
}

\newtheorem{assumptionInternalB}{\hspace{-0.4em}}

\newenvironment{assumptionB}[1][]{%
	\begin{assumptionInternalB}[#1]%
	}{%
	\end{assumptionInternalB}%
}

\newtheorem{theorem}{Theorem}[section]
\newtheorem{definition}[theorem]{Definition}
\newtheorem{lemma}[theorem]{Lemma}
\newtheorem{proposition}[theorem]{Proposition}
\newtheorem{corollary}[theorem]{Corollary}
\newtheorem{remark}[theorem]{Remark}

\numberwithin{equation}{section}
\numberwithin{theorem}{section}

\usepackage{lastpage}

\title[When to spend privacy]{When to spend privacy: optimal allocation in private learning}

\author{Umberto Biccari\textsuperscript{\,$\ast$}}  
\address{\textsuperscript{$\ast$}\, Chair of Computational Mathematics, DeustoTech, University of Deusto, Avenida de las Universidades 24, 48007 Bilbao, Basque Country, Spain.
\newline \indent \textsuperscript{$\dagger$}\, Division of Mathematical Sciences, School of Physical and Mathematical Sciences, Nanyang Technological University, 21 Nanyang Link 637371, Singapore.
} 
\email{umberto.biccari@deusto.es}
\thanks{This project has received funding from the European Research Council (ERC) under the European Union's Horizon 2030 research and innovation programme (grant agreement NO: 101096251-CoDeFeL). UB was partially supported by the Basque Government through the consolidated research group "CMRG-Computational Mathematics Research Group" under Grant IT1856-26, by the Grant PID2023-146872OB-I00-DyCMaMod of MICIU (Spain), and by the COST Actions CA24122-Multiscale Stochastics, Patterns, and Analysis of Combinatorial Environments and CA24136-Interactions between Control Theory and Machine Learning. YS was supported by the NTU Start-Up Grant.}

\author{Yongcun Song \textsuperscript{\,$\dagger$}}
\email{yongcun.song@ntu.edu.sg}


\begin{document}

\begin{abstract}
Privacy accounting bounds the cumulative privacy cost of Differentially Private Stochastic Gradient Descent but does not determine how to allocate a fixed budget across iterations. We formulate the complete Gaussian noise-variance schedule as an open-loop control under a composed zero-Concentrated Differential Privacy constraint, explicitly accounting for schedule selection. For expected quadratic terminal tracking, we prove that every local minimizer exhausts the certified budget, yielding an exact-budget simplex formulation. We introduce Privacy-Budget Adjoint Mirror Descent (PB-AMD), combining pathwise adjoint gradients with entropic updates that preserve positivity, charge floors, and the prescribed budget. Under regularity and coercivity assumptions, we establish existence and budget stability of optimal schedules, design-to-training utility stability, and deterministic, stochastic, and sample-average stationarity guarantees. A linear–quadratic reduction yields an explicit square-root allocation law. Experiments demonstrate tracking improvements over several allocation rules and predictive improvements over uniform allocation in a learning-rate restart setting, where learning-rate allocation performs similarly. Comparisons at a common stationarity tolerance show lower schedule-optimization times than projected gradient descent on the tested problems. The framework connects certified privacy accounting with optimization through nonlinear training dynamics.
\end{abstract}

\maketitle

\section{Introduction}

Differential Privacy (DP) provides a principled framework for limiting the dependence of a randomized algorithm on any individual data record \cite{dwork2014algorithmic}. In stochastic optimization, Differentially Private-Stochastic Gradient Descent (DP-SGD) implements this principle by combining gradient clipping with calibrated Gaussian perturbations \cite{abadi2016deep}. 
Nevertheless, since training involves repeated accesses to the private dataset, each noisy gradient update contributes to the overall privacy loss of the algorithm. A central question in DP-SGD is therefore how to quantify the cumulative privacy cost of the entire sequence of updates and ensure that it remains within a prescribed privacy budget. 

Several accounting frameworks have been developed for this purpose, including zero-concentrated DP (zCDP) and Rényi DP, which provide convenient composition rules for tracking privacy loss across iterations \cite{bun2016concentrated,gopi2021numerical,mironov2017renyi,wang2019subsampled}. Such accounting determines how much privacy expenditure is admissible over the entire training procedure, but it does not determine how this expenditure should be distributed. This leads to the central question of this paper: 
\begin{quote}
	\emph{given a fixed total zCDP budget, how should the Gaussian noise be allocated across the iterations of DP-SGD so as to optimize terminal utility?}    
\end{quote}

Different allocations with the same composed privacy charge need not have the same effect on the final model, since a perturbation introduced at one iteration is propagated through all subsequent optimization steps. The problem is therefore inherently dynamical: the DP-SGD iterates form the state, the sequence of Gaussian noise variances can be interpreted as controls, the expected terminal loss is the performance criterion, and the composed privacy charge acts as a finite resource.

We develop this viewpoint as a finite-horizon stochastic control problem over the complete noise-variance schedule. For the quadratic terminal tracking objective that constitutes the principal setting of the paper, we show that every locally optimal schedule exhausts the available certified privacy charge. Introducing the per-step zCDP charges as decision variables then transforms the saturated resource constraint into an exact-budget simplex. This structure leads to 
\begin{itemize}
	\item an adjoint characterization of the sensitivity of terminal utility to privacy allocation;
	\item a corresponding Karush-Kuhn-Tucker (KKT) optimality system;
	\item and a feasibility-preserving optimization method, Privacy-Budget Adjoint Mirror Descent (PB-AMD), whose iterates remain on the prescribed privacy-budget simplex.
\end{itemize}

The precise meaning of the terminal tracking objective and the privacy-valid information structure under which the schedule may be designed are clarified next.

\subsection{Problem formulation and scope}

To turn the temporal allocation question above into a well-defined optimization problem, two ingredients must be specified. First, one needs a criterion for comparing admissible privacy schedules, that is, a quantitative measure of how the distribution of noise across iterations affects the final outcome of training. Second, one must specify what information may be used to construct the schedule without introducing an additional, unaccounted privacy cost. These two choices determine, respectively, the objective of the control problem and the information structure under which the resulting schedule is privacy valid.

For the first point, we describe the evolution of the model parameters during training by a sequence $\{\theta_t\}_{t=0}^T$, where $\theta_t\in\RR^d$ denotes the parameter vector at iteration $t$. At the level relevant for the control formulation, the DP-SGD dynamics are written as
\begin{align*}
	\theta_{t+1} = \theta_t-\eta_t g_t(\theta_t) - \eta_t\Dcal_t\sqrt{u_t}\,\xi_t, \quad\xi_t\sim\mathcal N(0,I_d), \quad\text{ for all }t\in\inter{T}_0,
\end{align*}
where $g_t:\RR^d\to\RR^d$ is the gradient map of a fixed design model, $\eta_t>0$ is the step size, and $u_t>0$ determines the variance of the Gaussian perturbation at iteration $t$. Here, we have introduced the notation $\inter{T}_0=\{0,\ldots,T-1\}$, that will be adopted throughout the entire paper. The factor $\Delta_t^b$ is the deterministic scale used to calibrate this perturbation; its precise relation to the privacy mechanism is specified in Section \ref{subsec:DP-SGD}. The complete variance schedule
\begin{align*}
	u=(u_0,\ldots,u_{T-1})
\end{align*}
is the quantity to be optimized and plays the role of the control. Removing the Gaussian perturbations gives the corresponding noiseless reference dynamics
\begin{align*}
	\bar\theta_{t+1} = \bar\theta_t-\eta_t g_t(\bar\theta_t), \quad t\in\inter{T}_0.    
\end{align*}

The principal criterion used in this paper to compare privacy schedules is the expected squared deviation of the noisy terminal parameter vector from this reference,
\begin{align}\label{eq:terminal_tracking_intro}
	J(u) = \mathbb{E}\Bigl[\|\theta_T(u)-\bar\theta_T\|^2\Bigr].    
\end{align}

We refer to this criterion as the \emph{terminal tracking objective}. It measures how strongly the accumulated privacy perturbations displace the final model from the terminal state that would be produced by the same design dynamics in the absence of privacy noise. In this way, the effect of the entire temporal noise allocation is summarized through its impact on the final state. The control formulation developed below also accommodates more general terminal losses, but quadratic tracking is the principal objective considered in this work and the setting in which we establish the budget-saturation property underlying the exact-budget formulation.

The terminal tracking objective \eqref{eq:terminal_tracking_intro} should be interpreted as a surrogate for practical learning utility. In general, minimizing the distance to the noiseless terminal state is not equivalent to minimizing empirical or test loss, nor to maximizing predictive accuracy. Its relevance therefore depends on how well the design dynamics approximate the target training dynamics and on whether parameter-space proximity is informative for the downstream learning task. We quantify in Section \ref{sec:open_loop} how discrepancies between the design and target dynamics affect the terminal objective and the suboptimality of schedules optimized on the design model.

A separate issue concerns the information used to optimize the privacy schedule. Throughout the paper, the complete schedule is determined before private training begins and is subsequently kept fixed. When it is constructed from information independent of the sensitive training dataset, its optimization incurs no additional privacy cost. If private information is used instead to select the design model or the schedule, the corresponding selection procedure must itself be differentially private and its privacy cost included in the overall accounting. Section \ref{sec:DP} formalizes these two cases and establishes the corresponding privacy guarantees.

\subsection{Related work and positioning}

We position the present work with respect to three closely related strands of literature: temporal privacy allocation in private learning, optimal-control formulations of learning dynamics, and numerical methods for constrained stochastic optimization.

\subsubsection{Temporal privacy allocation}

Existing temporal allocation strategies include private line-search adaptation of per-step zCDP budgets \cite{lee2018adaptive} and dynamic schedules minimizing a terminal-utility bound that accounts for noise propagation \cite{hong2022dynamic}. Other methods jointly vary clipping thresholds and noise levels \cite{du2021dynamic} or combine dynamic budget allocation with layerwise gradient importance \cite{li2025importance}. Related approaches address adaptive clipping \cite{andrew2021adaptive,bu2023automatic}, learning-rate adaptation \cite{koskela2020learning}, a priori selection of global DP-SGD hyperparameters \cite{vandijk2024proactive}, and allocation across federated-learning rounds and clients \cite{kianidehkordi2025timeadaptive}. Model-based protectors can also be learned from non-private tasks \cite{hong2021learning}.

Our focus is the joint allocation of a fixed privacy budget to minimize the expected squared deviation of the final model from its noiseless counterpart under nonlinear training dynamics. We optimize this objective directly over the complete variance schedule, rather than through a prescribed schedule family or an analytical utility bound. Schedule design uses either information independent of the sensitive dataset or a private selector whose cost is included in composition.

Table~\ref{tab:related_work} summarizes the relevant distinctions: temporal variation, joint optimization of the complete schedule, treatment of terminal performance, privacy accounting for schedule selection, and use of the training dynamics in optimization.

\begin{table}[th]
    \centering
    \small
    \setlength{\tabcolsep}{4pt}
    \begin{tabularx}{\textwidth}
        {@{}lcccc>{\centering\arraybackslash}X@{}}
        \toprule
        Approach & \makecell{Temporal\\allocation} & \makecell{Joint schedule\\optimization} & \makecell{Terminal\\criterion} & \makecell{Privacy-valid\\selection} & \makecell{Schedule\\construction}
        \\
        \midrule
        \cite{lee2018adaptive} & Yes & No & No & Yes & Adaptive private line search
        \\[10pt]
        \cite{hong2022dynamic}
        & Yes & Yes & Upper bound & \textendash & Analytical minimization of a utility bound 
        \\[10pt]
        \cite{du2021dynamic}
        & Yes & No & No & \textendash & Prescribed clipping and noise schedules
        \\[10pt]
        \cite{vandijk2024proactive} & Indirect & No & Validation & Yes & A priori privacy-utility parameter selection
        \\[10pt]
        Present work & Yes & Yes & \makecell{Expected\\terminal loss} & Yes & Terminal-loss optimization through nonlinear dynamics with exact-budget updates
        \\
        \bottomrule
    \end{tabularx}
    \vspace{0.2cm}
    \caption{Comparison with representative approaches to temporal privacy allocation.}
    \label{tab:related_work}
\end{table}

\subsubsection{Optimal control perspective} 

Residual and continuous-depth neural networks have been interpreted as controlled dynamical systems, leading to Pontryagin-type optimality conditions and structure-preserving numerical methods \cite{benning2019control,e2019meanfield,li2018maximum}; classical discrete-time stochastic control provides the underlying state-control-cost framework \cite{bertsekas1978stochastic}. In private learning, \cite{yuan2026optimal} estimate the communication rounds needed to reach a target loss and use this estimate to adjust Gaussian noise in federated learning, while RLDP learns a feedback policy for adjusting clipping and noise during private fine-tuning \cite{khadangi2025rldp}.

The control perspective provides a characterization of the sensitivities governing this allocation problem. A pathwise adjoint propagates terminal-loss sensitivities backward through the noisy training trajectory, accounting for how subsequent nonlinear updates transform each perturbation. Combined with the privacy constraint, these sensitivities yield first-order optimality conditions that balance the marginal value of privacy expenditure across iterations and supply the gradient used by the allocation algorithm.

\subsubsection{Optimization methodology}

Numerical allocation requires terminal sensitivities for the full schedule and updates that respect positivity and the reciprocal privacy constraint. Quadratic penalties introduce a tunable parameter and do not guarantee feasibility at finite parameter values \cite{cplNocedalWright2006}. Alternatively, projected gradient descent in privacy-charge coordinates can use the same adjoint sensitivities and preserve feasibility through Euclidean projection of additive steps. 

When privacy charges differ substantially across iterations, however, its additive Euclidean steps suffer from scale sensitivity: a fixed absolute charge change corresponds to a larger relative change in noise variance at a smaller charge. To address this scale sensitivity, we use entropic mirror descent to update privacy charges multiplicatively. We prove that every optimizer of quadratic tracking exhausts the certified budget, yielding an exact-budget simplex in reciprocal privacy-charge coordinates. This reduction motivates using entropic mirror descent, a standard simplex method \cite{BeckTeboulle,NemirovskiYudin}, whose proximal term penalizes equal absolute charge changes more strongly at smaller charges and whose multiplicative updates control relative changes in noise variance. 

Implementing entropic mirror descent poses two difficulties: separate forward sensitivity calculations repeat propagation along the noisy nonlinear trajectory, while standard normalization can violate positive charge floors that bound noise variances and clipping would exceed the budget. We compute the full schedule gradient using one forward simulation and one backward adjoint sweep per Gaussian path. 

The standard entropic proximal step on the simplex with charge floors reduces to a scalar normalization enforcing both the floors and the budget equality. The resulting Privacy-Budget Adjoint Mirror Descent (PB-AMD) preserves positive, exact-budget iterates. For its analysis, relative smoothness provides the Bregman counterpart of Euclidean smoothness \cite{bauschke2017descent,cplLuFreundNesterov2018}, while stochastic mirror-descent and sample-average approximation provide frameworks for stochastic and Monte Carlo optimization \cite{fatkhullin2024taming,GhadimiLan,cplKleywegtEtAl2002,cplNemirovskiEtAl2009,cplShapiroEtAl2014}.

\subsection{Main contributions}

We advance temporal privacy allocation through three connected contributions.
\begin{itemize}
    \item[1.] \emph{A control-theoretic foundation for temporal privacy allocation.} We formulate joint optimization of the full Gaussian-variance schedule through nonlinear training dynamics as a finite-horizon stochastic control problem. For privacy validity, we certify private training for data-independent schedules and for schedules returned by a private selector whose cost is included in composition. Under the stated assumptions, we also establish existence and budget stability of optimal schedules. When design and training models differ, we bound the effects on tracking performance and allocation suboptimality.

    \item[2.] \emph{A budget-saturation theorem enabling exact-budget optimization.} We prove that every local minimizer of the expected quadratic terminal tracking objective exhausts the certified privacy budget. Consequently, reciprocal privacy-charge coordinates reduce the search to an exact-budget simplex. The pathwise adjoint supplies the full schedule gradient, from which we derive the KKT system characterizing marginal privacy values across the training horizon.

    \item[3.] \emph{A dynamics-aware algorithm with exact privacy-budget enforcement.} Building on this exact-budget structure, we introduce Privacy-Budget Adjoint Mirror Descent (PB-AMD), whose floor-aware entropic updates use adjoint gradients through the full nonlinear training dynamics and enforce positivity and the prescribed total privacy charge at every iteration, without a penalty parameter. For this method, we establish deterministic and stochastic stationarity guarantees and quantify the transfer from fixed-path sample-average stationarity to the population objective. The linear--quadratic reduction further yields an explicit square-root allocation law and a global convergence result.
\end{itemize}

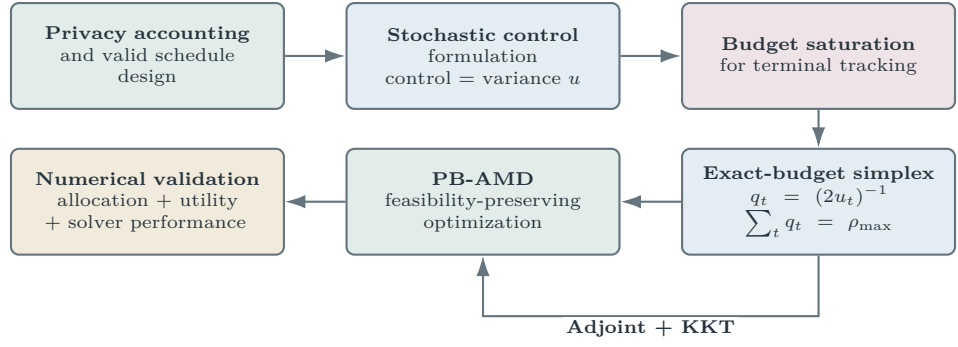
\begin{figure}[ht]
	\centering
	\begin{tikzpicture}[node distance=7mm and 8mm]
		
		\node[stage, 
		style={minimum height=14mm, inner ysep=1mm},
		text width=34mm, align=center,
		fill=simplexgreen] (privacy)
		{\scriptsize\textbf{Privacy accounting} \\ \scriptsize and valid schedule \\ \scriptsize design};
		
		\node[stage,
		style={minimum height=14mm, inner ysep=1mm},
		text width=34mm, align=center,
		fill=stateblue,
		right=of privacy] (control)
		{\scriptsize\textbf{Stochastic control} \\ \scriptsize formulation \\ \scriptsize control $=$ variance $u$};
		
		\node[stage,
		style={minimum height=14mm, inner ysep=1mm},
		text width=34mm, align=center,
		fill=adjointred,
		right=of control] (saturation)
		{\scriptsize\textbf{Budget saturation} \\ \scriptsize for terminal tracking};
		
		\node[stage,
		style={minimum height=14mm, inner ysep=1mm},
		text width=34mm, align=center,
		fill=gradientyellow,
		below=5mm of privacy] (numerics)
		{\scriptsize\textbf{Numerical validation} \\ \scriptsize allocation + utility \\ \scriptsize + solver performance};
		
		\node[stage,
		style={minimum height=14mm, inner ysep=1mm},
		text width=34mm, align=center,
		fill=simplexgreen,
		right=of numerics] (pbamd)
		{\scriptsize\textbf{PB-AMD} \\ \scriptsize feasibility-preserving \\ \scriptsize optimization};
		
		\node[stage,
		style={minimum height=14mm, inner ysep=1mm},
		text width=34mm, align=center,
		fill=stateblue,
		right=of pbamd] (simplex)
		{\scriptsize\textbf{Exact-budget simplex} \\ \scriptsize $q_t=(2u_t)^{-1}$ \\ \scriptsize $\sum_t q_t=\rho_{\max}$};
		
		\draw[flow] (privacy) -- (control);
		\draw[flow] (control) -- (saturation);
		\draw[flow] (saturation) -- (simplex);
		
		\draw[flow] (simplex) -- (pbamd);
		\draw[flow] (pbamd) -- (numerics);
		
		\draw[flow] (simplex.south) -- ++(0,-8mm) coordinate (aux)
		-- node[midway, below, font=\scriptsize\bfseries, fill=white, inner sep=1pt]
		{Adjoint + KKT}
		(pbamd.south |- aux)
		-- (pbamd.south);
		
	\end{tikzpicture}
	\caption{Conceptual roadmap of the paper.}
	\label{fig:roadmap}
\end{figure}

Figure \ref{fig:roadmap} summarizes the logical structure connecting these contributions, from privacy-valid schedule design and the stochastic control formulation to the exact-budget optimization method and its numerical validation.

\subsection{Notation}\label{subsec:notation} 

Table \ref{tab:notation} collects some recurrent notation used in this paper and its meaning. 

\renewcommand{\arraystretch}{1.12}
\begin{table}[th]
	\centering	
	\small
	\begin{tabular}{ll}
	\toprule
	\text{Notation} & \text{Meaning} \\ 
	\midrule
	
	$\NN$ & $\{1,2,\ldots\}$ 
	\\
	$\inter{q}$, $\inter{q}_0$ & $\{1,\ldots,q\}$ and $\{0,\ldots,q-1\}$, respectively, with $q\in\NN$
	\\
	$\interinv{q}$, $\interinv{q}_0$ & indices in $\inter{q}$ and $\inter{q}_0$ taken in reverse order, i.e., $q,\ldots,1$ and $q-1,\ldots,0$
	\\[10pt]
	$\norm{\cdot}_1$, $\norm{\cdot}_2$, $\norm{\cdot}_\infty$ & $\ell^1$, $\ell^2$ and $\ell^\infty$ norms on $\RR^n$, respectively, with $n\in\NN$
	\\
	$\ip{\cdot}{\cdot}$, $\norm{\cdot}_{\rm op}$ & Euclidean inner product and induced operator norm
	\\
	$I_n$ & $n\times n$ identity matrix
	\\
	$\preceq$ and $\succeq$ & Loewner order on symmetric matrices
	\\
	$\operatorname{Tr}$ and $\diag$ & matrix trace and the diagonal-matrix operator
	\\[10pt]
	$\Pr(\cdot)$ and $\E$ & probability and expectation
	\\
	$P\ll Q$ & the probability measure $P$ is absolutely continuous with respect to the probability
    \\
	& measure $Q$, i.e., $Q(A)=0$ implies $P(A)=0$ for every measurable set $A$
	\\
	\bottomrule
	\end{tabular}
	\caption{Summary of some notation employed in the paper.}\label{tab:notation}
\end{table}

\subsection{Organization}
	
Section~\ref{sec:DP} establishes the privacy-accounting framework, calibrates the DP-SGD mechanism, and separates private training from the design model used to select an open-loop schedule.  Section~\ref{sec:open_loop} formulates the finite-horizon control problem and develops its existence, stability, budget-saturation, adjoint, and KKT properties.  Section~\ref{sec:algorithm} derives PB-AMD on the exact-budget simplex and proves its deterministic, stochastic, sample-average, and linear-quadratic convergence results. 	Section~\ref{sec:numerics} evaluates the predicted allocation regimes, terminal utility, exact constraint realization, and solver efficiency. Section~\ref{sec:conclusion} concludes the paper.

\section{Privacy accounting and open-loop schedule certification}\label{sec:DP}

We recall here some known DP notions used in this paper. In particular, we give the formal definition of DP, explain the privacy loss originating from repeated application of a differentially private mechanism and motivating the notion of a privacy budget, and establish privacy guarantees for the DP-SGD mechanism considered below.  

\subsection{Zero-concentrated differential privacy and Gaussian privacy accounting}\label{subsec:DP_preliminary}

We recall the privacy framework used throughout the paper; formal definitions and additional details are collected in Appendix \ref{app:DP}.

We work under replace-one adjacency: two datasets $D$ and $D'$ of the same size $N$ are neighboring, written $D\sim D'$, if they differ in exactly one entry. Privacy is quantified using the notion of \emph{zero-Concentrated Differential Privacy} (zCDP) \cite{bun2016concentrated}. Given $\rho\geq0$, a randomized mechanism $M$ satisfies $\rho$-zCDP if, for every pair $D\sim D'$ and every order $\alpha>1$,
\begin{align*}
	\MF D_\alpha\big(\MC L(\MC M(D))\,\|\,\MC L(\MC M(D'))\big) \leq \rho\alpha.    
\end{align*}

Here $\MC L(\cdot)$ denotes the law of a random variable, and $\MF D_\alpha$ is the R\'enyi divergence of order $\alpha$ (see Definition \ref{def:reny}). This condition controls how much the output distribution can change when one data record is replaced. Smaller values of $\rho$ provide stronger privacy guarantees. We refer to $\rho$ as the \emph{privacy charge}.

For the present paper, the importance of zCDP is that it provides a particularly simple accounting rule for Gaussian perturbations. If two randomized mechanisms $\MC M_1$ and $\MC M_2$ satisfy $\rho_1$- and $\rho_2$-zCDP, then their composition $\MC M_1\circ\MC M_2$ fulfills $(\rho_1+\rho_2)$-zCDP. Moreover, zCDP is preserved under post-processing: if $\MC M$ satisfies $\rho$-zCDP, then any computation using only its output and independent randomness also satisfies $\rho$-zCDP.

These properties provide the privacy-accounting framework for our study of DP-SGD \cite{abadi2016deep}. Gaussian noise protects each gradient query, while composition bounds the privacy charge of the full training procedure. In the next subsection, we identify the certified per-step charge associated with each noise variance and establish the conditions under which the selected schedule yields a valid overall privacy guarantee.

\subsection{A calibrated DP-SGD mechanism}\label{subsec:DP-SGD}

We present here the DP-SGD approach of \cite{abadi2016deep} and discuss its zCDP properties. First, we define the private training dynamics of DP-SGD and identify the zCDP charge associated with each iteration. Second, we establish the privacy guarantees under two schedule-selection mechanisms: (i) schedules fixed independently of the sensitive dataset and (ii) schedules produced by a differentially private preliminary procedure. 

Fix a training horizon $T\geq 1$ and a parameter dimension $d\in\NN$. Denote by $\ell_t(\theta,z)$ the per-example loss at iteration $t\in\inter{T}_0$, and define the clipped gradient at level $C_{\mathrm{clip}}>0$ by
\begin{align*}
	c_t(\theta,z)\coloneqq \beta^{-1}\nabla_\theta\ell_t(\theta,z) \quad\text{ with }\quad \beta\coloneqq\max\left\{1,\frac{\|\nabla_\theta\ell_t(\theta,z)\|_2}{C_{\mathrm{clip}}}\right\},    
\end{align*}
so that $\|c_t(\theta,z)\|_2\leq C_{\mathrm{clip}}$. At each iteration $t\in\inter{T}_0$, we first sample a batch $B_t\subset\inter{N}$ of prescribed cardinality $b_t$, independently
of the data values. We then compute the corresponding clipped batch-average gradient:
\begin{align*}
	g_t^D(\theta) \coloneqq \frac{1}{b_t}\sum_{i\in B_t}c_t(\theta,z_i).
\end{align*}
Moreover, we define the exact uniform conditional sensitivity by
\begin{align}\label{eq:sensitivity_bound}
    \Delta_t \coloneqq \sup_{\substack{B\subset\inter{N}\\|B|=b_t}} \sup_{\theta\in\RR^d} \sup_{D\sim D'} \left\| \frac{1}{b_t}\sum_{i\in B} \bigl(c_t(\theta,z_i)-c_t(\theta,z_i')\bigr)\right\|_2,
\end{align}

Notice that at most one summand changes when one record is replaced, and each clipped gradient has norm at most $C_{\mathrm{clip}}$. Consequently,
\begin{align}\label{eq:sensitivity_clipping_bound}
	0\leq\Delta_t\leq\Dcal_t, \quad\text{ with }\quad \Dcal_t\coloneqq\frac{2C_{\mathrm{clip}}}{b_t}.
\end{align}

The supremum over $\theta$ makes this bound \eqref{eq:sensitivity_clipping_bound} valid uniformly over the random iterates generated during private training.

We use $\Dcal_t$ to calibrate the Gaussian perturbations in DP-SGD. Assume that the initialization $\theta_0$ and all shared public randomness have a joint law independent of the private dataset. Let $\{\xi_t\}_{t=0}^{T-1}$ be independent standard Gaussian vectors. At each step, conditional on the selected schedule, initialization, current batch indices, shared public randomness, and preceding transcript, the innovation $\xi_t$ has law $\mathcal N(0,I_d)$ under every dataset. Given step sizes $\eta_t>0$ and noise multipliers $\sigma_t>0$, the private training mechanism is
\begin{equation}\label{eq:param_update}
	\theta_{t+1} = \theta_t-\eta_t\left(g_t^D(\theta_t)+\Dcal_t\sigma_t\xi_t\right), \qquad \xi_t\sim\mathcal{N}(0,I_d), \quad t\in\inter{T}_0.
\end{equation}

Conditionally on the preceding training transcript and the public batch $B_t$, the current iterate $\theta_t$ is fixed. The query $g_t^D(\theta_t)$ has sensitivity at most $\Delta_t$, and the Gaussian perturbation in \eqref{eq:param_update} has standard deviation $\Delta_t^b\sigma_t$. By \cite[Proposition 1.6]{bun2016concentrated}, this noisy query satisfies
$\rho_t^{\mathrm{ex}}$-zCDP, where 
\begin{align}\label{eq:rho_ex}
	\rho_t^{\mathrm{ex}} \coloneqq \frac{\Delta_t^2}{2(\Dcal_t\sigma_t)^2} = \left(\frac{\Delta_t}{\Dcal_t}\right)^2 \frac{1}{2\sigma_t^2} \leq \frac{1}{2\sigma_t^2}.
\end{align}

The subsequent multiplication by $\eta_t$ and subtraction from the fixed iterate $\theta_t$ are post-processing operations, so the same privacy guarantee holds for the update $\theta_{t+1}$.

Thus, each iteration is certified to satisfy $(2\sigma_t^2)^{-1}$-zCDP. Introducing the noise-variance controls $u_t\coloneqq\sigma_t^2$, we define the certified per-step
privacy charge by
\begin{align}\label{eq:q}
	q_t\coloneqq\frac{1}{2u_t} = \frac{1}{2\sigma_t^2}.
\end{align}

Here, $\sigma_t$ is the noise multiplier relative to the replace-one sensitivity bound $2C_{\mathrm{clip}}/b_t$; the standard deviation of the noise added to the batch-average gradient is therefore $2C_{\mathrm{clip}}\sigma_t/b_t$.

This is the quantity used in the privacy optimization in Section \ref{sec:open_loop}. It equals $\rho_t^{\mathrm{ex}}$ when $\Dcal_t=\Delta_t$ and exceeds $\rho_t^{\mathrm{ex}}$ when $\Dcal_t>\Delta_t$.

\begin{remark}
The quantity $\Delta_t$ is the exact sensitivity, whereas $\Delta_t^b$ is an explicit upper bound supplied by clipping. Privacy certification requires only a valid upper bound, so computing $\Delta_t$ is unnecessary. For fixed $\sigma_t$, if $\Delta_t<\Delta_t^b$, using $\Delta_t^b$ rather than $\Delta_t$ increases the standard deviation of the noise added to the gradient from $\Delta_t\sigma_t$ to $\Delta_t^b\sigma_t$. A smaller valid upper bound on $\Delta_t$ may therefore replace
$\Delta_t^b$ in \eqref{eq:param_update} to reduce the noise.
\end{remark}

We can now certify the privacy of the complete training procedure. We begin with the setting used principally in the control formulation, in which the entire variance schedule is fixed independently of the sensitive dataset.

\begin{theorem}\label{thm:fixed-schedule-privacy}
	
Assume that the initialization $\theta_0$ and shared public randomness have a joint law independent of the dataset, and use the conditional Gaussian sampling rule above. For a deterministic schedule $u=(u_0,\ldots,u_{T-1})\in(0,\infty)^T$, let $\MC M_u$ release either the full iterate transcript $\{\theta_t\}_{t=1}^T$ of \eqref{eq:param_update} or only the terminal iterate $\theta_T$. Then $\MC M_u$ is $\rho(u)$-zCDP, where
\begin{align}\label{eq:rho_u}
    \rho(u) \coloneqq \sum_{t=0}^{T-1}\frac{1}{2u_t}.    
\end{align}
\end{theorem}

\begin{proof}
Fix the initialization and shared public randomness, whose joint law is independent of the dataset, as well as the deterministic schedule. At iteration $t$, condition on the preceding transcript, the public batch indices and the public randomness revealed so far. The current state $\theta_t$ is then fixed. The conditional Gaussian sampling rule keeps $\xi_t$ standard Gaussian under this conditioning. The sensitivity bound \eqref{eq:sensitivity_bound}, being uniform in $\theta$, applies to the query $D\mapsto g_t^D(\theta_t)$. The Gaussian mechanism therefore makes this conditional update $\rho_t^{\mathrm{ex}}$-zCDP, with $\rho_t^{\mathrm{ex}}$ given by \eqref{eq:rho_ex}. Adaptive composition then shows that the complete training transcript is $\rho(u)$-zCDP, since
\begin{align*}
    \sum_{t=0}^{T-1}\rho_t^{\mathrm{ex}} \leq \sum_{t=0}^{T-1}\frac{1}{2u_t} =\rho(u).
\end{align*}

The bound remains valid after averaging over the data-independent initialization and public randomness. The terminal iterate $\theta_T$ is a function of this transcript and hence satisfies the same guarantee by post-processing. In particular, if $\rho(u)\leq\rho_{\max}$, the mechanism is $\rho_{\max}$-zCDP.

Now let $U$ be a random schedule whose joint law with the initialization and shared public randomness is independent of the dataset, with $\rho(U)\leq\bar\rho$ almost surely. Write $\mu$ for its law, which is the same under neighboring datasets. Conditional on $U=u$, the preceding argument gives a $\bar\rho$-zCDP guarantee. We show that this guarantee also holds when $U$ is released together with the training output.

For neighboring datasets $D\sim D'$, let $P_u,Q_u$ denote the conditional laws of the training output given $U=u$. For every $\alpha>1$,
\begin{align*}
    \MF D_\alpha(P_u\|Q_u)\leq\alpha\bar\rho \qquad\text{for }\mu\text{-almost every }u.    
\end{align*}

The joint output laws are $P(du,dy)=\mu(du)P_u(dy)$ and $Q(du,dy)=\mu(du)Q_u(dy)$. Because they have the same schedule marginal $\mu$,
\begin{align*}
    \exp\bigl((\alpha-1)\MF D_\alpha(P\|Q)\bigr) = \int \exp\bigl((\alpha-1)\MF D_\alpha(P_u\|Q_u)\bigr)\,\mu(du) \leq \exp\bigl((\alpha-1)\alpha\bar\rho\bigr).    
\end{align*}

Thus $\MF D_\alpha(P\|Q)\leq\alpha\bar\rho$ for every $\alpha>1$, proving that the joint mechanism is $\bar\rho$-zCDP. The same guarantee holds if $U$ is omitted, by post-processing.
\end{proof}

Theorem \ref{thm:fixed-schedule-privacy} shows that optimizing a schedule does not itself consume privacy budget when the information used to construct that schedule is independent of the sensitive dataset. This is the setting formalized through the fixed design model in Section \ref{subsec:design-bridge}. A different situation arises when the sensitive dataset is used to choose the schedule. Such selection is still admissible, but its privacy cost must be included explicitly.

\begin{theorem}\label{thm:private-selector}
	
Under the same data-independent initialization assumption as in Theorem~\ref{thm:fixed-schedule-privacy}, let $\rho_{\mathrm{sel}}\geq 0$ and $\rho_{\max}>0$. Let $R_{\rm pub}$ collect the initialization and public randomness shared by schedule selection and training. Suppose that a schedule selector $\MC A$ makes the joint release $(R_{\rm pub},\MC A(D))$ $\rho_{\mathrm{sel}}$-zCDP and returns, almost surely, a schedule $u=\MC A(D)\in(0,\infty)^T$ satisfying $\rho(u) \leq\rho_{\max}$ with $\rho(u)$ given by \eqref{eq:rho_u}. Conditional on the selected schedule, run \eqref{eq:param_update} using fresh Gaussian randomness. Then the joint mechanism releasing both the selected schedule and the training output is $(\rho_{\mathrm{sel}}+\rho_{\max})$-zCDP.    
\end{theorem}

\begin{proof}
The first mechanism releases $(R_{\rm pub},u)$ and incurs charge at most $\rho_{\mathrm{sel}}$. For almost every returned pair, the initialization and shared public randomness are fixed, and the conditional Gaussian sampling rule applies. The argument of Theorem~\ref{thm:fixed-schedule-privacy} therefore bounds the conditional training charge by $\rho_{\max}$. Adaptive composition gives charge at most $\rho_{\mathrm{sel}}+\rho_{\max}$ for the joint release of $(R_{\rm pub},u)$ and the training output. Omitting $R_{\rm pub}$ or releasing only the terminal model preserves this bound by post-processing.
\end{proof}

Theorems \ref{thm:fixed-schedule-privacy} and \ref{thm:private-selector} identify the information structures under which the schedule optimization developed later is privacy valid. If the schedule is constructed entirely from information independent of the sensitive dataset, its selection carries no additional privacy charge. If private information is used to select the schedule, the selection procedure must itself be private and its charge composed with the training charge. The next subsection makes this distinction explicit by separating the data-dependent training mechanism $g_t^D$ from the fixed design model $g_t$ used in the control problem.

\subsection{Private training mechanism versus design model}\label{subsec:design-bridge}

In the strategy proposed in this work, the privacy schedule is optimized before private training begins. To determine this schedule, we need a model of the training dynamics that describes how the noise injected at different iterations affects the terminal objective. Using the private queries $g_t^D$ directly for this purpose would make the schedule data-dependent and would therefore require accounting for the privacy cost of its selection. Our main formulation avoids this additional cost by optimizing the schedule on a fixed, data-independent design model. We denote this design model by
\begin{align*}
	\MC D_{\rm des} \coloneqq \left(\bar\theta_0, \{g_t,\eta_t,\Delta_t^b\}_{t=0}^{T-1}, L\right),    
\end{align*}

The tuple $\MC D_{\rm des}$ represents the complete model used to simulate the training dynamics and evaluate candidate privacy schedules before private training. Here, $\bar\theta_0$ is the design initialization, $g_t:\RR^d\to\RR^d$ is the design gradient map, $\eta_t>0$ is the step size, $\Delta_t^b>0$ is the sensitivity bound used to scale the Gaussian perturbation, and $L$ is the terminal loss. The maps $g_t$ may, for example, be constructed from public or synthetic data or obtained from another data-independent approximation of the training dynamics. The concrete construction of the design models used in this work is specified in Section \ref{subsec:experiment_design}, together with the experimental setup.

Once the schedule has been optimized on $\MC D_{\rm des}$, it is fixed and applied to the actual private training mechanism \eqref{eq:param_update}, where the design maps $g_t$ are replaced by the private queries $g_t^D$. If $\MC D_{\rm des}$ is independent of the sensitive dataset, schedule optimization incurs no additional privacy cost and Theorem \ref{thm:fixed-schedule-privacy} applies. If private information is instead used to construct the design model or select the schedule, the selection procedure must itself be differentially private, as described in Theorem \ref{thm:private-selector}.

Besides, since the schedule is designed using $g_t$ but deployed using $g_t^D$, its performance may be affected by discrepancies between the design and private training dynamics. Theorem \ref{thm:design-transfer} quantifies the resulting effect on the terminal objective and schedule suboptimality. In all cases, the condition $\Delta_t^b\geq\Delta_t$ ensures valid calibration of the Gaussian perturbation used during private training.

\section{Open-loop control formulation and optimality structure}\label{sec:open_loop}

This section develops the main mathematical foundations of the proposed privacy-allocation framework by formulating the fixed design problem introduced in Subsection~\ref{subsec:design-bridge} as a finite-horizon stochastic control problem, with the complete noise-variance schedule as the control. We first show that, for the quadratic terminal tracking objective, every locally optimal schedule exhausts the available certified privacy budget. This property allows us to reformulate the problem on an exact-budget simplex using per-step privacy charges as decision variables. We then derive a pathwise adjoint formula for the gradient of the terminal objective and establish the corresponding KKT optimality conditions. Finally, we investigate the existence and stability of optimal schedules, their dependence on the privacy budget, and the effects of discrepancies between the design model and the actual training dynamics. 

To emphasize the main ideas and their implications for privacy allocation, we accompany the results with their interpretation and, where appropriate, a brief explanation of the underlying arguments. Complete proofs are collected in Appendix \ref{app:proofs_open_loop}.

Fix $T\geq 1$ and maps $g_t:\RR^d\to\RR^d$, step sizes $\eta_t>0$, scales $\Dcal_t>0$, and a design initialization $\bar\theta_0\in\RR^d$. Let the noiseless reference trajectory be
\begin{equation*}
	\bar\theta_{t+1}=\bar\theta_t-\eta_tg_t(\bar\theta_t),
	\qquad
	\theta^\star:=\bar\theta_T,
	\quad t\in\inter{T}_0,
\end{equation*}
and, for $t\in\inter{T}_0$, let the controlled stochastic dynamics be
\begin{equation}\label{eq:controlled-dynamics}
	\theta_{t+1}(u) = \theta_t(u)-\eta_tg_t(\theta_t(u)) - \eta_t\Dcal_t\sqrt{u_t}\,\xi_t, \qquad \theta_0(u)=\bar\theta_0.
\end{equation}
For a terminal loss $L:\RR^d\to\RR\cup\{+\infty\}$, set
\begin{equation}\label{eq:J}
	J(u)\coloneqq\E\left[L(\theta_T(u))\right].
\end{equation}
Moreover, let $\rho(u)$ be given by \eqref{eq:rho_u} and, for $\rho_{\max}>0$, define
\begin{align}\label{eq:admissible_set}
	\uad(\rho_{\max})\coloneqq \Big\{u\in(0,\infty)^T:\rho(u)\le\rho_{\max}\Big\}.
\end{align}
The open-loop problem associated with DP noise scheduling is 
\begin{equation}\label{eq:open_loop}
	\min_{u\in\uad(\rho_{\max})}J(u) \quad\text{subject to the dynamics } \eqref{eq:controlled-dynamics}.
\end{equation}

\subsection{Analytical assumptions}\label{subsec:assumptions}

We collect here the assumptions used in the analysis of the control problem. They fall naturally into two groups. Assumptions~\ref{ass:gaussian}-\ref{ass:noncollapse} concern the stochastic dynamics, the terminal loss, and the coercivity properties needed for the basic well-posedness and stability results. Assumptions~\ref{ass:smooth}-\ref{ass:second-order} impose additional differentiability and growth conditions required only for the adjoint, first-order optimality, and algorithmic regularity analysis.

\subsubsection{Basic assumptions for well-posedness and stability}

\begin{assumption}[Independent Gaussian innovations]\label{ass:gaussian}
The random vectors $\xi_0,\ldots,\xi_{T-1}$ are mutually independent and satisfy $\xi_t\sim\MC N(0,I_d)$.
\end{assumption}

\begin{assumption}[Local regularity and linear growth]\label{ass:lipschitz}
For every $t\in\inter{T}_0$, the map $g_t$ is locally Lipschitz and there are $a_t,b_t\ge0$ such that $\norm{g_t(\theta)}\le a_t+b_t\norm{\theta}$ for all $\theta\in\RR^d$.
\end{assumption}

\begin{assumption}[Continuous polynomial growth]\label{ass:poly}
The loss $L$ is continuous and bounded below by a finite constant $L_{\rm lb}\in\RR$. Moreover, there exist $r_L\ge1$ and $K_L>0$ such that $|L(\theta)|\le K_L\big(1+\norm{\theta}^{r_L}\big)$ for all $\theta\in\RR^d$.
\end{assumption}

\begin{assumption}[Quadratic coercive lower bound]\label{ass:coercive-loss}
There exist $c_1>0$ and $c_0\ge0$ such that $L(\theta)\ge c_1\norm{\theta-\theta^\star}^2-c_0$ for all $\theta\in\RR^d$.
\end{assumption}

\begin{assumption}[Quadratic non-collapse of deterministic steps]\label{ass:noncollapse}
Define $\mathsf G_t(\theta):=\theta-\eta_tg_t(\theta)$.  For every $t\in\inter{T}_0$, there exist $\chi_t>0$ and $\omega_t\ge0$ such that
\begin{equation}\label{eq:quadratic-noncollapse}
    \norm{\mathsf G_t(x)-\mathsf G_t(y)}^2 \geq \chi_t^2\norm{x-y}^2-\omega_t, \qquad x,y\in\RR^d.
\end{equation}
\end{assumption}

Assumptions \ref{ass:gaussian}-\ref{ass:poly} provide the basic probabilistic and continuity framework for the controlled dynamics and terminal objective. Assumptions \ref{ass:coercive-loss}-\ref{ass:noncollapse} are the additional coercivity conditions used to prevent minimizing sequences from escaping to unbounded noise-variance schedules and hence to establish existence and stability of optimal controls.

\subsubsection{Additional assumptions for first- and second-order analysis}

The following assumptions are not required for the existence theory. They are imposed when differentiability of the reduced objective and higher-order regularity are needed.

\begin{assumption}[Smoothness for first-order analysis]\label{ass:smooth}
Each $g_t$ is continuously differentiable, and there are $K_g>0$ and $r_g\geq 0$ such that $\norm{\nabla g_t(\theta)}_{\rm op}\leq K_g\big(1+\norm{\theta}^{r_g}\big)$ for all $\theta\in\RR^d$ and $t\in\inter{T}_0$. The loss $L$ is continuously differentiable with
$\norm{\nabla L(\theta)}\le K_\nabla\big(1+\norm{\theta}^{r_\nabla}\big)$ for some $K_\nabla>0$ and $r_\nabla\ge0$.
\end{assumption}

\begin{assumption}[Second-order polynomial regularity]\label{ass:second-order}
Each $g_t$ is twice continuously differentiable and $L$ is twice continuously differentiable. There are constants $K_2>0$ and $r_2\ge0$ such that, for every $\theta\in\RR^d$ and $t\in\inter{T}_0$, 
\begin{align*}
	\norm{D^2g_t(\theta)}_{\rm op}+\norm{\nabla^2L(\theta)}_{\rm op} \leq K_2\big(1+\norm{\theta}^{r_2}\big),	
\end{align*}
where $D^2g_t(\theta):\RR^d\times\RR^d\to\RR^d$ is the second derivative of $g_t$, viewed as a bilinear map, with
\begin{align*}
	\norm{D^2g_t(\theta)}_{\rm op}\coloneqq \sup_{\norm{v}_2=\norm{w}_2=1}\norm{D^2g_t(\theta)[v,w]}_2.	
\end{align*}
\end{assumption}

To help the reader navigate the assumptions used in the different parts of the analysis, Table \ref{tab:assumptionsA} summarizes their roles and the main results in which they enter.

\begin{table}[th]
	\centering	
	\small
	\begin{tabular}{ccc}
		\toprule
		\textbf{Assumptions} & \textbf{Main role} & \textbf{Main results requiring them}
		\\
		\midrule
		\ref{ass:gaussian}-\ref{ass:lipschitz} & \makecell{Basic stochastic dynamics \\ and moment control}  & \multirow{2}{*}{\makecell{Continuity of $J$, basic control formulation \\[5pt] Theorems \ref{thm:budget-saturation}, \ref{thm:existence}; Propositions \ref{prop:u-adjoint}, \ref{prop:J-cont}, \ref{prop:coercivity}}} 
		\\
		\ref{ass:poly} & \makecell{Integrability and continuity \\ of the terminal objective} &
		\\[20pt]
		\ref{ass:coercive-loss}-\ref{ass:noncollapse} & \makecell{Coercivity and compactness \\ of minimizing sequences} & \makecell{Existence and compactness of minimizers \\[5pt] Proposition \ref{prop:coercivity}; Theorems \ref{thm:existence}, \ref{thm:value-continuity}}
		\\[20pt]
		\ref{ass:smooth} & \makecell{First-order \\ sensitivity analysis} & \makecell{Adjoint gradient, KKT conditions, \\ privacy-charge gradient \\[5pt] Propositions \ref{prop:u-adjoint}, \ref{prop:q-gradient}, Theorem \ref{thm:KKT}, Corollary \ref{cor:lambda-positive}}
		\\[20pt]
		\ref{ass:second-order} & \makecell{Second-order regularity \\ for optimization} & \makecell{Smoothness estimates needed \\ for PB-AMD convergence \\ Proposition \ref{prop:primitive-B}}
		\\        
		\bottomrule
	\end{tabular}
	\vspace{0.2cm}
	\caption{Analytical assumptions and the results in which they are used.}
	\label{tab:assumptionsA}    
\end{table}

\begin{remark}
For the smoothly clipped logistic and Softplus design models in Section~\ref{sec:numerics}, bounded data-gradient terms with polynomially growing derivatives and linear $\ell^2$-regularization gradients ensure \ref{ass:lipschitz}, \ref{ass:smooth}, and \ref{ass:second-order}. The Gaussian innovations satisfy \ref{ass:gaussian}, and $L_{\rm tr}(\theta)\coloneqq\norm{\theta-\theta^\star}^2$ satisfies \ref{ass:coercive-loss} with $c_1=1$ and $c_0=0$. For \ref{ass:noncollapse}, if $g_t(\theta)=r_t(\theta)+R_t\theta$, $\sup_\theta\|r_t(\theta)\|\le C_t$, and $I_d-\eta_tR_t$ is nonsingular,
then \eqref{eq:quadratic-noncollapse} holds with $\chi_t^2=\frac12\,s_{\min}(I_d-\eta_tR_t)^2$ and $\omega_t=4\eta_t^2C_t^2$, where $s_{\min}$ denotes the smallest singular value.
\end{remark}

\subsection{Exact certified-budget saturation for terminal tracking}

We begin with a structural property of the control problem that will be essential for the exact-budget reformulation developed in the next subsection. Recall that the admissible set in \eqref{eq:admissible_set} is defined by the inequality $\rho(u)\leq \rho_{\max}$, so a feasible schedule is not required a priori to expend the entire certified privacy budget. Since the certified privacy charge at iteration $t$ is $(2u_t)^{-1}$, decreasing $u_t$ increases privacy expenditure while reducing the variance of the Gaussian perturbation. It is therefore natural to ask whether an optimal schedule can leave part of the available budget unused.

For the quadratic tracking objective, the answer is no. Indeed, once $u_0,\ldots,u_{T-2}$ are fixed, the state $\theta_{T-1}$ depends only on the previous Gaussian innovations and is therefore independent of $\xi_{T-1}$. The last update can be written as
\begin{align*}
	\theta_T-\theta^\star = \theta_{T-1}-\eta_{T-1}g_{T-1}(\theta_{T-1})-\theta^\star -\eta_{T-1}\Dcal_{T-1}\sqrt{u_{T-1}}\,\xi_{T-1}.    
\end{align*}

For the squared tracking loss, independence and the zero mean of the final innovation eliminate the cross term after taking expectation. Hence the dependence of the objective on $u_{T-1}$ is explicit and strictly increasing. If a feasible schedule had slack in the privacy constraint, one could therefore decrease $u_{T-1}$ slightly, remain feasible, and strictly improve the objective. The following theorem formalizes this observation.

\begin{theorem}\label{thm:budget-saturation}
Assume $T\ge1$, Assumptions~\ref{ass:gaussian} and \ref{ass:lipschitz}, and set the loss function to be 
\begin{align}\label{eq:ltrain}
    L(\theta)=\|\theta-\theta^\star\|^2\eqqcolon L_{\rm tr}(\theta).    
\end{align}
If $\eta_{T-1}\Dcal_{T-1}>0$, then, for fixed $(u_0,\ldots,u_{T-2})$, the reduced objective is finite, affine, and strictly increasing in $u_{T-1}$:
\begin{equation}\label{eq:last-step-affine}
    J(u) = \widetilde J(u_0,\ldots,u_{T-2}) + d\eta_{T-1}^2(\Dcal_{T-1})^2u_{T-1},
\end{equation}
where
\begin{align*}
    \widetilde J(u_0,\ldots,u_{T-2}) \coloneqq \E\norm{\theta_{T-1}-\eta_{T-1}g_{T-1}(\theta_{T-1})-\theta^\star}^2 <\infty.
\end{align*}
Consequently, every local minimizer, and hence every global minimizer, of \eqref{eq:open_loop} satisfies $\rho(u^\star)=\rho_{\max}$.
\end{theorem}

Two aspects of the result are worth emphasizing. First, saturation is necessary not only for a global optimizer, but for every local minimizer. Second, proving it does not require understanding how the terminal objective varies with every component of the schedule. The explicit dependence on the final variance $u_{T-1}$ is already sufficient: whenever the constraint has slack, reducing this single variance produces a feasible descent direction.

Theorem \ref{thm:budget-saturation} therefore makes the certified privacy constraint active at every optimizer of the quadratic tracking problem: $\rho(u^\star)=\rho_{\max}$, with $\rho(\cdot)$ as in \eqref{eq:rho_u}. This equality concerns the composition certificate used in the optimization, which need not be the smallest valid privacy parameter of the terminal-release mechanism. The argument does not establish monotonicity in every earlier variance or saturation for arbitrary terminal losses.

This suggests using the per-step certified privacy charges themselves as decision variables. Setting as in \eqref{eq:q} $q_t=(2u_t)^{-1}$ transforms the active resource constraint into the exact simplex condition
\begin{align}\label{eq:q_budget}
	\sum_{t=0}^{T-1}q_t=\rho_{\max}.    
\end{align}

The next subsection develops this reformulation and clarifies its relation to the original variance-based problem. 

\subsection{Privacy-charge reformulation and exact-budget simplex}

Theorem \ref{thm:budget-saturation} shows that, for the quadratic terminal tracking objective $L_{\rm tr}$ in \eqref{eq:ltrain}, every optimal variance schedule exhausts the available certified privacy budget. We now exploit this property to obtain an equivalent formulation of the control problem in which the privacy constraint has a simpler structure. The key idea is to use the privacy expenditure at each iteration, rather than the corresponding noise variance, as the optimization variable.

Recall that the variance $u_t>0$ determines the magnitude of the Gaussian perturbation introduced at iteration $t$, while its certified zCDP charge is given by $q_t=(2u_t)^{-1}$. This relation is invertible, with $u_t=(2q_t)^{-1}$. Consequently, specifying the complete sequence of positive privacy charges $q=(q_0,\ldots,q_{T-1})$ is equivalent to specifying the variance schedule $u=(u_0,\ldots,u_{T-1})$. 

The two parametrizations describe the same controlled dynamics, but they lead to different representations of the privacy constraint. Indeed, in the original variance variables, admissibility requires $\rho(u) \leq \rho_{\max}$. In terms of the privacy charges, this becomes the linear inequality $\sum_{t=0}^{T-1}q_t\leq\rho_{\max}$; the equality \eqref{eq:q_budget} holds at every optimizer of the quadratic tracking problem by Theorem~\ref{thm:budget-saturation}. We may therefore restrict the search for optimal schedules to the exact-budget simplex
\begin{align*}
	Q^\circ \coloneqq\left\{q\in(0,\infty)^T: \sum_{t=0}^{T-1}q_t=\rho_{\max} \right\}.
\end{align*}

This reformulation has an important practical interpretation. Instead of deciding how much noise to inject at each iteration while satisfying a nonlinear constraint, we directly decide how to distribute a fixed amount of certified privacy expenditure across the training horizon. Each coordinate $q_t$ represents the portion of the available budget assigned to iteration $t$, and increasing this allocation reduces the corresponding noise variance.

The change of variables also modifies the expression of the controlled dynamics. Substituting $u_t=(2q_t)^{-1}$ into \eqref{eq:controlled-dynamics}, we obtain
\begin{equation}\label{eq:q-dynamics}
	\theta_{t+1}(q) = \theta_t(q)-\eta_tg_t(\theta_t(q)) - \frac{\eta_t\Dcal_t}{\sqrt{2q_t}}\xi_t, \qquad t\in\inter{T}_0.
\end{equation}

Although $Q^\circ$ provides the natural domain for the reformulated problem, it is not compact. In particular, sequences of admissible charges may approach the boundary $q_t=0$, where the noise amplitude in \eqref{eq:q-dynamics} becomes unbounded. This creates difficulties for both the analysis and the numerical solution of the optimization problem. To avoid these difficulties, we introduce a small positive lower bound on the individual privacy charges. Given $0<q_{\min}<\rho_{\max}/T$, we define the compact simplex
\begin{align}\label{eq:Qqmin}
	Q(q_{\min})\coloneqq \left\{ q\in\RR^T: q_t\geq q_{\min},\quad \sum_{t=0}^{T-1}q_t=\rho_{\max} \right\}.
\end{align}

The condition on $q_{\min}$ guarantees that this set is nonempty. Moreover, every admissible charge is bounded away from zero, so the noise coefficients in \eqref{eq:q-dynamics} remain finite and uniformly bounded over the feasible set. Writing $Q\coloneqq Q(q_{\min})$, we arrive at the compact optimization problem
\begin{equation}\label{eq:q-problem}
	\min_{q\in Q}F(q), \qquad F(q)\coloneqq\E\big[L(\theta_T(q))\big].
\end{equation}

The introduction of the floor $q_{\min}$ is primarily a technical and computational device. It is therefore important to establish that this restriction does not alter the solutions of the original problem when the floor is chosen sufficiently small. The following proposition makes this equivalence precise.

\begin{proposition}\label{prop:q-equivalence}
The change of variables $q_t=(2u_t)^{-1}$ bijectively maps $\uad(\rho_{\max})$ onto
\begin{align*}
	Q_{\le}\coloneqq\left\{q\in(0,\infty)^T: \sum_{t=0}^{T-1}q_t\le\rho_{\max}\right\},
\end{align*}
preserving the dynamics and objective. Assume $L=L_{\mathrm{tr}}$ as in \eqref{eq:ltrain} and that the minimizer set of \eqref{eq:open_loop} is nonempty and compact. Then there exists $q_\ast\in(0,\rho_{\max}/T)$ such that, for every $0<q_{\min}<q_\ast$,
\begin{align*}
    \argmin_{q\in Q(q_{\min})}F(q) = \argmin_{q\in Q^\circ}F(q) = \left\{\bigl((2u_t)^{-1}\bigr)_{t=0}^{T-1}: u\in\argmin_{\uad}(\rho_{\max})J\right\}.
\end{align*}
\end{proposition}

\subsection{Differentiability, adjoints, and KKT conditions}
To characterize optimal privacy allocations, we first study the sensitivity of the expected terminal loss with respect to the noise-variance schedule. To this end, for $t\in\inter{T}_0$, we introduce the one-step state-transition Jacobian
\begin{align}\label{eq:Jacobian}
	A_t(\theta):=I_d-\eta_t\nabla g_t(\theta).
\end{align}

The following proposition establishes the differentiability of the objective and provides an adjoint representation of its gradient, which will subsequently be used to derive the KKT optimality conditions. 

\begin{proposition}\label{prop:u-adjoint}
Under Assumptions~\ref{ass:gaussian}, \ref{ass:lipschitz}, \ref{ass:poly}, and \ref{ass:smooth}, the map $J$ is continuously differentiable on $(0,\infty)^T$. For each realization $\xi=(\xi_0,\ldots,\xi_{T-1})$ of the Gaussian innovations, let $\theta_t=\theta_t(u,\xi)$ denote the corresponding state trajectory and define the pathwise adjoint by the backward recursion
\begin{equation}\label{eq:adjoint}
    p_T=\nabla L(\theta_T), \qquad p_t=A_t(\theta_t)^\top p_{t+1}, \quad t\in\interinv{T}_0,
\end{equation}
with $A_t(\cdot)$ given by \eqref{eq:Jacobian}. Then the pathwise objective $j(u;\xi):=L(\theta_T(u;\xi))$ satisfies
\begin{equation}\label{eq:path-u-gradient}
    \partial_{u_t}j(u;\xi) =-\frac{\eta_t\Dcal_t}{2\sqrt{u_t}} \ip{p_{t+1}}{\xi_t}, \qquad t\in\inter{T}_0, 
\end{equation}
and differentiation commutes with expectation:
\begin{equation}\label{eq:u-gradient}
    \partial_{u_t}J(u) = \E[\partial_{u_t}j(u;\xi)] = -\frac{\eta_t\Dcal_t}{2\sqrt{u_t}} \E\left[\ip{p_{t+1}}{\xi_t}\right], \qquad t\in\inter{T}_0.
\end{equation}
\end{proposition}

The adjoint representation established in Proposition \ref{prop:u-adjoint} allows us to characterize the first-order optimality conditions for the privacy-allocation problem \eqref{eq:open_loop}. Although the state dynamics are stochastic, the control is an open-loop schedule chosen before training and is therefore deterministic. Moreover, the objective $J(u)$ is an expectation, so the optimization takes place over the finite-dimensional vector $u$, subject to the deterministic privacy-budget constraint. To formulate the corresponding optimality conditions, define
\begin{align}\label{eq:Phi}
	\Phi(u)\coloneqq \rho(u)-\rho_{\max}, \qquad \mathscr L(u,\lambda):=J(u)+\lambda\Phi(u), \quad \lambda\geq 0.
\end{align}

Here, $\mathscr L$ is the reduced Lagrangian, obtained after incorporating the stochastic state dynamics into the objective $J$. Consequently, the only optimization variables are the deterministic schedule $u$ and the scalar multiplier $\lambda$. The state and adjoint trajectories appearing in the gradient formula remain random, and their contribution to the stationarity conditions is therefore expressed through an expectation.

The following theorem combines this gradient representation with the privacy-budget constraint to obtain the KKT conditions for locally optimal schedules. In particular, the multiplier $\lambda$ represents the marginal value of the available privacy budget, while the stationarity conditions characterize how the sensitivity of the terminal objective is balanced across iterations.

\begin{theorem}\label{thm:KKT}
Assume the hypotheses of Proposition \ref{prop:u-adjoint}. Let $u^\star\in \uad(\rho_{\max})$ be a local minimizer of \eqref{eq:open_loop}, and let $\{\theta_t(u^\star)\}_{t=0}^T$ and $\{p_t^\star\}_{t=0}^T$ be its state and pathwise adjoint, defined by \eqref{eq:controlled-dynamics} and \eqref{eq:adjoint}, respectively. Then there exists a deterministic multiplier $\lambda^\star\geq0$ such that
$\lambda^\star\big(\rho(u^\star)-\rho_{\max}\big)=0$ and
\begin{align}\label{eq:OS_stationarity}
    \eta_t\Dcal_t(u_t^\star)^{3/2} \E\left[\ip{p_{t+1}^\star}{\xi_t}\right] + \lambda^\star=0, \quad\text{ for all }t\in\inter{T}_0.
\end{align}
\end{theorem}

Combining the budget-saturation property of Theorem \ref{thm:budget-saturation} with the KKT conditions of Theorem \ref{thm:KKT}, we obtain the following corollary, which establishes the strict positivity of the privacy multiplier and provides a more explicit characterization of optimal schedules for the quadratic terminal tracking objective.

\begin{corollary}\label{cor:lambda-positive}
For $L=L_{\rm tr}$, every local minimizer satisfying the hypotheses of Theorem~\ref{thm:KKT} uses the full budget and has $\lambda^\star>0$.  In fact,
\begin{align*}
    \partial_{u_{T-1}}J(u) = d\eta_{T-1}^2(\Dcal_{T-1})^2 \qquad\text{ and }\qquad \lambda^\star = 2d\eta_{T-1}^2(\Dcal_{T-1})^2(u_{T-1}^\star)^2.
\end{align*}
Consequently, $\E\left[\ip{p_{t+1}^\star}{\xi_t}\right]<0$ for every $t\in\inter{T}_0$, and \eqref{eq:OS_stationarity} may be written
\begin{align}\label{eq:ustar_corollary}
    u_t^\star = \left(-\frac{\lambda^\star}{\eta_t\Dcal_t\E[\ip{p_{t+1}^\star}{\xi_t}]}\right)^{2/3}.
\end{align}
\end{corollary}

The preceding results characterize optimal privacy allocation in terms of the noise variances. We now return to the privacy-charge formulation introduced in Section~3.3 and derive the corresponding adjoint gradient formula, which will be used in the exact-budget optimization method developed in Section~\ref{sec:algorithm}.

For each realization $\xi=(\xi_0,\ldots,\xi_{T-1})$ of the Gaussian innovations, let $\theta_t=\theta_t(q;\xi)$ denote the corresponding state trajectory and define the pathwise terminal objective by $f(q;\xi):=L(\theta_T(q;\xi))$. The following proposition expresses the gradient with respect to the privacy charges in terms of this adjoint.

\begin{proposition}\label{prop:q-gradient}
	Under the assumptions of Proposition~\ref{prop:u-adjoint}, the functional $F$ defined in \eqref{eq:q-problem} is continuously differentiable on an open neighborhood of $Q$, and
	\begin{align}
		\partial_{q_t}f(q;\xi) & =\eta_t\Dcal_t(2q_t)^{-3/2}\ip{p_{t+1}}{\xi_t}, &&t\in\inter{T}_0,\label{eq:path-q-gradient}
		\\
		\partial_{q_t}F(q) &=\E\left[\eta_t\Dcal_t(2q_t)^{-3/2}\ip{p_{t+1}}{\xi_t}\right], &&t\in\inter{T}_0. \notag
	\end{align}
	where $p_t$ is the adjoint state defined in \eqref{eq:adjoint}.
\end{proposition}

The adjoint representation in Proposition \ref{prop:q-gradient} provides a practical way to evaluate the gradient of the expected terminal objective. Since the expectation cannot generally be computed exactly, we approximate it by averaging the pathwise gradients over independent realizations of the Gaussian innovations. Given $M\in\NN$ independent Gaussian paths, define the Monte Carlo gradient estimator
\begin{align}\label{eq:mc-gradient}
	h_t(q) \coloneqq \frac{1}{M}\sum_{m=1}^M \eta_t\Dcal_t(2q_t)^{-3/2} \langle p_{t+1}^{(m)},\xi_t^{(m)}\rangle, \qquad t\in\inter{T}_0.
\end{align}
Here, $p^{(m)}$ denotes the adjoint associated with the $m$-th simulated trajectory, and
\begin{align*}
	h(q)\coloneqq(h_0(q),\ldots,h_{T-1}(q))\in\RR^T.	
\end{align*}
 
Two sampling strategies are possible, with different interpretations. If new Gaussian paths are generated independently of the current iterate $q$, the estimator is conditionally unbiased: $\E[h(q)| q]=\nabla F(q)$. Alternatively, the same Gaussian paths can be fixed and reused throughout the optimization. In this case, $h(q)$ is the exact gradient of the corresponding sample-average objective $F_M(q)$, introduced in Subsection~\ref{subsec:saa}, rather than an unbiased fresh-sample estimator of $\nabla F(q)$.

These two approaches provide the gradient information required for numerical privacy allocation: the first supports stochastic gradient-based optimization of the expected objective $F$, while the second leads to deterministic optimization of its sample-average approximation $F_M$. Both will be considered in the numerical framework developed in Section~\ref{sec:algorithm}.

\subsection{Well-posedness and stability of the control problem}

Having established the first-order optimality conditions and adjoint gradient formulas, we now investigate the well-posedness of the privacy-allocation problem, the dependence of optimal schedules on the privacy budget, and their stability with respect to perturbations of the design model.

Concerning well-posedness, we prove existence of a minimizer by showing that $\uad(\rho)$ is closed, $J$ is lower semicontinuous, and coercivity bounds every minimizing sequence. These properties are established in Propositions \ref{prop:uad}, \ref{prop:J-cont} and \ref{prop:coercivity}, and yield the following result.

\begin{theorem}\label{thm:existence}
	Under Assumptions~\ref{ass:gaussian}, \ref{ass:lipschitz}, \ref{ass:poly}, \ref{ass:coercive-loss}, and \ref{ass:noncollapse}, problem \eqref{eq:open_loop} admits at least one minimizer. Its minimizer set is compact.
\end{theorem}

Dependence on the privacy budget is obtained by proving continuity of the optimal value and outer stability of optimal schedules. For $\rho>0$, write
\begin{align*}
	V(\rho):=\min_{u\in\uad(\rho)}J(u) \quad\text{ and }\quad S(\rho):=\argmin_{u\in\uad(\rho)}J(u).	
\end{align*}
We then have the following result.

\begin{theorem}\label{thm:value-continuity}
	Under the assumptions of Theorem~\ref{thm:existence}, the following hold.
	\begin{enumerate}[(i)]
		\item $V:(0,\infty)\to\RR$ is nonincreasing and continuous.
		\item If $\rho_n\to\rho>0$ and $u_n\in S(\rho_n)$, then $\{u_n\}$ is bounded and every cluster point belongs to $S(\rho)$; equivalently,
		\[
		\limsup_{n\to\infty}S(\rho_n)\subset S(\rho).
		\]
		\item If $S(\rho)=\{u^\star\}$ is a singleton, then every sequence $u_n\in S(\rho_n)$ converges to $u^\star$.
	\end{enumerate}
\end{theorem}

To conclude, we compare the public design dynamics and target training dynamics under the same control and Gaussian innovations. For $a\in\{\mathrm{des},\mathrm{tar}\}$, let $g_t^a$, initial state $\theta_0^a\in\RR^d$, and reference $\theta^{\ast,a}\in\RR^d$ specify model $a$. Both models use the same $\eta_t$, $\Dcal_t$, control, and Gaussian innovations:
\begin{align*}
	\theta_{t+1}^a=\theta_t^a-\eta_tg_t^a(\theta_t^a) - \eta_t\Dcal_t\sqrt{u_t}\xi_t,\qquad J^a(u)=\E\norm{\theta_T^a-\theta^{\ast,a}}^2.
\end{align*}

\begin{theorem}\label{thm:design-transfer}
Assume \ref{ass:gaussian}. Let $\MC K\subset(0,\infty)^T$ be nonempty and compact, and suppose all maps $g_t^a$, $a\in\{\mathrm{des},\mathrm{tar}\}$, are globally Lipschitz, with
\begin{equation}\label{eq:design-map-discrepancy}
	\varepsilon_t^g\coloneqq\sup_{\theta\in\RR^d} \|g_t^{\rm tar}(\theta)-g_t^{\rm des}(\theta)\|<\infty, \quad \text{ for all } t\in\inter{T}_0.
\end{equation}
There is a finite constant $C_{\MC K}$, depending only on the two models and $\MC K$, such that, with 
\begin{align*}
    \varepsilon_{\MC K} \coloneqq C_{\MC K}\left(\|\theta_0^{\mathrm{tar}}-\theta_0^{\mathrm{des}}\| + \sum_{t=0}^{T-1}\eta_t\varepsilon_t^g + \|\theta^{*,\mathrm{tar}}-\theta^{*,\mathrm{des}}\|\right),
\end{align*}
we have
\begin{align*}
    \sup_{u\in \MC K}|J^{\mathrm{tar}}(u)-J^{\mathrm{des}}(u)| \leq\varepsilon_{\MC K} \qquad\text{ and }\qquad \left|\min_{\MC K} J^{\mathrm{tar}}-\min_{\MC K} J^{\mathrm{des}}\right|
\leq\varepsilon_{\MC K}.
\end{align*}
Moreover, every $\gamma$-minimizer $\widehat u\in \MC K$ of $J^{\mathrm{des}}$, meaning $J^{\mathrm{des}}(\widehat u)\leq\min_{\MC K}J^{\mathrm{des}}+\gamma$, satisfies
\begin{equation*}
	J^{\rm tar}(\widehat u) \le\min_{u\in\MC K}J^{\rm tar}(u)+\gamma+2\varepsilon_{\MC K}.
\end{equation*}
\end{theorem} 

Theorem \ref{thm:design-transfer} applies, in particular, to the exact-budget privacy allocations considered in this paper. Indeed, the constraints
$\sum_{t=0}^{T-1}q_t=\rho_{\max}$ and $q_t\geq q_{\min}>0$
ensure that the corresponding noise variances $u_t=(2q_t)^{-1}$ belong to a compact set $\mathcal K$. The theorem therefore provides a uniform stability estimate over all such admissible schedules.

The result can also be applied to the actual training dynamics. Conditional on a fixed dataset and the public batch sequence, we may take $g_t^{\rm tar}=g_t^D$, provided that the required global Lipschitz assumption and the uniform discrepancy bound \eqref{eq:design-map-discrepancy} are satisfied. This allows us to quantify how discrepancies between the fixed design model and the actual training dynamics affect the terminal outcome. Finally, if estimating the model discrepancy or selecting and releasing the schedule involves accessing private data, the corresponding privacy expenditure must be accounted for separately in the private selection stage.

\section{Privacy-budget adjoint mirror descent}\label{sec:algorithm}

We now develop an optimization method for the exact-budget problem \eqref{eq:q-problem}. We use the adjoint gradient in Proposition~\ref{prop:q-gradient} to determine how reallocating the per-step privacy charges changes the terminal objective to first order. We construct Privacy-Budget Adjoint Mirror Descent (PB-AMD) using multiplicative entropic mirror updates and scalar normalization to preserve positivity and enforce the prescribed budget exactly at every iteration. We then establish its deterministic, stochastic, and sample-average convergence guarantees, with complete proofs in Appendix~\ref{app:proofs_algo}.

\subsection{PB-AMD: construction and algorithm}\label{subsec:pbamd-update}

Throughout this section, we construct and analyze PB-AMD on the compact exact-budget simplex introduced in \eqref{eq:Qqmin}, namely,
\begin{align*}
    Q(q_{\min})=\left\{q\in\RR^T:q_t\ge q_{\min},\ \sum_{t=0}^{T-1}q_t=\rho_{\max}\right\}.    
\end{align*}

The floor $q_{\min}>0$ keeps the computation away from the singular boundary $q_t=0$, which would correspond to infinite noise variance. As shown in Proposition~\ref{prop:q-equivalence}, every sufficiently small floor leaves the global minimizer set of the unfloored exact-budget problem unchanged. 

To compare the benefit of assigning additional privacy charge to different training iterations, we define $m_t(q)\coloneqq -\partial_{q_t}F(q)$ as the marginal utility of certified privacy expenditure at iteration $t$. A large $m_t(q)$ means that assigning more charge to that iteration, and hence reducing its noise, gives a large first-order decrease in terminal loss. The KKT conditions then express the required balance of these marginal utilities under the budget and floor constraints: there are $\lambda\in\RR$ and $\mu_t\ge 0$ such that
\begin{align*}
	\partial_{q_t}F(q^\star)+\lambda-\mu_t=0, \qquad \mu_t(q_t^\star-q_{\min})=0, \qquad t\in\inter{T}_0.
\end{align*}
Consequently,
\begin{align*}
	m_t(q^\star)=\lambda\quad\text{if }q_t^\star>q_{\min}\quad\text{ and }\quad m_t(q^\star)\le\lambda\quad\text{if }q_t^\star=q_{\min}.  
\end{align*}

Thus, if $q_s,q_t>q_{\min}$ and $m_t(q)>m_s(q)$, a sufficiently small transfer $q_s\mapsto q_s-\delta$, $q_t\mapsto q_t+\delta$ with $\delta>0$ preserves feasibility in $Q$ and decreases the objective $F$ in \eqref{eq:q-problem} to first order. To implement this budget redistribution through multiplicative mirror updates on $Q$ in \eqref{eq:Qqmin}, we use the negative entropy 
\begin{align*}
	\psi(q):=\sum_{t=0}^{T-1}q_t\log q_t,
\end{align*}
whose Bregman divergence is
\begin{equation*}
	D_\psi(z,q) \coloneqq\sum_{t=0}^{T-1} \left[z_t\log\!\left(\frac{z_t}{q_t}\right)-z_t+q_t\right].
\end{equation*}

Because $q_t,z_t\ge q_{\min}>0$, this divergence is finite and smooth on all of $Q\times Q$.  At an iterate $q^k\in Q$, given a gradient estimate $h^k\in\RR^T$ and a step size $\alpha_k>0$, PB-AMD performs 
\begin{equation}\label{eq:md-implicit}
	q^{k+1} \coloneqq \argmin_{q\in Q} \left\{\ip{h^k}{q}+\frac{1}{\alpha_k}D_\psi(q,q^k)\right\}.
\end{equation}

The next proposition shows that the mirror-descent step admits an explicit multiplicative representation, with a scalar normalization that enforces the exact privacy budget while respecting the lower bounds on individual charges.

\begin{proposition}\label{prop:floor-update}
The minimizer in \eqref{eq:md-implicit} is unique and has the form
\begin{equation}\label{eq:md-update}
	q_t^{k+1} = \max\left\{q_{\min},\ \tau_kq_t^k e^{-\alpha_k h_t^k}\right\}, \qquad t\in\inter{T}_0
\end{equation}
where $\tau_k>0$ is the unique scalar satisfying
\begin{equation}\label{eq:tau-normalization}
	\sum_{t=0}^{T-1} \max\left\{q_{\min},\ \tau_kq_t^k e^{-\alpha_k h_t^k}\right\} = \rho_{\max}.
\end{equation}
In particular, every iterate belongs to $Q$, and a coordinate at the lower bound can return to the interior at a later iteration.
\end{proposition}

In the update \eqref{eq:md-update} given by Proposition~\ref{prop:floor-update}, coordinates with smaller $h_t^k$, equivalently larger estimated marginal privacy utilities, receive larger multiplicative factors $e^{-\alpha_k h_t^k}$ than coordinates with larger $h_t^k$. The scalar $\tau_k$ then enforces 
\begin{align*}
    \sum_{t=0}^{T-1}q_t^{k+1}=\rho_{\max}    
\end{align*}
through \eqref{eq:tau-normalization}, and the floor $q_{\min}$ excludes the singular limit of zero charge (infinite noise variance).

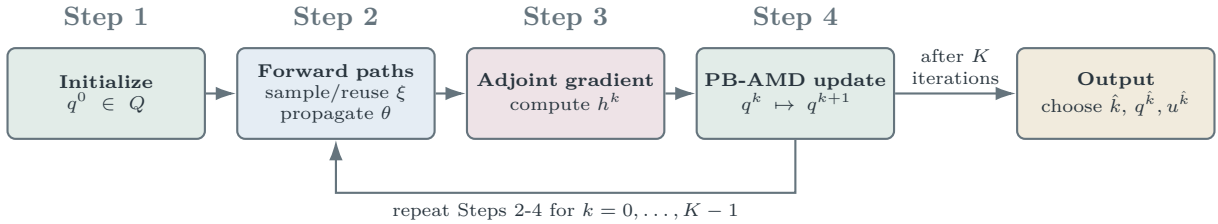
\begin{figure}[h]
	\centering
	\begin{tikzpicture}[node distance=4mm and 4mm]
		
		\node[stage, style={minimum height=12mm, inner ysep=1mm}, fill=simplexgreen] (iterate)
		{\scriptsize\textbf{Initialize}\\ $q^0\in Q$};
		\node[stage, style={minimum height=12mm, inner ysep=1mm}, fill=stateblue, right=of iterate] (forward)
		{\scriptsize\textbf{Forward paths}\\ sample/reuse $\xi$\\propagate $\theta$};
		\node[stage, style={minimum height=12mm, inner ysep=1mm}, fill=adjointred, right=of forward] (adjoint)
		{\scriptsize\textbf{Adjoint gradient}\\compute $h^k$};
		\node[stage, style={minimum height=12mm, inner ysep=1mm}, fill=simplexgreen, right=of adjoint] (update)
		{\scriptsize\textbf{PB-AMD update}\\ $q^k\mapsto q^{k+1}$};
		\node[stage, style={minimum height=12mm, inner ysep=1mm}, fill=gradientyellow, right=16mm of update] (output)
		{\scriptsize\textbf{Output}\\ choose $\hat{k}$, $q^{\hat{k}},u^{\hat{k}}$};
		
		\foreach \boxname/\steplabel in {iterate/Step 1,forward/Step 2,adjoint/Step 3,update/Step 4} \node[above=1.0mm of \boxname, font=\bfseries, text=diagramline] {\steplabel};
		
		\draw[flow] (iterate) -- (forward);
		\draw[flow] (forward) -- (adjoint);
		\draw[flow] (adjoint) -- (update);
		\draw[flow] (update) -- node[above, font=\scriptsize, align=center] {after $K$\\iterations} (output);
		
		\draw[flow] (update.south) -- ++(0,-7mm) coordinate (aux) -- node[midway, below, font=\scriptsize, align=center] {repeat Steps 2-4 for $k=0,\ldots,K-1$} (forward.south |- aux)    -- (forward.south);
	\end{tikzpicture}
\caption{Workflow of PB-AMD.}\label{fig:pbamd-workflow}
\end{figure}

We combine this budget update with the adjoint gradient estimator \eqref{eq:mc-gradient} to obtain the complete PB-AMD procedure in Algorithm~\ref{alg:pbamd}, which supports both fresh-path and fixed-path sample-average approximation (SAA) modes. Figure~\ref{fig:pbamd-workflow} summarizes this procedure.

\begin{algorithm}[h]
	\caption{Privacy-Budget Adjoint Mirror Descent (PB-AMD)}\label{alg:pbamd}
	\small
	\begin{algorithmic}[1]
		\Statex \textbf{Input:} Horizon $T$, iteration count $K\in\NN$, budget $\rho_{\max}$, floor $q_{\min}$ with $Tq_{\min}<\rho_{\max}$, sampling mode $\mathsf{mode}\in\{\mathsf{fresh},\mathsf{SAA}\}$, path counts $M$ or $M_k$, step sizes $\alpha_k>0$, and $q^0\in Q$.
		\If{$\mathsf{mode}=\mathsf{SAA}$}
		\State Draw and store $M$ Gaussian paths and set $M_k=M$ for all $k$.
		\EndIf
		\For{$k=0,\ldots,K-1$}
		\If{$\mathsf{mode}=\mathsf{fresh}$}
		\State Draw $M_k$ paths with $\xi_t^{(m)}\sim\MC N(0,I_d)$ independently over $t$ and $m$.
		\EndIf
		\For{$m=1,\ldots,M_k$}
		\State Set $\theta_0^{(m)}=\bar\theta_0$ and, for $t\in\inter{T}_0$, use \eqref{eq:q-dynamics} to compute $\displaystyle \theta_{t+1}^{(m)}=\theta_t^{(m)}-\eta_tg_t(\theta_t^{(m)}) - \frac{\eta_t\Dcal_t}{\sqrt{2q_t^k}}\xi_t^{(m)}$. 
		\State Using \eqref{eq:adjoint}, set $p_T^{(m)}=\nabla L(\theta_T^{(m)})$ and, for $t\in\interinv{T}_0$, compute $\displaystyle p_t^{(m)} = \left(I_d-\eta_t\nabla g_t(\theta_t^{(m)})\right)^\top p_{t+1}^{(m)}.$
		\EndFor
		\State {For each $t\in\inter{T}_0$, evaluate the gradient estimator in \eqref{eq:mc-gradient} as $\displaystyle h_t^k=\frac{1}{M_k}\sum_{m=1}^{M_k} \eta_t\Dcal_t(2q_t^k)^{-3/2} \ip{p_{t+1}^{(m)}}{\xi_t^{(m)}}.$}
		\State Find the unique $\tau_k>0$ solving \eqref{eq:tau-normalization}, for example by bisection in the log domain.
		\State For each $t\in\inter{T}_0$, apply \eqref{eq:md-update} to obtain $\displaystyle q_t^{k+1}=\max\left\{q_{\min},\tau_kq_t^ke^{-\alpha_kh_t^k}\right\}$.
		\EndFor
		\Statex \textbf{Output:} Choose $\widehat k\in\{0,\ldots,K\}$ by a prespecified rule that uses no private training data, and return $q^{\widehat k}\in Q$ and $u_t^{\widehat k}=(2q_t^{\widehat k})^{-1}$ for $t\in\inter{T}_0$, using \eqref{eq:q}.
	\end{algorithmic}
\end{algorithm}

Each PB-AMD iteration consists of a forward simulation of \eqref{eq:q-dynamics}, a backward adjoint sweep using \eqref{eq:adjoint}, and the multiplicative budget update \eqref{eq:md-update} with normalization \eqref{eq:tau-normalization}. For one Gaussian path, the forward-adjoint sweep costs $O(T)$ evaluations of $g_t$ and Jacobian-transpose products and stores $O(Td)$ states. The normalization equation \eqref{eq:tau-normalization} for $\tau_k$ is a one-dimensional monotone solve, so exact budget enforcement
adds little cost relative to the forward-backward sweep. For comparison, a penalty formulation with $\beta>0$ would solve
\begin{equation*}
    \min_{u\in(0,\infty)^T}J(u)+\frac{\beta}{2}[\Phi(u)]_+^2, \qquad [x]_+:=\max\{x,0\}.
\end{equation*}

For finite $\beta$, this surrogate need not satisfy the budget constraint $\Phi(u)\le0$ in \eqref{eq:Phi} and introduces an additional parameter to tune. PB-AMD instead remains on the exact-budget simplex $Q$ in \eqref{eq:Qqmin} at every iterate. Section~\ref{sec:numerics} compares this exact-simplex approach with projected-gradient and penalty-based alternatives.

\subsection{Convergence guarantees}\label{subsec:pbamd-stationarity}
Having constructed PB-AMD, we now establish first-order stationarity
guarantees for its deterministic and stochastic implementations. We base the analysis on three properties of the reduced optimization problem: differentiability and lower boundedness of the objective, smoothness relative to the entropy geometry, and control of the error introduced by the Monte Carlo gradient estimator.

We first recall a useful property of the entropy geometry on the compact simplex $Q$ defined in \eqref{eq:Qqmin}. Since the total privacy charge is fixed at $\rho_{\max}$ and each coordinate is bounded below by $q_{\min}$, the largest admissible value of any individual charge is $q_{\max}\coloneqq \rho_{\max}-(T-1)q_{\min}$. Moreover, because every $q\in Q$ satisfies $q_t\in[q_{\min},q_{\max}]$, the Hessian of the negative entropy obeys
$q_{\max}^{-1}I_T\preceq\nabla^2\psi(q) \preceq q_{\min}^{-1}I_T$. Consequently, the associated Bregman divergence is uniformly comparable to the squared Euclidean distance:
\begin{equation}\label{eq:entropy-quadratic-comparison}
	\frac{\mu_\psi}{2}\norm{z-q}_2^2 \le D_\psi(z,q) \le\frac{M_\psi}{2}\norm{z-q}_2^2, \quad\text{ with } \mu_\psi:=\frac{1}{q_{\max}}, \quad M_\psi:=\frac{1}{q_{\min}}.
\end{equation}

This comparison allows us to relate the entropy geometry underlying the mirror-descent updates to standard smoothness and stationarity estimates. We now state the three properties required for the convergence analysis.

\begin{assumptionB}[Differentiability and lower boundedness]\label{ass:B1}
The objective $F$ is continuously differentiable on a neighborhood of $Q$ and satisfies $F(q)\geq F_{\inf}>-\infty$ for all $q\in Q$. 
\end{assumptionB}

\begin{assumptionB}[Relative smoothness]\label{ass:B2}
There exists $L_F\ge 0$ such that
\begin{align*}
    F(z)\le F(q)+ \langle\nabla F(q),z-q\rangle +L_FD_\psi(z,q), \qquad q,z\in Q.     
\end{align*}
\end{assumptionB}

\begin{assumptionB}[Unbiasedness and bounded variance of the Monte Carlo gradient]\label{ass:B3} 
Let $\mathscr F_k$ denote the sigma-field generated by the initial point, all iterates, and all oracle randomness revealed before the fresh Monte Carlo batch at iteration $k$ is drawn. In particular, $q^k$ is $\mathscr F_k$-measurable. For fresh Monte Carlo paths independent of $\mathscr F_k$, the gradient estimator $h^k$ satisfies $\mathbb E[h^k\mid\mathscr F_k] =\nabla F(q^k)$ and
\begin{align*}
    \mathbb E\left[\|h^k-\nabla F(q^k)\|_2^2 \,\middle|\,\mathscr F_k \right] \le \frac{\nu^2}{M_k}.     
\end{align*}
\end{assumptionB}

These properties provide the analytical framework for the convergence results developed below. Proposition \ref{prop:primitive-B} verifies \ref{ass:B1}, \ref{ass:B2}, and \ref{ass:B3} for fresh independent path sampling, under Assumptions \ref{ass:gaussian}, \ref{ass:lipschitz}, \ref{ass:poly}, \ref{ass:smooth}, and \ref{ass:second-order}. Thus, this verification uses the second-order regularity assumption (A7), in addition to the assumptions required for the adjoint gradient.

Because $F$ is generally nonconvex, the relevant convergence target is first-order stationarity rather than global objective suboptimality. We measure stationarity in the same entropy geometry used by the algorithm. For $\alpha>0$, define the exact mirror step and residual by
\begin{align}
	T_\alpha(q) &:=\argmin_{z\in Q} \left\{\ip{\nabla F(q)}{z}+\frac{1}{\alpha}D_\psi(z,q)\right\}, \label{eq:exact-mirror-step}
	\\
	R_\alpha(q) &:=\frac{1}{\alpha^2}D_\psi(T_\alpha(q),q). \label{eq:mirror-residual}
\end{align}

The map $T_\alpha$ is continuous, and the residual vanishes exactly at points satisfying the KKT variational inequality:
\begin{equation}\label{eq:residual-kkt-equivalence}
	R_\alpha(q)=0 \quad\Leftrightarrow\quad \ip{\nabla F(q)}{z-q}\ge0\quad\text{for every }z\in Q.
\end{equation}

Having established the regularity properties required for the analysis, we now investigate the convergence of PB-AMD. The following theorem provides first-order stationarity guarantees for both the deterministic method, based on exact gradients, and its stochastic counterpart, based on fresh Monte Carlo estimates. 

\begin{theorem}\label{thm:stochastic-pbamd}
For the fresh-path update, assume (B1)-(B3), let $q^0\in Q$, use \eqref{eq:md-implicit} with $\alpha_k\equiv\alpha>0$ satisfying $\alpha L_F\le 1/2$, and let $R$ be uniform on $\inter{K}_0$ independently of the algorithm. Then
\begin{equation}\label{eq:stochastic-rate}
    \E R_\alpha(q^R) \leq \frac{4M_\psi}{\mu_\psi} \frac{F(q^0)-F_{\inf}}{\alpha K} + \frac{5M_\psi\nu^2}{\mu_\psi^2} \left(\frac{1}{K}\sum_{k=0}^{K-1}\frac{1}{M_k}\right).
\end{equation}
Thus a constant batch size gives an $O(K^{-1}+M^{-1})$ expected-stationarity bound.  More generally,
\[
\frac{1}{K}\sum_{k=0}^{K-1}\frac{1}{M_k}\rightarrow 0 \quad\Rightarrow\quad \E[R_\alpha(q^R)]\rightarrow 0.
\]
In the exact-gradient case, assume only (B1)-(B2) and $0<\alpha L_F<1$. Then
\begin{equation}\label{eq:deterministic-rate}
    \min_{0\le k\le K-1}R_\alpha(q^k) \le \frac{F(q^0)-F_{\inf}}{\alpha^2(1/\alpha-L_F)K},
\end{equation}
$R_\alpha(q^k)\to0$, and every cluster point of $\{q^k\}$ satisfies \eqref{eq:residual-kkt-equivalence}.
\end{theorem}

Theorem \ref{thm:stochastic-pbamd} shows how the convergence of PB-AMD depends on both the number of optimization iterations and the accuracy of the Monte Carlo gradient estimates. When exact gradients are available, \eqref{eq:deterministic-rate} gives an $O(K^{-1})$ bound on $\min_{0\le k<K}R_\alpha(q^k)$. With stochastic gradients, an additional error arises from the finite number of Monte Carlo paths used at each iteration, as quantified by \eqref{eq:stochastic-rate}. In particular, if the batch size $M$ is fixed, this contribution remains of order $O(M^{-1})$, regardless of how many optimization iterations are performed. Thus, the theoretical bound does not guarantee convergence to a stationary point of the population objective with a fixed batch size. To obtain asymptotic stationarity, the batch sizes must instead increase sufficiently fast to ensure
\begin{align*}
	\frac{1}{K}\sum_{k=0}^{K-1}\frac{1}{M_k}\rightarrow0 \quad\text{ as }\quad K\to\infty.		
\end{align*}

\subsection{Fixed-path/SAA implementation}\label{subsec:saa}

As an alternative to drawing fresh Monte Carlo paths at every iteration, we now sample a fixed collection of Gaussian paths and reuse them throughout the optimization. With these paths fixed, we obtain deterministic and reproducible PB-AMD iterations, but optimize the
sample-average objective $F_M$ rather than the population objective $F$. We therefore investigate whether approximate stationarity for $F_M$ also implies approximate stationarity for $F$. To this end, for $M$ independent Gaussian paths $\xi^{(1)},\ldots,\xi^{(M)}$, we define
\begin{equation}\label{eq:saa-objective}
	F_M(q)\coloneqq\frac{1}{M}\sum_{m=1}^Mf(q;\xi^{(m)}).
\end{equation}
Here, $f(q;\xi)=L(\theta_T(q;\xi))$ is the pathwise terminal loss.

If the maps $g_t$ and the terminal loss $L$ are twice continuously differentiable, the state recursion \eqref{eq:q-dynamics} ensures that $F_M$ has a Lipschitz-continuous gradient on the compact set $Q$. Consequently, the exact-gradient convergence result of Theorem \ref{thm:stochastic-pbamd} applies to $F_M$, provided that the step size satisfies its condition with the relative-smoothness constant of the realized sample-average objective.

To relate the stationarity of the two objectives, we introduce a measure that depends directly on their gradients. For any continuously differentiable function $H$ on a neighborhood of $Q$, define the variational gap
\begin{equation}\label{eq:variational-gap}
	G_H(q)\coloneqq\max_{z\in Q}\ip{\nabla H(q)}{q-z}.
\end{equation}

We also denote by $T_{\alpha,H}$ and $R_{\alpha,H}$ the mirror step and its residual obtained by replacing $F$ with $H$ in \eqref{eq:exact-mirror-step}-\eqref{eq:mirror-residual}. Both $G_H(q)=0$ and $R_{\alpha,H}(q)=0$ characterize the KKT condition for minimizing $H$ over $Q$.

The advantage of the variational gap $G_H$ in \eqref{eq:variational-gap} is that it allows us to compare the stationarity of $F_M$ and $F$ through the difference between their gradients. The following proposition makes this connection precise, providing the basis for transferring the stationarity guarantees obtained for the sample-average problem to the original population objective.

\begin{proposition}\label{prop:saa-transfer}
	Let
$\varepsilon_M :=\sup_{q\in Q}\norm{\nabla F_M(q)-\nabla F(q)}_\infty$, $D_{Q,r}:=\max_{q,z\in Q}\norm{q-z}_r$ $(r=1,2)$, and $B_F:=\max_{q\in Q}\norm{\nabla F(q)}_2.$
	Suppose that $F$ and $F_M$ are continuously differentiable on a neighborhood of $Q$. If $q_M\in Q$ satisfies $R_{\alpha,F_M}(q_M)\le r_M$ for some $\alpha,r_M>0$, then
	\begin{equation}\label{eq:saa-stationarity-transfer}
		G_F(q_M) \le \left[\alpha\bigl(B_F+\sqrt T\,\varepsilon_M\bigr)+M_\psi D_{Q,2} \right] \sqrt{\frac{2r_M}{\mu_\psi}} + D_{Q,1}\varepsilon_M.
	\end{equation}
	
	Consequently, for any fixed $\alpha>0$, if $r_M\to0$ and $\varepsilon_M\to0$, then $G_F(q_M)\to0$. The convergence holds in the same deterministic, almost-sure, or in-probability sense as the convergence of $\varepsilon_M$, provided $r_M\to0$ deterministically.
\end{proposition}

\begin{corollary}\label{cor:saa-consistency}
Under the assumptions of Proposition~\ref{prop:primitive-B}, for independent Gaussian paths, 
\[
	\sup_{q\in Q}\norm{\nabla F_M(q)-\nabla F(q)}_\infty \rightarrow 0 \qquad\text{almost surely}.
\]
	
For any fixed $\alpha>0$, deterministic $r_M\downarrow0$, and possibly sample-dependent outputs $q_M\in Q$, $R_{\alpha,F_M}(q_M)\le r_M$ almost surely implies $G_F(q_M)\to0$ almost surely. Moreover, if $\alpha\E[\Gamma]<\mu_\psi$, with $\Gamma$ as in \eqref{eq:path-gradient-envelope}, then $\alpha$ lies below the relative-smoothness threshold of $F_M$ for all sufficiently large $M$, almost surely, and fixed-path PB-AMD can attain any prescribed positive residual tolerance.
\end{corollary}

Proposition \ref{prop:saa-transfer} quantifies how stationarity for the sample-average objective $F_M$ transfers to the population objective $F$. The bound \eqref{eq:saa-stationarity-transfer} separates two sources of error: the optimization residual tolerance $r_M$ in $R_{\alpha,F_M}(q_M)\le r_M$, which measures how accurately PB-AMD solves the sample-average problem, and the sampling error $\varepsilon_M$, which measures the uniform discrepancy between the gradients of $F_M$ and $F$. In particular, for a fixed step size, both errors must vanish to guarantee population stationarity. Corollary \ref{cor:saa-consistency} establishes that, under the stated regularity assumptions, the sampling error converges to zero almost surely as the number of independent Gaussian paths increases. It also shows that a fixed step size satisfying the population-based condition is eventually admissible for the sample-average problems. Together, these results justify the fixed-path implementation of PB-AMD: by increasing the number of sampled paths and reducing the optimization residual, one obtains asymptotically stationary solutions of the original population problem while retaining the deterministic and reproducible character of each sample-average optimization.

\subsection{Linear-quadratic model: what determines the optimal schedule?}\label{subsec:lq}

Finally, we consider a Linear-Quadratic (LQ) reduction of the dynamics to understand what determines the temporal shape of an optimal privacy schedule. We use this reduction only for interpretation and apply PB-AMD itself to the original nonlinear design dynamics. In the LQ model, we explicitly propagate the effect of noise injected at each iteration to the terminal criterion.

Let $\delta_t:=\theta_t-\bar\theta_t$. For $t\in\inter{T}_0$, the first-order linearization of the deterministic step around the reference trajectory uses $\mathsf A_t\coloneqq I_d-\eta_t\nabla g_t(\bar\theta_t)$. More generally, the analysis below applies to any deterministic matrices $\mathsf A_t\in\RR^{d\times d}$. For $t\in\inter{T}_0$, consider 
\begin{equation*}
	\delta_{t+1} = \mathsf A_t\delta_t-\frac{\eta_t\Dcal_t}{\sqrt{2q_t}}\xi_t, \qquad \delta_0=0,
\end{equation*}
with terminal loss $L(\theta_T)=\delta_T^\top W_T\delta_T$,
where $W_T\in\RR^{d\times d}$ and $W_T=W_T^\top\succeq0$.
For $0\le s\le t\le T$, define the transition matrices by
$\mathsf P_{s,t}:=\mathsf A_{t-1}\cdots\mathsf A_s$ for $s<t$
and $\mathsf P_{t,t}:=I_d$.
Then
\[
\delta_T = -\sum_{t=0}^{T-1} \mathsf P_{t+1,T}\frac{\eta_t\Dcal_t}{\sqrt{2q_t}}\xi_t.
\]

Expanding the quadratic form gives a double sum over $(s,t)$. For $s\ne t$, independence and zero mean imply
\[
\E\left[\xi_s^\top\mathsf P_{s+1,T}^\top W_T\mathsf P_{t+1,T}\xi_t\right]=0,
\]
whereas $\E[\xi_t\xi_t^\top]=I_d$ yields the diagonal trace terms. Therefore
\begin{equation}\label{eq:LQ-objective}
	F(q)=\sum_{t=0}^{T-1}\frac{a_t}{q_t}, \qquad a_t:=\frac{\eta_t^2(\Dcal_t)^2}{2} \operatorname{Tr}\left(\mathsf P_{t+1,T}^\top W_T\mathsf P_{t+1,T}\right)\ge0.
\end{equation}

The coefficient $a_t$ summarizes the terminal sensitivity of noise injected at iteration $t$: $\eta_t$ and $\Dcal_t$ set the perturbation scale, while $\mathsf P_{t+1,T}$ propagates that perturbation through the remaining dynamics. Large $a_t$ therefore means that noise at iteration $t$ is more costly for the terminal objective.

\begin{proposition}\label{prop:LQ-solution}
If all $a_t=0$, every point of $Q$ minimizes $F$ over $Q$ in \eqref{eq:LQ-objective}. Otherwise, the minimizer $q^\star$ is unique and satisfies
\begin{equation}\label{eq:LQ-waterfilling}
    q_t^\star = \max\left\{q_{\min},\sqrt{\frac{a_t}{\lambda^\star}}\right\}, \qquad \sum_{t=0}^{T-1}q_t^\star = \rho_{\max},
\end{equation}
where $\lambda^\star>0$ is uniquely determined by the budget equality. If $q_t^\star>q_{\min}$ for every $t$, then
\begin{equation}\label{eq:LQ-closed-form}
    q_t^\star = \rho_{\max} \sqrt{a_t}\left(\sum_{s=0}^{T-1}\sqrt{a_s}\right)^{-1}.
\end{equation}
\end{proposition}

In the interior case, \eqref{eq:LQ-closed-form} gives $q_t^\star\propto\sqrt{a_t}$. Thus the optimizer has no intrinsic early- or late-time preference: it allocates more privacy, and hence less noise, where downstream propagation makes perturbations more consequential. Changes in the learning rates $\eta_t$ or state-transition Jacobians $\mathsf A_t$ can therefore shift the optimal charges $q_t^\star$ across time, a mechanism
examined numerically in Section~\ref{sec:numerics}.

\begin{corollary}\label{thm:LQ-global}
Assume that at least one $a_t$ is positive, and set 
\begin{align*}
	L_{\rm LQ}\coloneqq 2q_{\min}^{-2} \max_{t\in\Tset}a_t.
\end{align*}
If exact-gradient PB-AMD uses $0<\alpha\le1/L_{\rm LQ}$, then
\begin{equation}\label{eq:LQ-rate}
    F\left(\frac1K\sum_{k=1}^Kq^k\right)-F(q^\star) \le\frac{D_\psi(q^\star,q^0)}{\alpha K},
\end{equation}
and the raw iterates converge to the unique optimizer: $q^k\to q^\star$.
\end{corollary}

\section{Numerical experiments}
\label{sec:numerics}

In this section, we evaluate how the control-theoretic framework improves terminal tracking and prediction, and how efficiently PB-AMD solves the exact-budget problem. We optimize privacy charges in \eqref{eq:q-problem} using the quadratic loss \eqref{eq:ltrain} and evaluate all schedules through the training dynamics \eqref{eq:q-dynamics}. Allocation-rule comparisons assess tracking and prediction; solver comparisons assess computational efficiency at a common first-order tolerance. All the code employed in our experiments is available on \texttt{GitHub} \cite{github}.

\subsection{Experimental design and comparison methods}\label{subsec:experiment_design}

The experiments address four questions: whether privacy allocation adapts to the training dynamics (Section \ref{subsec:allocation-regimes}), whether optimizing through the full noisy nonlinear dynamics improves on simpler allocation rules (Section \ref{subsec:nonlinear-allocation}), whether tracking gains translate into better prediction (Section \ref{subsec:predictive-performance}), and how efficiently PB-AMD solves the allocation problem (Section \ref{subsec:solver-convergence}). Moreover, we distinguish \emph{allocation-rule comparisons}, which assess tracking and prediction, from \emph{solver comparisons}, which assess computational cost for the same optimization problem.

\medskip
\paragraph{\textbf{Learning tasks and privacy budgets}.} We consider the four learning tasks summarized in Table \ref{tab:learning_taks}. The Fashion-MNIST experiments use the dataset of \cite{xiao2017fashion}.

\begin{table}[th]
	\centering
	\small
	\begin{tabular}{cccc}
		\toprule
		Task & Model and initialization & \makecell{Parameters trained \\ and perturbed} & Learning rate
		\\
		\midrule
		Tabular & \makecell{Regularized logistic regression \\ on generated data with \\ correlated features and \\ nonlinear feature interactions \\ in the data-generating model} & All model parameters & Constant
		\\[30pt]
        Fashion-constant & \makecell{Softplus classifier on \\ Fashion-MNIST, \\ initialized randomly} & All network parameters & Constant 
		\\[20pt]
		Fashion-restart & \makecell{Softplus classifier on \\ Fashion-MNIST, pretrained \\ on public data} & All network parameters & \makecell{Increased at a \\ prescribed restart}
		\\[20pt]
		Nonlinear head & \makecell{Randomly initialized Softplus \\ classifier applied to features \\ from an extractor pretrained \\ on public Fashion-MNIST data} & \makecell{Classification head only; \\ the extractor is frozen} & Constant 
		\\
		\bottomrule
	\end{tabular}
	\vspace{0.2cm}
	  \caption{Learning tasks, initialization, trainable parameters, and learning-rate settings.}
	\label{tab:learning_taks}
\end{table}

Tabular and Fashion-constant test allocation without learning-rate changes, whereas Fashion-restart tests adaptation to a learning-rate increase. Nonlinear head tests the additional value of optimizing through noisy nonlinear dynamics when learning-rate scaling alone gives a uniform allocation. Appendix \ref{sec:models} specifies the data sizes, architectures, training parameters, and Gaussian path counts.

All allocation rules are compared on the exact-budget simplex $Q$ in \eqref{eq:Qqmin}, with $q_{\min}=10^{-5}$. By \eqref{eq:q}, a larger charge $q_t$ corresponds to a smaller noise variance $u_t=(2q_t)^{-1}$. The certified budgets are $\rho_{\max}\in\{0.25,0.5,1\}$ for Tabular and the two full-network tasks, and $\rho_{\max}\in\{0.25,0.5\}$ for Nonlinear head. Before batch averaging, each record's loss gradient is smoothly rescaled to have norm at most $C_{\mathrm{clip}}$, as detailed in Appendix \ref{sec:models}. The sensitivity argument in Section \ref{subsec:DP-SGD} applies to this transformation because it requires only the per-example norm bound; smooth rescaling also permits differentiation of the design dynamics.

The total certified privacy charge is the sum of the per-step charges in \eqref{eq:q}, without subsampling amplification. All allocation comparisons therefore use the same conservative additive z-CDP certificate.

\medskip
\paragraph{\textbf{Data roles and schedule evaluation}.} Following the open-loop formulation, we separate records according to their use:
\begin{itemize}
    \setlength{\itemsep}{2pt}\setlength{\parskip}{0pt}
    \item \emph{Public auxiliary data} provide preprocessing and, where required, pretraining.
    \item \emph{Public-design data} determine the schedules. For Fashion-MNIST, these records form a subset of the public auxiliary data.
    \item \emph{Private training data} are disjoint from public records and are used only after schedules have been fixed.
    \item \emph{Held-out test data} assess prediction after training.
\end{itemize}

For any training sample, the \emph{No-noise reference} is the terminal model obtained with the same initialization, batches, and training updates, but without Gaussian noise. Public-design and private training samples therefore have separate No-noise references. Schedule construction and evaluation follow three stages:
\begin{itemize}
    \setlength{\itemsep}{2pt}\setlength{\parskip}{0pt}
    \item \emph{Design}. PB-AMD (Algorithm~\ref{alg:pbamd}) minimizes the sample-average objective $F_M$ in \eqref{eq:saa-objective} on public-design data, using that sample’s No-noise reference as the tracking target.
    \item \emph{Selection}. If different parameter settings or saved iterates produce several candidate schedules, we select the schedule with the lowest mean tracking error on the same public-design data, evaluated using independent Gaussian paths.
    \item \emph{Private evaluation}. We freeze the selected schedules before private training and evaluate the tracking error $J(u)$ in \eqref{eq:J} against the No-noise reference computed on the private training sample. Held-out test records are used for predictive evaluation.
\end{itemize}

Thus, private records are not used to construct or select the schedules. The stated privacy budget certifies an individual noisy training run under a fixed schedule. The noiseless reference computed on private records and the associated tracking diagnostics are used only for offline evaluation and are not outputs covered by this guarantee. Joint release of multiple training runs or private-data diagnostics would require separate privacy accounting. Appendix \ref{sec:evaluation} and \ref{sec:utility-solvers} give the complete design, validation, and evaluation protocol.

\medskip
\paragraph{\textbf{Allocation-rule comparisons}.} These comparisons test the value of jointly optimizing all privacy charges through the full noisy nonlinear dynamics. We use the baselines of Table \ref{tab:allocation}.

\begin{table}[th]
	\centering
	\small
	\begin{tabular}{cc}
		\toprule
		Rule & Allocation principle
		\\
		\midrule
		Uniform & Assigns $q_t=\rho_{\max}/T$, providing the reference for assessing redistribution 
		\\[10pt]
        \makecell{Learning-rate \\ allocation (LR)} & \makecell{Uses weights $w_t=\eta_t$, isolating learning-rate scaling from \\ subsequent training dynamics}
		\\[10pt]
		Decay & Combines learning-rate scaling with subsequent regularization shrinkage alone
		\\[10pt]
		Linearized & \makecell{Uses $w_t=\sqrt{a_t}$ with $a_t$ from \eqref{eq:LQ-objective}, $W_T=\mathrm{Id}$, and Jacobians \\ evaluated along the public-design No-noise reference}
        \\[20pt]
		Geometric & Uses $w_t=\eta_t e^{\gamma t/(T-1)}$, with $\gamma$ selected by mean tracking error on public-design data 
		\\
		\bottomrule
	\end{tabular}
	\vspace{0.2cm}
	  \caption{Baseline allocation rules compared with PB-AMD under a common certified privacy budget.}
	\label{tab:allocation}
\end{table}

For LR, Decay, Linearized, and Geometric, weights are converted to feasible charges through $q_t=\max\{q_{\min},cw_t\}$, where $c>0$ is chosen to meet the common budget. LR, Decay, and Linearized simplify Proposition~\ref{prop:LQ-solution} in different ways. Geometric adapts the exponential schedule family of \cite{hong2022dynamic}, testing whether a tuned single-parameter profile can match PB-AMD’s optimization of all $T$ charges.

\medskip
\paragraph{\textbf{Solver comparisons}.} To assess computational efficiency separately from allocation quality, we fix $F_M$ and the training and adjoint routines, and compare solvers at a common first-order tolerance:
\begin{itemize}
    \setlength{\itemsep}{2pt}\setlength{\parskip}{0pt}
    \item \emph{Projected gradient descent (PGD)} uses the same adjoint gradient as PB-AMD, but projects additive Euclidean steps onto $Q$, instead of using multiplicative reallocation.
    \item \emph{Penalty} uses logarithmic variances and the quadratic budget-excess penalty (4.4), following the penalty-method framework of \cite{cplNocedalWright2006}.
    \item \emph{Penalty + repair} rescales the raw charges above the floor to enforce the exact budget, enabling feasible utility and timing comparisons.
\end{itemize}

Appendix \ref{sec:utility-solvers} specifies the allocation weights, solver settings, candidate selection, and repair procedure.

\medskip 
\paragraph{\textbf{Repetitions and statistical reporting}.} Each task–budget comparison uses ten problem seeds shared across methods; the two Nonlinear-head budgets use independent seed groups. Within each evaluation stage, methods share standard Gaussian vectors to obtain paired comparisons. We report the mean across seeds of the tracking-error reduction
\begin{align*}
    100\left(1-\frac{J_{\mathrm{PB}}}{J_{\mathrm{base}}}\right),
\end{align*}
where $J_{\mathrm{PB}}$ and $J_{\mathrm{base}}$ are path-averaged tracking errors for PB-AMD and the comparator. Positive values favor PB-AMD; for example, a reduction of $10\%$ means that its tracking error is $90\%$ of the comparator’s error. Reported wins count seeds with lower PB-AMD tracking error.

Confidence intervals across the ten problem seeds are unadjusted for multiple testing. Statistical significance is assessed using Holm's correction \cite{holm1979simple}, controlling the probability of any false rejection at $5\%$ within each comparison group. Appendix \ref{sec:evaluation} specifies these groups and the statistical protocol.

\subsection{Temporal adaptation of privacy allocation}\label{subsec:allocation-regimes}

\begin{figure}[htbp]
    \centering
    \includegraphics[width=0.95\textwidth,trim=0 5.5pt 0 0,clip]{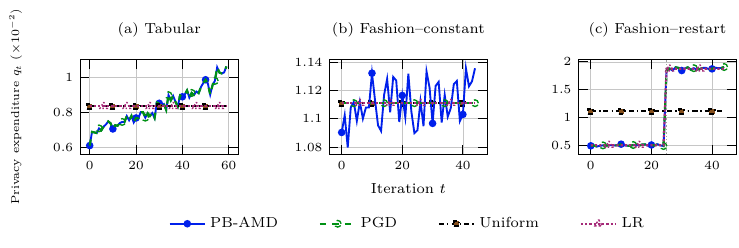}
    \caption{Single-instance exact-budget schedules at $\rho_{\max}=0.5$. LR coincides with Uniform in (a,b); the vertical dotted line in (c) marks the learning-rate restart.}  \label{fig:allocation-regimes}
\end{figure}

We first examine how privacy allocation responds to the training dynamics. Tabular and Fashion-constant test whether redistribution improves tracking at a constant learning rate, whereas Fashion-restart tests adaptation to a learning-rate increase. Figure \ref{fig:allocation-regimes} illustrates one instance per task at $\rho_{\max}=0.5$; the tracking-error comparisons below use all ten problem seeds at each of the three budgets and are evaluated on private training data.

For Tabular, PB-AMD allocates more privacy expenditure to later iterations despite the constant learning rate (Figure \ref{fig:allocation-regimes}(a)). This redistribution reduces tracking error by $1.04\%$–$1.78\%$ relative to Uniform across the tested budgets. Since LR coincides with Uniform in this setting, the improvement reflects an allocation benefit beyond learning-rate scaling.

For Fashion-constant, PB-AMD instead produces an approximately uniform allocation (Figure \ref{fig:allocation-regimes}(b)), with tracking errors differing from Uniform by less than $0.02\%$. Thus, nonuniform expenditure offers essentially no tracking benefit in this task. Together, the two constant-rate experiments show that the usefulness of redistribution depends on the training dynamics: a constant learning rate alone does not determine whether a nonuniform schedule is beneficial.

For Fashion-restart, PB-AMD sharply increases expenditure after the learning-rate increase (Figure \ref{fig:allocation-regimes}(c)). PB-AMD and LR reduce tracking error relative to Uniform by approximately $27.45\%$ and $27.47\%$, respectively, at each budget. Learning-rate scaling therefore captures the substantial restart benefit, although it misses the smaller improvement in Tabular. PGD produces allocation profiles close to those of PB-AMD, indicating agreement between the two solvers on these instances.

We also examine sensitivity to the public-design sample. At $\rho_{\max}=0.5$, we perturb that sample and recompute the PB-AMD schedule, keeping the private training data unchanged. The redesigned schedule retains an approximately $27.45\%$ tracking-error reduction relative to Uniform (Appendix \ref{sec:public}). This shows that the improvement over Uniform survives the particular public-design perturbation tested, without using private records for schedule selection; it does not establish robustness to general public–private distribution shifts. Theorem \ref{thm:design-transfer} provides a conditional explanation for such transfer: under its global Lipschitz and uniform discrepancy assumptions, it bounds the target-objective suboptimality of a near-optimal design schedule. The bound depends on the model constants and training horizon and is not numerically evaluated here.

These experiments distinguish three outcomes: a small benefit beyond learning-rate scaling in Tabular, negligible benefit from redistribution in Fashion-constant, and a substantial restart benefit already captured by LR. PB-AMD accommodates all three without imposing a fixed preference for early or late expenditure.

\subsection{Value of nonlinear schedule optimization}\label{subsec:nonlinear-allocation}

The Fashion-restart experiment shows that learning-rate scaling can capture a substantial allocation benefit. We now use Nonlinear head to test whether optimizing through the full noisy nonlinear dynamics improves on simpler allocation rules. Here the learning rate is constant, so LR coincides with Uniform, while weak regularization makes Decay nearly uniform. Geometric tunes a single shape parameter using public-design tracking error. Linearized accounts for perturbation propagation through sensitivities computed along the public-design No-noise reference. In contrast, PB-AMD jointly optimizes all $T$ charges using the adjoint gradient \eqref{eq:mc-gradient} evaluated along noisy trajectories.

\paragraph{Aggregate tracking performance.} Table \ref{tab:nonlinear-allocation} reports paired tracking-error reductions on private training data for ten instances at each budget. PB-AMD outperforms LR, Decay, and Geometric in all ten instances at both budgets. At $\rho_{\max}=0.5$, the mean reductions are $7.72\%$, $7.85\%$, and $3.32\%$, respectively; at $\rho_{\max}=0.25$, they are $4.32\%$, $4.41\%$, and $2.16\%$. These gains remain statistically significant after Holm correction across the eight comparisons involving four allocation rules and two budgets. Thus, in this task, optimizing all charges improves tracking beyond both learning-rate scaling and a tuned geometric profile.

The comparison with Linearized is less conclusive. PB-AMD achieves mean reductions of $5.07\%$ at $\rho_{\max}=0.5$ and $2.75\%$ at $\rho_{\max}=0.25$, winning in eight and seven instances, respectively. Neither comparison remains statistically significant after the same correction. 

\begin{table}[htbp]
    \centering
    \small
    \begin{tabular*}{\textwidth}{@{\extracolsep{\fill}}lcccc@{}}
        \toprule
        & \multicolumn{2}{c}{$\rho_{\max}=0.5$} & \multicolumn{2}{c}{$\rho_{\max}=0.25$} 
        \\
        \cmidrule(lr){2-3}\cmidrule(l){4-5} 
        Comparator & Error reduction [95\% CI] & Wins & Error reduction [95\% CI] & Wins 
        \\
        \midrule
        LR & $7.72\ [6.03,9.41]$ & $10/10$ & $4.32\ [3.55,5.09]$ & $10/10$ 
        \\
        Decay & $7.85\ [6.15,9.55]$ & $10/10$ & $4.41\ [3.63,5.18]$ & $10/10$ 
        \\
        Linearized & $5.07\ [-2.10,12.23]$ & $8/10$ & $2.75\ [0.35,5.15]$ & $7/10$ 
        \\
        Geometric & $3.32\ [2.59,4.04]$ & $10/10$ & $2.16\ [1.77,2.56]$ & $10/10$ 
        \\
        \bottomrule
    \end{tabular*}
    \vspace{6pt}
    \caption{Nonlinear-head training: mean paired PB-AMD tracking-error reductions (\%) on private training data relative to each comparator, with unadjusted $95\%$ confidence intervals. \emph{Wins} counts seeds with lower PB-AMD tracking error, out of ten per budget.}
    \label{tab:nonlinear-allocation}
\end{table}

\begin{figure}[h]
    \centering
    \includegraphics[width=0.95\textwidth]{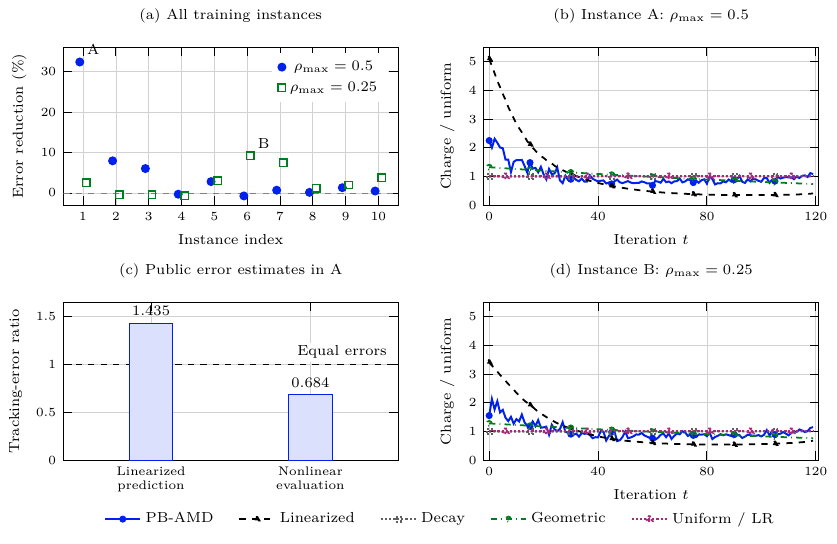}
    \caption{Nonlinear-head allocation: (a) instance-level PB-AMD tracking-error reductions relative to Linearized on private training data, for two independent groups of ten seeds; (b,d) schedules for instances A and B, selected as the largest-gain examples at their respective budgets, with charges divided by the uniform charge. Uniform and LR coincide; (c) PB-AMD-to-Linearized error ratios for the two fixed schedules in instance A on public-design data, comparing the reference linearized prediction with full nonlinear evaluation using fresh Gaussian paths. Positive reductions in (a) and ratios below one in (c) favor PB-AMD.}\label{fig:nonlinear-mechanism}
\end{figure}

The confidence intervals in Table \ref{tab:nonlinear-allocation} are unadjusted; consequently, the interval excluding zero at $\rho_{\max}=0.25$ does not imply significance under the corrected tests. Figure \ref{fig:nonlinear-mechanism}(a) displays all instance-level comparisons, including those favoring Linearized. These results support an advantage over the simpler rules, while leaving the average advantage over Linearized statistically unresolved.

\paragraph{Selected examples of nonlinear reallocation.} To examine how the schedules differ, we consider instances A ($\rho_{\max}=0.5$) and B ($\rho_{\max}=0.25$). These were selected after the full comparison because they produced the largest PB-AMD gains over Linearized at their respective budgets: $32.40\%$ and $9.20\%$ on private training data. They therefore illustrate cases where nonlinear optimization is particularly beneficial, rather than typical gains across instances. Appendix \ref{sec:evaluation} reports Gaussian-path uncertainty intervals adjusted for this selection.

Figures \ref{fig:nonlinear-mechanism}(b,d) show each charge divided by the uniform charge. A value of one represents both Uniform and LR, with Decay remaining close to this level. Both PB-AMD and Linearized favor early expenditure, but PB-AMD allocates less to the earliest steps and redistributes part of that expenditure to later iterations. In these two instances, this less concentrated allocation accompanies lower tracking error than Linearized.

\paragraph{Linearized prediction versus nonlinear evaluation.} For instance A, we further test whether the reference linearization correctly ranks the two schedules. We fix the schedules produced by PB-AMD and Linearized and evaluate both on the same public-design records, first using the linearized objective \eqref{eq:LQ-objective} and then using the full noisy dynamics with fresh Gaussian paths.

Figure \ref{fig:nonlinear-mechanism}(c) reports the ratio of PB-AMD’s error to Linearized’s error; values below one favor PB-AMD. The linearized prediction gives a ratio of $1.435$, favoring Linearized. Full nonlinear evaluation reverses this ranking, giving PB-AMD $31.65\%$ lower tracking error, consistent with its advantage on private training data. This example shows that sensitivities computed along a fixed No-noise reference can favor a schedule that performs worse under the actual noisy nonlinear dynamics, explaining why optimizing those dynamics can improve allocation.

\subsection{Predictive consequences of privacy allocation}\label{subsec:predictive-performance}

We next examine whether the tracking gains observed for Fashion-restart translate into better prediction. We train on private records using the previously fixed schedules and evaluate the resulting models on held-out test records. Tracking error $J(u)$ in \eqref{eq:J}, with the quadratic loss \eqref{eq:ltrain}, measures squared parameter distance to the No-noise reference trained on the same private sample. Predictive performance is assessed separately through test accuracy, the fraction of correctly classified records, and unregularized cross-entropy, which penalizes low probabilities assigned to the true labels.

\begin{figure}[!htbp]
    \centering
    \includegraphics[width=0.95\textwidth]{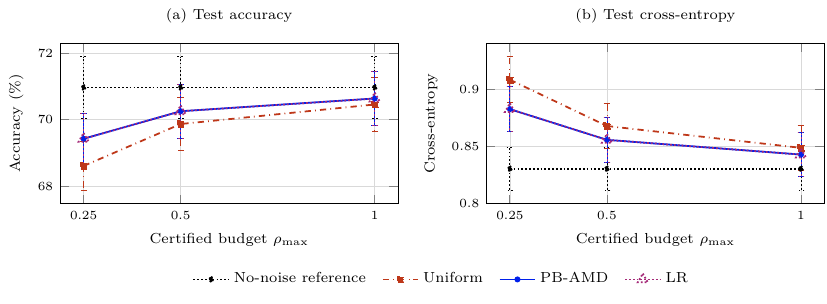}
    \caption{Test performance for Fashion-restart models trained on private data: means and pointwise $95\%$ confidence intervals over ten problem seeds. Each noise schedule uses $256$ Gaussian training runs per seed.}\label{fig:network-performance}
\end{figure}

Figure \ref{fig:network-performance} reports Fashion-restart results across the three certified budgets. PB-AMD improves test accuracy and reduces cross-entropy relative to Uniform at every budget. At $\rho_{\max}=0.25$, mean accuracy rises from $68.61\%$ to $69.43\%$, an increase of $0.82$ percentage points. At each budget, PB-AMD closes approximately one-third of the accuracy and cross-entropy gaps between Uniform and the No-noise reference. LR nearly coincides with PB-AMD in both metrics, consistently with their similar tracking gains in Section \ref{subsec:allocation-regimes}.

The statistical analysis treats the predictive comparisons for Fashion-constant and Fashion-restart as one family: two tasks, three budgets, and two metrics, giving twelve tests. Although Figure \ref{fig:network-performance} focuses on Fashion-restart, its significance assessment includes the Fashion-constant comparisons in this correction. 

For Fashion-restart, the improvements in test accuracy and test cross-entropy relative to Uniform remain significant after Holm correction across the twelve-comparison family; Appendix \ref{sec:evaluation} details the statistical protocol.

The figure shows means and pointwise $95\%$ confidence intervals across ten problem seeds, with each noisy schedule evaluated using $256$ Gaussian training runs per seed. These results demonstrate that, in the Fashion-restart setting, an allocation optimized for parameter tracking also improves held-out prediction at a fixed privacy budget.

\subsection{Solver convergence and computational efficiency}\label{subsec:solver-convergence}

This section compares the time required by PB-AMD, PGD, and Penalty + repair to reach the same first-order tolerance for the exact-budget problem \eqref{eq:q-problem}. We use ten new public-design instances each of Tabular, Fashion-restart, and Nonlinear head at $\rho_{\max}=0.5$. Within each instance, all solvers share the tracking term $F_M$ in \eqref{eq:saa-objective}, Uniform initialization $q^0$, minibatches, and Gaussian paths. Fixed stepsizes and penalty parameters are calibrated on separate public-design instances before timing; for Nonlinear head, PB-AMD and PGD use the same sufficient-decrease line search.

We measure time to $G_{F_M}(q)/F_0\le10^{-4}$, where $G_{F_M}$ is the variational gap in \eqref{eq:variational-gap} and $F_0=F_M(q^0)$ is the objective at Uniform initialization. The gap is zero exactly when the constrained first-order conditions hold, giving a common stationarity criterion for the three solvers. For Penalty + repair, we evaluate it at the repaired charges. Timed solves include required objective and adjoint evaluations, line search, and repair, but exclude calibration and evaluations used only to draw the curves. Private training and prediction scoring are not timed.

\begin{figure}[!htbp]
  	\centering
  	\includegraphics[width=0.95\textwidth]{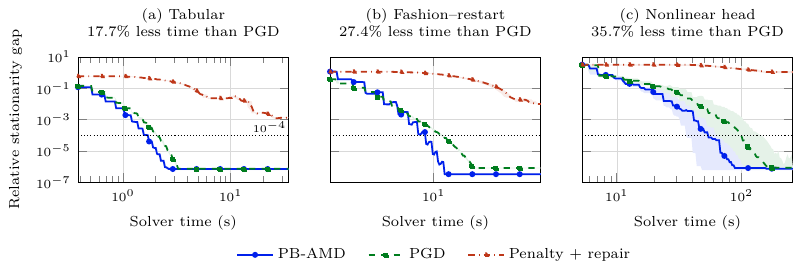}
	\caption{Solver convergence on public-design instances at $\rho_{\max}=0.5$: median normalized gap $G_{F_M}(q)/F_0$ from \eqref{eq:variational-gap}, with inter-quartile bands over ten convergence instances per task. The horizontal dotted line is the $10^{-4}$ timing target.}\label{fig:solver-efficiency}
\end{figure}

At this tolerance, PB-AMD reduces solver time relative to PGD by $17.7\%$ for Tabular (Figure~\ref{fig:solver-efficiency}(a)), $27.4\%$ for Fashion-restart (Figure~\ref{fig:solver-efficiency}(b)), and $35.7\%$ for Nonlinear head (Figure~\ref{fig:solver-efficiency}(c)).

These percentages use the median ratio of PGD time to PB-AMD time over instances where both solvers reach the target; the curves include all ten instances per task. 

Both methods reach the target in all ten Tabular instances, all ten Fashion-restart instances, and the same nine head instances. None of the $30$ repaired penalty runs reaches this tolerance within the update caps. 

PB-AMD maintains the exact budget at every update and has lower median schedule-optimization time than PGD among the tested instances where both methods reach the common first-order tolerance.

Additional results on raw penalty-budget violations, exact-budget preservation, returned-schedule tracking errors, and agreement with the analytical linear-quadratic optimizer are reported in Appendices \ref{sec:penalty}
and \ref{sec:numerical-checks}. Appendices \ref{sec:matched}-\ref{sec:timing} give the complete convergence and timing protocol.

\section{Conclusion}\label{sec:conclusion}

We have developed a control-theoretic mathematical framework and an exact-budget optimization method for temporal privacy allocation in differentially private stochastic gradient descent. By treating the complete Gaussian noise-variance schedule as an open-loop control, the framework relates the timing of privacy expenditure to its effect on the final model under a composed zero-concentrated differential privacy constraint. At the same time, separating schedule design from private training makes the privacy cost of schedule selection an explicit part of the certification.

The central structural result is budget saturation for expected quadratic terminal tracking: every locally optimal schedule exhausts the certified budget. Consequently, the change from noise variances to per-iteration privacy charges yields an exact-budget simplex. Privacy-Budget Adjoint Mirror Descent (PB-AMD) exploits this geometry to preserve positivity and the prescribed budget at every update without penalty-parameter tuning. The analysis supports this construction with existence and stability results for optimal schedules and stationarity guarantees for the optimization method.

Numerical experiments show that the optimized allocations adapt to the training dynamics. In the Nonlinear-head task, PB-AMD improves tracking relative to learning-rate, decay, and tuned geometric allocation, while its average advantage over Linearized remains statistically unresolved after multiplicity correction. In Fashion-restart, PB-AMD improves prediction relative to Uniform, with learning-rate allocation achieving similar performance. Comparisons at a common stationarity tolerance show lower schedule-optimization times than PGD on the tested problems. These results support the framework as a method for optimizing certified privacy allocations through nonlinear training dynamics, while leaving its predictive advantage over the strongest allocation baselines open.

\appendix
\section{Differential privacy preliminaries}\label{app:DP}

This appendix collects the formal privacy notions underlying the accounting framework of Section \ref{sec:DP}. The material is classical \cite{bun2016concentrated,dwork2014algorithmic,dwork2016concentrated}; we include it to make precise the definitions and properties used in the privacy certification of the iterative Gaussian mechanisms considered in the paper.

\subsection{Adjacency and differential privacy}

We begin with the notion of replace-one adjacency and then recall R\'enyi divergence and the definition of zCDP.

\begin{definition}\label{def:neighboring}
Let $(\MC X,\Sigma_{\MC X})$ be a measurable sample space, fix $N\in\NN$, and set $\MC U_N\coloneqq \MC X^N$. Two datasets $D,D'\in\MC U_N$ are called \emph{neighboring}, written $D\sim D'$, if they differ in exactly one coordinate. 
\end{definition}

Throughout this work, privacy guarantees are expressed using zero-Concentrated Differential Privacy (zCDP). This notion is particularly convenient for privacy accounting in the iterative Gaussian mechanism underlying DP-SGD. Its definition is based on R\'enyi divergence. 
\begin{definition}\label{def:reny}
Let $P$ and $Q$ be probability measures. For $\alpha>1$ and $P\ll Q$, the \emph{Rényi divergence} of order $\alpha$ is
\begin{align*}
    \MF D_\alpha(P\|Q) \coloneqq \frac{1}{\alpha-1} \log\int \left(\frac{dP}{dQ}\right)^\alpha \,dQ.    
\end{align*}
If $P\not\ll Q$, set $\MF D_\alpha(P\|Q)=+\infty$.
\end{definition}

\begin{definition}\label{def:zcdp}
A randomized mechanism $\MC M:\MC U_N\rightsquigarrow\MC O$ satisfies $\rho$-zCDP, with $\rho\geq 0$, if, for every $D\sim D'$ and every $\alpha>1$, 
\begin{align*}
    \MF D_\alpha\big(\MC L(\MC M(D)) \,\|\, \MC L(\MC M(D'))\big) \leq \rho\alpha,    
\end{align*}
where $\MC L(X)$ denotes the law of a random variable $X$.    
\end{definition}

The parameter $\rho$ will be referred to as the zCDP privacy charge. Its main advantage for the purposes of this paper is its additive behavior under composition. Consider a finite sequence $\{\MC M_t\}_{t=0}^{T-1}$ of possibly adaptive randomized mechanisms. Suppose that, conditionally on all outputs and public information available before iteration $t$, the mechanism used at that iteration satisfies $\rho_t$-zCDP. Then the complete mechanism after $T$ iterations satisfies $\rho_{\mathrm{tot}}$-zCDP with
\begin{align}\label{eq:rho_tot}
	\rho_{\mathrm{tot}} = \sum_{t=0}^{T-1}\rho_t.
\end{align}

Given a prescribed total zCDP budget $\rho_{\max}>0$, a sequence of mechanisms is therefore certified whenever $\rho_{\mathrm{tot}}\leq \rho_{\max}$. 

In this accounting process, the mechanisms need not be independent and the query used at a given iteration may depend on the preceding private outputs. What is required is that, after conditioning on the previous transcript, a valid zCDP bound $\rho_t$ holds uniformly for the conditional mechanism at that iteration \cite{dwork2016concentrated}.

This adaptive composition property is particularly suited to private stochastic optimization. The current optimization iterate depends on all preceding noisy updates, but once the preceding transcript is conditioned upon, the iterate is fixed and the privacy charge of the next Gaussian query can be bounded conditionally. The individual bounds may then be summed to obtain a privacy guarantee for the complete training procedure.

\subsection{Gaussian mechanism and privacy accounting as a resource constraint} Let $f:\MC U_N\to\RR^d$ be a measurable query over the dataset universe $\MC U_N$, with finite $\ell_2$-sensitivity
\begin{align*}
	\Delta_{2,f}\coloneqq\sup_{D\sim D'}\|f(D)-f(D')\|_2.	
\end{align*}

For $s>0$, consider the Gaussian mechanism $\MC M_s(D) = f(D)+s\xi$, with $\xi\sim\MC N(0,I_d)$. For two Gaussian distributions having the same covariance $s^2I_d$, their Rényi divergence satisfies
\begin{align*}
	\MF D_\alpha \Big(\MC N(f(D),s^2I_d) \big\| \mathcal N(f(D'),s^2I_d)\Big) = \frac{\alpha}{2s^2} \|f(D)-f(D')\|_2^2 \leq
	\alpha\frac{\Delta_{2,f}^2}{2s^2}.
\end{align*}

Hence the Gaussian mechanism is $\rho$-zCDP with $\rho = \Delta_{2,f}^2/(2s^2)$. This relation is the calibration formula used throughout the paper. It shows directly the trade-off between noise and privacy expenditure: for fixed sensitivity, increasing the Gaussian standard deviation decreases the zCDP charge, whereas decreasing the noise increases the privacy expenditure.

Combining Gaussian calibration and adaptive composition gives the accounting structure used throughout the paper. If the Gaussian mechanism applied at iteration $t$ has certified charge $\rho_t$, then $\rho_{\mathrm{tot}}$ defined in \eqref{eq:rho_tot} is a valid zCDP bound for the full sequence of private updates.

A prescribed privacy budget $\rho_{\max}$ therefore imposes the resource constraint $\rho_{\mathrm{tot}} \leq \rho_{\max}$. A \emph{privacy schedule} specifies how this total expenditure is distributed across iterations. The accountant specifies which temporal allocations are privacy feasible, but does not identify which allocation is preferable for learning. In the setting of this paper, the latter question is treated as an optimization problem: the total certified privacy resource is distributed across iterations so as to optimize the terminal performance criterion.

\subsection{Post-processing}

DP is invariant under data-independent post-processing. If $\MC M$ is $\rho$-zCDP and $\MC K$ is a randomized transformation whose conditional distribution depends only on the output of $\MC M$ and on randomness independent of the private dataset, then $\MC K\circ \MC M$ is also $\rho$-zCDP. 

This property is used repeatedly in the DP-SGD construction of Section \ref{subsec:DP-SGD}. Once a data-dependent gradient query has been privatized by Gaussian perturbation, deterministic operations performed on that private output do not generate an additional privacy charge. Similarly, releasing only the terminal iterate of an already private training transcript cannot increase the privacy loss.

These properties, together with the Gaussian calibration above, provide the privacy accounting used throughout the paper; see, e.g., \cite{bun2016concentrated}.

\section{Proofs of the results in Section \ref{sec:open_loop}}\label{app:proofs_open_loop}

\subsection{Proofs: Exact certified-budget saturation for terminal tracking}

\begin{proof}[Proof of Theorem \ref{thm:budget-saturation}]
Let $\MC F_{T-1}:=\sigma(\xi_0,\ldots,\xi_{T-2})$, interpreted as the trivial sigma-field when $T=1$, and define the $\MC F_{T-1}$-measurable vector $Y\coloneqq\theta_{T-1}-\eta_{T-1}g_{T-1}(\theta_{T-1})-\theta^\star$.

Lemma~\ref{lem:moments} (applied to the singleton control set) and the linear-growth condition \ref{ass:lipschitz} give $Y\in L^2$. Moreover, $\theta_T-\theta^\star = Y-\eta_{T-1}\Dcal_{T-1}\sqrt{u_{T-1}}\,\xi_{T-1}$. Conditioning on $\MC F_{T-1}$ and using independence, zero mean, and $\E\norm{\xi_{T-1}}^2=d$ gives \eqref{eq:last-step-affine}.

If a feasible local minimizer had $\rho(u^\star)<\rho_{\max}$, then for every sufficiently small $h>0$ the point $u^\star-h\mathsf e_{T-1}$, where $\mathsf e_{T-1}$ is the last coordinate vector of $\RR^T$, would remain positive and feasible and would lie in an arbitrarily small neighborhood of $u^\star$. Equation~\eqref{eq:last-step-affine} would lower the objective by $d\eta_{T-1}^2(\Dcal_{T-1})^2h>0$, contradicting local optimality.
\end{proof}

\subsection{Proofs: Privacy-charge reformulation and exact-budget simplex}

\begin{proof}[Proof of Proposition \ref{prop:q-equivalence}]
The transformation $q_t=(2u_t)^{-1}$, with inverse $u_t=(2q_t)^{-1}$, gives the asserted bijection and preserves the dynamics and objective. Let $S_u$ denote the nonempty compact minimizer set of \eqref{eq:open_loop}. Its image 
\begin{align*} 
	S_q\coloneqq \left\{\big((2u_t)^{-1}\big)_{t=0}^{T-1}:u\in S_u\right\}
\end{align*} 
is nonempty and compact. Theorem \ref{thm:budget-saturation} implies $S_q\subset Q^\circ$; together with the bijection, this gives 
\begin{align*}
	S_q=\argmin_{q\in Q^\circ}F(q).
\end{align*}
Since every coordinate is positive on $S_q$, compactness yields $\delta\coloneqq\min_{q\in S_q}\min_{0\leq t<T}q_t>0$.

Moreover, $\delta\leq\rho_{\max}/T$. Set $q_\ast=\delta/2$. Then $q_\ast\in(0,\rho_{\max}/T)$, and for every $0<q_{\min}<q_\ast$, we have $S_q\subset Q(q_{\min})\subset Q^\circ$. These inclusions show that the minimum over $Q(q_{\min})$ equals that over $Q^\circ$, and that all minimizers over $Q^\circ$ belong to $Q(q_{\min})$. Consequently,
\begin{align*}
    \argmin_{q\in Q(q_{\min})}F(q) = S_q = \argmin_{q\in Q^\circ}F(q).
\end{align*}
\end{proof}

\subsection{Proofs: Differentiability, adjoints, and KKT conditions}

\begin{proof}[Proof of Proposition \ref{prop:u-adjoint}]
Fix a compact set $K\subset(0,\infty)^T$.  For every noise path, the finite recursion \eqref{eq:controlled-dynamics} is a composition of $C^1$ maps of $u$.  If
$S_{s,t}:=\partial_{u_t}\theta_s$, then $S_{s,t}=0$ for $s\leq t$, while
\[
S_{t+1,t} = -\frac{\eta_t\Dcal_t}{2\sqrt{u_t}}\xi_t, \qquad S_{s+1,t}=A_s(\theta_s)S_{s,t}\quad\text{ for } s\geq t+1.
\]
Consequently,
\[
\partial_{u_t}j(u;\xi) = \ip{\nabla L(\theta_T)}{S_{T,t}} = -\frac{\eta_t\Dcal_t}{2\sqrt{u_t}} \ip{p_{t+1}}{\xi_t},
\]
which proves \eqref{eq:path-u-gradient}. It remains to justify expectation and differentiation interchange. Assumption~\ref{ass:smooth} gives 
\begin{align*}
	\norm{A_s(\theta_s)}_{\rm op} \le 1+\eta_sK_g\Big(1+\norm{\theta_s}^{r_g}\Big).
\end{align*}
Lemma~\ref{lem:moments} provides a constant $C_K$ such that, uniformly over $u\in K$,
\[
\max_{0\le s\le T}\norm{\theta_s(u;\xi)} \le C_K\left(1+\sum_{s=0}^{T-1}\norm{\xi_s}\right) \quad\text{almost surely}.
\]

The finite adjoint recursion and the polynomial growth of $\nabla L$ consequently bound every $p_s$ by a polynomial in the Gaussian norms. Because $\inf_{u\in K}u_t>0$, there are deterministic constants $C_{K,t}$ and an integer $\bar r$ for which
\begin{equation}\label{eq:u-gradient-dominator}
    \sup_{u\in K}|\partial_{u_t}j(u;\xi)| \le C_{K,t}\left(1+\sum_{s=0}^{T-1}\norm{\xi_s}\right)^{\bar r}.
\end{equation}

The right-hand side is integrable, indeed it has moments of every finite order. For almost every $\xi$, the state, adjoint, and pathwise derivative are continuous in $u$.

To justify differentiation under the expectation at an arbitrary $u^\circ\in(0,\infty)^T$, choose a compact rectangular neighborhood $K'\Subset(0,\infty)^T$ of $u^\circ$. Let $\mathsf e_t$ denote the $t$-th coordinate vector of $\RR^T$. For every sufficiently small nonzero $h$, the segment $\{u^\circ+\tau h\mathsf e_t:0\le\tau\le1\}$ lies in $K'$. The one-dimensional mean-value theorem and \eqref{eq:u-gradient-dominator}, with $K'$ in place of $K$, give
\[
\left|\frac{j(u^\circ+h\mathsf e_t;\xi)-j(u^\circ;\xi)}{h}\right| \le \sup_{v\in K'}|\partial_{u_t}j(v;\xi)| \le C_{K',t} \left(1+\sum_{s=0}^{T-1}\norm{\xi_s}\right)^{\bar r}.
\]
Thus the difference quotients admit an integrable envelope.

Dominated convergence now yields \eqref{eq:u-gradient}.  Applying the same domination to a convergent sequence of controls gives continuity of each partial derivative on $K$.  Since $K$ was arbitrary, $J\in C^1((0,\infty)^T)$.
\end{proof}

\begin{proof}[Proof of Theorem \ref{thm:KKT}]
By Proposition \ref{prop:u-adjoint}, $J\in C^1((0,\infty)^T)$. The privacy constraint $\Phi(u)$ in \eqref{eq:Phi} is also continuously differentiable, with $\nabla\Phi(u)=\big(-(2u_t^2)^{-1}\big)_{t=0}^{T-1}$.

If $\Phi(u^\star)=0$, the direction $v=-\nabla\Phi(u^\star)$ satisfies $\langle\nabla\Phi(u^\star),v\rangle =-\|\nabla\Phi(u^\star)\|^2<0$. If $\Phi(u^\star)<0$, there are no active constraints. Thus the Mangasarian-Fromovitz constraint qualification (MFCQ) holds at $u^\star$. The finite-dimensional KKT theorem \cite[Chapter~12]{cplNocedalWright2006} therefore provides a deterministic multiplier $\lambda^\star\geq0$ such that
\begin{align*}
    \lambda^\star\Phi(u^\star)=0 \qquad\text{ and }\qquad \partial_{u_t}J(u^\star) - \frac{\lambda^\star}{2(u_t^\star)^2}=0 \;\text{ for all } t\in\inter{T}_0.
\end{align*}
For the state and adjoint associated with $u^\star$, formula \eqref{eq:u-gradient} therefore gives
\begin{align*}
    -\frac{\eta_t\Delta b_t}{2\sqrt{u_t^\star}} \mathbb E\left[\left\langle p_{t+1}^\star,\xi_t\right\rangle\right]-\frac{\lambda^\star}{2(u_t^\star)^2}=0.
\end{align*}
Multiplying by $-2(u_t^\star)^2$, which is well defined since $u_t^\star>0$, finally yields \eqref{eq:OS_stationarity}. 
\end{proof}

\begin{proof}[Proof of Corollary \ref{cor:lambda-positive}]
Budget saturation follows from Theorem \ref{thm:budget-saturation}. Differentiating \eqref{eq:last-step-affine} gives the identity $\partial_{u_{T-1}}J(u) = d\eta_{T-1}^{2}(\Delta b_{T-1})^{2}$. Combining \eqref{eq:u-gradient} with \eqref{eq:OS_stationarity} yields $\lambda^\star = 2(u_t^\star)^2\,\partial_{u_t}J(u^\star)$ for all $t\in\inter{T}_0$. Taking $t=T-1$, we obtain $\lambda^\star = 2d\eta_{T-1}^{2}(\Dcal_{T-1})^{2}(u_{T-1}^\star)^2$. Since $\eta_t,\Dcal_t,u_t^\star>0$, equation \eqref{eq:OS_stationarity} therefore implies
$\mathbb E\left[\left\langle p_{t+1}^\star,\xi_t\right\rangle\right]
=-\frac{\lambda^\star}{\eta_t\Delta b_t(u_t^\star)^{3/2}}$.
Solving this identity for $u_t^\star$ gives \eqref{eq:ustar_corollary}. 
\end{proof}

\begin{proof}[Proof of Proposition \ref{prop:q-gradient}]
The derivative of the noise term in \eqref{eq:q-dynamics} is
\[
\frac{d}{dq_t}\left[-\eta_t\Dcal_t(2q_t)^{-1/2}\xi_t\right] = \eta_t\Dcal_t(2q_t)^{-3/2}\xi_t.
\]

The chain rule and the adjoint recursion give \eqref{eq:path-q-gradient}. Since $Q$ is compact and $q_t\ge q_{\min}$ on $Q$, there is an open neighborhood of $Q$ on which every coordinate is at least $q_{\min}/2$. The same domination argument as in Proposition~\ref{prop:u-adjoint} therefore justifies interchanging expectation and differentiation.
\end{proof}

\subsection{Proofs: Well-posedness and stability of the control problem}

\subsubsection{Existence and continuity}

\begin{proposition}\label{prop:uad}
For every $\rho>0$, the set $\uad(\rho)$ is nonempty, closed in $\RR^T$, and convex.  Its extreme points are exactly the budget-saturating schedules:
\[
\operatorname{Ext}(\uad(\rho)) = \Big\{u\in(0,\infty)^T:\rho(u)=\rho\Big\}.
\]
\end{proposition}

\begin{proof}
The constant vector $u_t=T/(2\rho)$ is feasible. Moreover, every feasible vector satisfies $u_t\ge(2\rho)^{-1}$, because $(2u_t)^{-1}\le\rho(u)\le\rho$.  Hence a convergent sequence in $\uad(\rho)$ has a strictly positive limit, and continuity of $u\mapsto\rho(u)$ on $[(2\rho)^{-1},\infty)^T$ proves closedness. Convexity follows from convexity of $s\mapsto s^{-1}$ on $(0,\infty)$.

The function $\rho(\cdot)$ is strictly convex. If $\rho(u)<\rho$, sufficiently small perturbations $u\pm v$ remain feasible, so $u$ is not extreme.  Conversely, if $\rho(u)=\rho$ and $u=\lambda v+(1-\lambda)w$ with $v,w\in\uad(\rho)$ and $\lambda\in(0,1)$, then $\rho=\rho(u)\le\lambda\rho(v)+(1-\lambda)\rho(w)\le\rho$.
Equality throughout and strict convexity imply $v=w=u$.
\end{proof}

\begin{lemma}\label{lem:moments}
Under \ref{ass:gaussian} and \ref{ass:lipschitz}, let $K\subset(0,\infty)^T$ be nonempty and bounded. For every $p\ge1$ there is a finite constant $C_{K,p}$ such that
\[
\sup_{u\in K}\E\norm{\theta_T(u)}^p\le C_{K,p}.
\]
More precisely, there is a deterministic $C_K<\infty$ such that
\[
\sup_{u\in K}\max_{0\le s\le T}\norm{\theta_s(u)} \le C_K\left(1+\sum_{t=0}^{T-1}\norm{\xi_t}\right) \quad\text{almost surely}.
\]
\end{lemma}

\begin{proof}
The linear-growth condition gives $\norm{g_t(\theta)}\le a_t+b_t\norm{\theta}$. If we set $\bar u\coloneqq\sup_{\substack{u\in K,\; t\in\inter{T}_0}}u_t$, then
\[
\norm{\theta_{t+1}(u)} \le(1+\eta_tb_t)\norm{\theta_t(u)} + \eta_ta_t + \eta_t\Dcal_t\sqrt{\bar u}\,\norm{\xi_t}.
\]

Finite iteration gives the displayed pathwise bound.  Gaussian random vectors have moments of every finite order, so the stated moment estimate follows.
\end{proof}

\begin{proposition}\label{prop:J-cont}
Under \ref{ass:gaussian}, \ref{ass:lipschitz}, and \ref{ass:poly}, $J$ is finite and continuous on $(0,\infty)^T$, and hence on every $\uad(\rho)$.
\end{proposition}

\begin{proof}
Let $u^n\to u$ in $(0,\infty)^T$. Continuity of the square-root function gives $\sqrt{u_t^n}\to\sqrt{u_t}$ for every $t\in\inter{T}_0$. Since each $g_t$ is continuous by Assumption \ref{ass:lipschitz}, induction in the state recursion \eqref{eq:controlled-dynamics} gives 
\begin{align*}
	\theta_t(u^n;\xi)\to\theta_t(u;\xi) 
\end{align*}
for all $t=0,\ldots,T$ and for almost every noise realization $\xi$. Then by continuity of $L$, we have 
\begin{align*}
	L(\theta_T(u^n;\xi))\to L(\theta_T(u;\xi)) \quad\text{ almost surely}. 	
\end{align*}

The set $K\coloneqq\{u^n\}_{n\in\NN}\cup\{u\}$ is bounded. Hence, by Lemma \ref{lem:moments}, there exists a random variable with finite moments of every order that uniformly bounds $\|\theta_T(v)\|$ for $v\in K$. Together with the polynomial-growth assumption on $L$, this yields an integrable random variable dominating $|L(\theta_T(u^n))|$ uniformly in $n$. The dominated convergence theorem therefore gives $J(u^n)=\E[L(\theta_T(u^n))] \rightarrow \E[L(\theta_T(u))]=J(u)$.

The same polynomial-growth bound and Lemma \ref{lem:moments} show that $J(u)$ is finite for every $u\in(0,\infty)^T$. Thus $J$ is finite and continuous on $(0,\infty)^T$, and consequently on every $\uad(\rho)$. 
\end{proof}

\begin{lemma}\label{lem:orthogonality}
Let $Z$ and $\xi$ be independent square-integrable $\RR^d$-valued random vectors, and suppose $\E\xi=0$. Then, for every $a\in\RR$,
\[
\E\ip{Z}{\xi}=0, \qquad \E\norm{Z-a\xi}^2=\E\norm{Z}^2+a^2\E\norm{\xi}^2.
\]
\end{lemma}

\begin{proof}
Independence and square integrability imply $\E\ip{Z}{\xi}=\ip{\E Z}{\E\xi}=0$. Expanding the square gives the second identity.
\end{proof}

\begin{proposition}\label{prop:coercivity}
Under Assumptions~\ref{ass:gaussian}, \ref{ass:lipschitz}, \ref{ass:coercive-loss}, and \ref{ass:noncollapse}, there are constants $\kappa>0$ and $C_{\rm coer}\ge0$ such that, for every $\rho>0$, 
\begin{equation}\label{eq:coercive-bound}
    J(u)\ge\kappa\sum_{t=0}^{T-1}u_t-C_{\rm coer}, \qquad u\in\uad(\rho).
\end{equation}
Consequently, $J(u^n)\to\infty$ for every feasible sequence with $\norm{u^n}\to\infty$.
\end{proposition}

\begin{proof}
Set $\delta_t:=\theta_t(u)-\bar\theta_t$. Then $\delta_{t+1}=Z_t-\eta_t\Dcal_t\sqrt{u_t}\,\xi_t$, with $Z_t\coloneqq\mathsf G_t(\theta_t(u))-\mathsf G_t(\bar\theta_t)$. The vector $Z_t$ is measurable with respect to $\sigma(\xi_0,\ldots,\xi_{t-1})$, is square integrable by Lemma~\ref{lem:moments}, and is independent of $\xi_t$. Lemma~\ref{lem:orthogonality} therefore gives
\[
\E\norm{\delta_{t+1}}^2 = \E\norm{Z_t}^2+d\eta_t^2(\Dcal_t)^2u_t,
\]
and Assumption~\ref{ass:noncollapse} yields
\[
\E\norm{\delta_{t+1}}^2 \ge \chi_t^2\E\norm{\delta_t}^2-\omega_t + d\eta_t^2(\Dcal_t)^2u_t.
\]
Iterating this inequality gives
\[
\E\norm{\delta_T}^2 \ge d\sum_{t=0}^{T-1}\eta_t^2(\Dcal_t)^2u_t \prod_{j=t+1}^{T-1}\chi_j^2 -\Omega \ge\kappa_0\sum_{t=0}^{T-1}u_t-\Omega,
\]
where
\[
\kappa_0\coloneqq d\min_{t\in\Tset} \left(\eta_t^2(\Dcal_t)^2\prod_{j=t+1}^{T-1}\chi_j^2\right)>0 \quad\text{ and }\quad \Omega\coloneqq\sum_{t=0}^{T-1}\omega_t \prod_{j=t+1}^{T-1}\chi_j^2.
\]
Assumption~\ref{ass:coercive-loss} gives \eqref{eq:coercive-bound} with $\kappa=c_1\kappa_0$ and $C_{\rm coer}=c_1\Omega+c_0$.
\end{proof}

\begin{proof}[Proof of Theorem \ref{thm:existence}]
The feasible set is nonempty and closed by Proposition~\ref{prop:uad}; $J$ is finite and lower semicontinuous by Proposition~\ref{prop:J-cont}.  A minimizing sequence is bounded by coercivity, hence has a convergent subsequence in finite dimension.  Closedness preserves feasibility, and lower semicontinuity gives optimality of the limit.  The same coercivity argument shows that the minimizer set is bounded; it is closed by lower semicontinuity and therefore compact.  This is the finite-dimensional direct-method argument; see, e.g., \cite{cplRockafellarWets1998}.
\end{proof}

\subsubsection{Dependence on the privacy budget}

\begin{proof}[Proof of Theorem \ref{thm:value-continuity}]
Monotonicity follows from $\uad(\rho_1)\subset\uad(\rho_2)$ when $\rho_1\le\rho_2$. Let $\rho_n\to\rho>0$ and $u_n\in S(\rho_n)$. 

For all sufficiently large $n$, we have $\rho/2\le\rho_n\le2\rho$. Fix $\bar u\in\uad(\rho/2)$. Then $\bar u\in\uad(\rho_n)$ and $J(u_n)=V(\rho_n)\le J(\bar u)$.  The coercive estimate in Proposition~\ref{prop:coercivity}, whose constants are independent of the budget, makes $\{u_n\}$ bounded. Moreover, $u_{n,t}\ge(2\rho_n)^{-1}\ge(4\rho)^{-1}$ for all large $n$, so $\rho$ defined in \eqref{eq:rho_u} is continuous on a common neighborhood containing the tail of the sequence.
	
Choose a subsequence $\{n_j\}$ such that $V(\rho_{n_j})\to\liminf_nV(\rho_n)$. Boundedness gives a further subsequence, not relabeled, with $u_{n_j}\to u$. Continuity of $\rho$ yields $u\in\uad(\rho)$, and continuity of $J$ gives 
\[
	V(\rho)\le J(u) = \lim_{j\to\infty}J(u_{n_j}) = \liminf_{n\to\infty}V(\rho_n).
\]

Conversely, choose $u^\star\in S(\rho)$ and set $\alpha_n:=\max\left\{1,\rho\rho_n^{-1}\right\}$ and $\tilde u_n:=\alpha_nu^\star$.
Since $\rho(\alpha u)=\alpha^{-1}\rho(u)$, we have $\tilde u_n\in\uad(\rho_n)$ and $\tilde u_n\to u^\star$.  Hence
\[
	\limsup_{n\to\infty}V(\rho_n) \le\lim_{n\to\infty}J(\tilde u_n) = J(u^\star)=V(\rho).
\]
Thus $V(\rho_n)\to V(\rho)$. If a subsequence $u_{n_k}\to u$, then $u\in\uad(\rho)$ and
\[
    J(u)=\lim_{k\to\infty}J(u_{n_k})\allowbreak =\lim_{k\to\infty}V(\rho_{n_k})=V(\rho) \quad\Rightarrow\quad u\in S(\rho).
\]
If $S(\rho)=\{u^\star\}$, boundedness and uniqueness of every cluster point imply $u_n\to u^\star$.
\end{proof}

\subsubsection{Design-to-training utility stability}

\begin{proof}[Proof of Theorem \ref{thm:design-transfer}]
Couple the design and target dynamics using the same Gaussian innovations. Their noise terms cancel. Global Lipschitz continuity therefore gives
\begin{align*}
    \|\theta_T^{\mathrm{tar}}-\theta_T^{\mathrm{des}}\| \leq A\left(\|\theta_0^{\mathrm{tar}}-\theta_0^{\mathrm{des}}\| + \sum_{t=0}^{T-1}\eta_t\varepsilon_t^g\right), \qquad\text{ with } A\coloneqq\prod_{t=0}^{T-1}(1+\eta_tL_t^{\mathrm{tar}}).
\end{align*}
By compactness of $\MC K$ and Lemma \ref{lem:moments}, both objectives are continuous and
\begin{align*}
    B_a\coloneqq\sup_{u\in \MC K}\bigl(\mathbb E\|\theta_T^a(u)-\theta^{*,a}\|^2\bigr)^{1/2}<\infty.
\end{align*}

Applying $\bigl|\|x\|^2-\|y\|^2\bigr|\leq\|x-y\|(\|x\|+\|y\|)$ to the terminal tracking errors and using Cauchy-Schwarz proves the uniform objective bound with $C_{\MC K}=A(B_{\mathrm{tar}}+B_{\mathrm{des}})$. Evaluating this bound at minimizers gives the minimum-value estimate. Finally,
\begin{align*}
    J^{\mathrm{tar}}(\widehat u) \leq J^{\mathrm{des}}(\widehat u)+\varepsilon_{\MC K} \leq \min_{\MC K}J^{\mathrm{des}}(u)+\gamma+\varepsilon_{\MC K} \leq  \min_{\MC K}J^{\mathrm{tar}}(u)+\gamma+2\varepsilon_{\MC K}.
\end{align*}
\end{proof}

\section{Proofs of the results in Section \ref{sec:algorithm}}\label{app:proofs_algo}

\subsection{Proofs: PB-AMD: construction and algorithm} 

\begin{proof}[Proof of Proposition \ref{prop:floor-update}]
Strict convexity of $D_\psi(\cdot,q^k)$ gives uniqueness. The KKT conditions of the prox subproblem yield, for coordinates above the floor, $\log(q_t^{k+1}/q_t^k)+\alpha_kh_t^k+\alpha_k\lambda_k=0$, while a floor coordinate is obtained whenever the corresponding unconstrained value is at most $q_{\min}$. Setting $\tau_k=e^{-\alpha_k\lambda_k}$ gives \eqref{eq:md-update}.  The left-hand side of \eqref{eq:tau-normalization} is continuous and strictly increasing once it exceeds $Tq_{\min}$, ranges from $Tq_{\min}$ to $+\infty$, and therefore has a unique solution because $\rho_{\max}>Tq_{\min}$.
\end{proof}

\subsection{Proofs: Convergence guarantees}

\begin{proposition}\label{prop:primitive-B}
Assume Assumptions~\ref{ass:gaussian}, \ref{ass:lipschitz}, \ref{ass:poly}, \ref{ass:smooth}, and \ref{ass:second-order}.  Fix $q_{\min}>0$ and let $Q=Q(q_{\min})$.  Then there is an open neighborhood $\MC Q$ of $Q$ and a nonnegative random variable $\Gamma(\xi)$ with finite moments of every order such that, for all $q,z\in\MC Q$,
\begin{equation}\label{eq:path-gradient-envelope}
    \norm{\nabla f(q;\xi)}_2 \le\Gamma(\xi), \qquad \norm{\nabla f(q;\xi)-\nabla f(z;\xi)}_2 \le\Gamma(\xi)\norm{q-z}_2
\end{equation}
almost surely. Consequently:
\begin{enumerate}[(i)]
    \item $F\in C^1(\MC Q)$ and $\nabla F$ is Lipschitz on $Q$ with 
    $\norm{\nabla F(q)-\nabla F(z)}_2\le L_{\rm E}\norm{q-z}_2$,
    where $L_{\rm E}:=\E\Gamma(\xi)<\infty$;
    \item \ref{ass:B1}-\ref{ass:B2} hold, and one may take $L_F=L_{\rm E}/\mu_\psi$; 
    \item for an average of $M_k$ fresh independent path gradients, \ref{ass:B3} holds with 
    $\nu^2:=\sup_{q\in Q}\E\norm{\nabla f(q;\xi)-\nabla F(q)}_2^2<\infty$.
\end{enumerate}
\end{proposition}

\begin{proof}
Choose a bounded open convex neighborhood $\MC Q$ of $Q$ with $\overline{\MC Q}\Subset(0,\infty)^T$, and put $\beta_t(r):=\eta_t\Dcal_t(2r)^{-1/2}$, $\gamma_t(r):=\eta_t\Dcal_t(2r)^{-3/2}$, and $\Xi(\xi):=1+\sum_{t=0}^{T-1}\norm{\xi_t}$.

The functions $\beta_t$, $\gamma_t$, and their first derivatives are uniformly bounded on every coordinate projection of $\overline{\MC Q}$. The associated controls $u_t=(2q_t)^{-1}$ form a bounded set, so Lemma~\ref{lem:moments} and the forward recursion therefore give a polynomial Gaussian envelope for all states, uniformly over $q\in\MC Q$.

\medskip 
\noindent Let $S_{s,j}(q):=\partial_{q_j}\theta_s(q)$. Differentiating the finite recursion gives
\[
S_{s,j}(q)=0~(s\le j),\quad S_{j+1,j}(q)=\gamma_j(q_j)\xi_j,\quad S_{s+1,j}(q)=A_s(\theta_s(q))S_{s,j}(q)~(s\ge j+1).
\]

Applying the polynomial Jacobian and Hessian bounds in Assumptions~\ref{ass:smooth} and \ref{ass:second-order} recursively to the forward states and sensitivities and the backward adjoints yields uniform polynomial estimates. Hence there are deterministic $C<\infty$ and an integer $m\ge1$ such that, for all $q,z\in\MC Q$,
\[
\max_s\norm{\theta_s(q)} + \max_{s,j}\norm{S_{s,j}(q)} + \max_s\norm{p_s(q)} \le C\Xi(\xi)^m
\]
and
\[
\max_s\norm{\theta_s(q)-\theta_s(z)} + \max_{s,j}\norm{S_{s,j}(q)-S_{s,j}(z)} + \max_s\norm{p_s(q)-p_s(z)} \le C\Xi(\xi)^m\norm{q-z}_2
\]
almost surely. Applying these estimates to $\partial_{q_j}f(q;\xi) = \gamma_j(q_j)\ip{p_{j+1}(q)}{\xi_j}$ proves \eqref{eq:path-gradient-envelope} with $\Gamma(\xi)=C'(1+\Xi(\xi)^{m'})$ for suitable deterministic $C'$ and integer $m'$. Since $\Xi$ is a finite sum of Gaussian norms, $\Gamma$ has moments of every finite order. The envelope justifies differentiation under the expectation and yields
$\norm{\nabla F(q)-\nabla F(z)}_2
\le\E[\Gamma(\xi)]\norm{q-z}_2$.

The Euclidean descent lemma and the lower bound in \eqref{eq:entropy-quadratic-comparison} give \ref{ass:B2} with $L_F=\E[\Gamma]/\mu_\psi$, while continuity on compact $Q$ gives \ref{ass:B1}. Finally, the square-integrable envelope makes $\nu^2$ finite, and independence of the $M_k$ paths gives the variance reduction in \ref{ass:B3}.
\end{proof}

\begin{proof}[Proof of Theorem \ref{thm:stochastic-pbamd}]
For the stochastic estimate, write $x=q^k$, $y=q^{k+1}$, $v=\nabla F(x)$, $e=h^k-v$, and $z=T_\alpha(x)$. Relative smoothness and the stochastic prox inequality give
\[
F(y)\le F(x)-(1/\alpha-L_F)D_\psi(y,x)-\ip{e}{y-x}.
\]

Testing the exact prox inequality with $y$, testing the stochastic prox inequality with $z$, and adding the two inequalities yields
\[
\norm{y-z}_2\le\frac{\alpha}{\mu_\psi}\norm{e}_2.
\]
Conditionally on $\mathscr F_k$, the vectors $x$ and $z-x$ are fixed and $\E[e\mid\mathscr F_k]=0$. Hence
\[
\E[-\ip{e}{y-x}\mid\mathscr F_k] \le \frac{\alpha}{\mu_\psi} \E[\norm{e}_2^2\mid\mathscr F_k].
\]
Summing this descent estimate and using (B3) bounds the average of $D_\psi(q^{k+1},q^k)$. Moreover, 
\[
D_\psi(z,x) \le\frac{2M_\psi}{\mu_\psi}D_\psi(y,x) + \frac{M_\psi\alpha^2}{\mu_\psi^2}\norm{e}_2^2,
\]
by \eqref{eq:entropy-quadratic-comparison} and the preceding prox stability. Division by $\alpha^2$ and averaging give \eqref{eq:stochastic-rate}.

\medskip 
\noindent For exact gradients, prox optimality and relative smoothness reduce to
\[
F(q^{k+1}) \le F(q^k)-(1/\alpha-L_F)D_\psi(q^{k+1},q^k).
\]

Telescoping gives \eqref{eq:deterministic-rate} and $\sum_k R_\alpha(q^k)<\infty$. Compactness of $Q$, continuity of $T_\alpha$, and \eqref{eq:residual-kkt-equivalence} then give the cluster-point claim.
\end{proof}

\subsection{Proofs: Fixed-path/SAA implementation}

\begin{proof}[Proof of Proposition \ref{prop:saa-transfer}]
For $z=T_{\alpha,H}(q)$, the prox variational inequality and \eqref{eq:entropy-quadratic-comparison} give
\[
G_H(q) \le \left(\alpha B_H+M_\psi D_{Q,2}\right) \sqrt{\frac{2R_{\alpha,H}(q)}{\mu_\psi}}, \qquad B_H:=\max_{q\in Q}\norm{\nabla H(q)}_2.
\]

Indeed, for an arbitrary $w\in Q$, one writes $q-w=(q-z)+(z-w)$ and bounds the second term by $\alpha^{-1}\norm{\nabla\psi(z)-\nabla\psi(q)}_2\norm{w-z}_2$. Moreover,
$B_{F_M}\le B_F+\sqrt T\,\varepsilon_M$ and, by H\"older's inequality, $G_F(q)\le G_{F_M}(q)+D_{Q,1}\varepsilon_M$. Combining these estimates at $q=q_M$ proves \eqref{eq:saa-stationarity-transfer}.
\end{proof}

\begin{proof}[Proof of Corollary \ref{cor:saa-consistency}]
The envelope \eqref{eq:path-gradient-envelope} is integrable. On any finite net of compact $Q$, the strong law of large numbers applies simultaneously to the finitely many gradient vectors and to the empirical mean of $\Gamma$. The Lipschitz part of \eqref{eq:path-gradient-envelope} extends convergence from the net to all of $Q$; letting the mesh size tend to zero gives the displayed uniform convergence, as in the standard finite-dimensional SAA argument \cite{cplKleywegtEtAl2002,cplShapiroEtAl2014}. Proposition~\ref{prop:saa-transfer} 	then gives population stationarity. The final assertion follows from the bound $L_{F_M}\le M^{-1}\sum_{m=1}^M\Gamma(\xi^{(m)})/\mu_\psi$, the strong law, and the exact-gradient part of Theorem~\ref{thm:stochastic-pbamd}.
\end{proof}

\subsection{Proofs: Linear-quadratic model}

\begin{proof}[Proof of Proposition \ref{prop:LQ-solution}]
The case $a_t=0$ for all $t$ is immediate. Otherwise, let $P\coloneqq\{t:a_t>0\}$ and $Z\coloneqq\{t:a_t=0\}$. At any minimizer, $q_t=q_{\min}$ for $t\in Z$: transferring budget from such a coordinate above the floor to a coordinate in $P$ would strictly decrease $F$. On the remaining coordinates, the objective is strictly convex, so compactness of $Q$ gives a unique minimizer. The KKT conditions on $P$ are
\begin{align*}
    -\frac{a_t}{(q_t^\star)^2}+\lambda^\star-\mu_t=0, \qquad \mu_t\geq0, \qquad \mu_t(q_t^\star-q_{\min})=0.
\end{align*}

Since $\rho_{\max}>Tq_{\min}$, some coordinate in $P$ lies above the floor, implying $\lambda^\star>0$. These conditions give \eqref{eq:LQ-waterfilling}, including the coordinates in $Z$. Its normalization function is continuous and strictly decreasing wherever its value exceeds $Tq_{\min}$, with limits $+\infty$ and $Tq_{\min}$ as $\lambda\downarrow0$ and $\lambda\uparrow\infty$, respectively. Hence $\lambda^\star$ is unique. When every coordinate exceeds the floor, summing $q_t^\star=\sqrt{a_t/\lambda^\star}$ yields \eqref{eq:LQ-closed-form}. 
\end{proof}

\begin{proof}[Proof of Corollary \ref{thm:LQ-global}]
The Hessians satisfy
\[
\nabla^2F(q)=\diag\left(\frac{2a_t}{q_t^3}\right) \preceq L_{\rm LQ}\diag\left(\frac1{q_t}\right) = L_{\rm LQ}\nabla^2\psi(q).
\]
Thus $F$ is convex and $L_{\rm LQ}$-smooth relative to $\psi$. The three-point inequality gives
\[
F(q^{k+1})-F(q^\star) \le\frac1\alpha \left[D_\psi(q^\star,q^k)-D_\psi(q^\star,q^{k+1})\right].
\]

Telescoping and convexity yield \eqref{eq:LQ-rate} and $F(q^k)\to F(q^\star)$. Compactness of $Q$ and uniqueness from Proposition~\ref{prop:LQ-solution} then imply $q^k\to q^\star$.
\end{proof}

\section{Experimental design}\label{app:experiments}

This appendix specifies the experiments in Section~\ref{sec:numerics}. The studies have two distinct aims:
\begin{itemize}
	\item \emph{Allocation quality.} We compare terminal parameter tracking and prediction under different privacy schedules.  Section~\ref{sec:models} defines the models, data, and evaluation samples, and Section~\ref{sec:utility-solvers} specifies how schedules are computed and selected. These settings apply to Figures~3--5 and Table~4 of the main article, and to the budget and implementation checks in Sections~\ref{sec:penalty} and~\ref{sec:numerical-checks} below. 
	\item \emph{Solver convergence.} We measure the time needed to reach the same first-order tolerance when computing a schedule from public records. Section~\ref{sec:matched} specifies the independent instances, stepsize calibration, and stopping rules for Figure~6; Section~\ref{sec:timing} defines the timed operations.
\end{itemize}

Within each solver comparison, Privacy-Budget Adjoint Mirror Descent (PB-AMD), projected gradient descent (PGD), and penalty optimization use the same fixed-path sample-average tracking term $F_M$ in \eqref{eq:saa-objective}, Gaussian paths, and noise calibration. PB-AMD and PGD enforce the budget directly. Penalty + repair first minimizes a budget-penalized objective and then rescales the charges to meet the exact budget.

\subsection{Models, data, and evaluation samples}\label{sec:models}

We first specify the records and models used in Section~5.1 of the main article.  Public records are used to construct schedules; private training records are excluded from schedule construction and used to train models under the fixed schedules; held-out test records measure prediction after training.  Where preprocessing or pretraining is needed, we compute it from public auxiliary records. The public-design sample contains the records used to optimize and select schedules. The sample sizes and, for Fashion-MNIST, the subset relation between these two public samples are given below.

The optimization variables are the per-step privacy charges $q_t=(2u_t)^{-1}$ from (2.5) of the main article, where $u_t$ is the Gaussian noise-variance parameter. For $T$ training iterations, the exact-budget feasible set is (3.10) of the main article: 
$Q=\{q\in\RR^T:q_t\ge q_{\min},\ \sum_tq_t=\rho_{\max}\}$, with $\rho_{\max}$ the total certified budget and $q_{\min}$ the per-step
charge floor.  We use $q_{\min}=10^{-5}$ in the learning experiments.  The separate linear--quadratic (LQ) check in Section~\ref{sec:numerical-checks} uses $q_{\min}=10^{-3}$. The budgets are $0.25,0.5,1$ for Tabular and the two full-network tasks,
Fashion-constant and Fashion-restart, and $0.5,0.25$ for Nonlinear head.

Table~\ref{tab:model-settings} gives the training parameters.  At step $t$, Gaussian noise is added to the gradient query with standard deviation $(2C_{\rm clip}/b)/\sqrt{2q_t}$ in each coordinate, where $b$ is the batch size and $C_{\rm clip}$ bounds the norm of each clipped data gradient. Individual data gradients $v$ are smoothly clipped as $v/\sqrt{1+\norm{v}^2/C_{\rm clip}^2}$, then averaged; the regularization gradient $\lambda_{\rm reg}\theta$, where $\theta$ is the parameter vector, is added afterwards, including for neural-network biases; $\lambda_{\rm reg}$ controls the quadratic regularization. We use replace-one adjacency and compute the total certified privacy charge by summing the per-step charges, as in (2.6) of the main article. We do not reduce this charge to account for random batch selection.

\begin{table}[htbp]
	\centering
	\small
	\begin{tabular}{lrrrrrr}
		\toprule
		Task & Parameters & $T$ & $b$ & Learning rate $\eta_t$ & $C_{\rm clip}$ & $\lambda_{\rm reg}$\\
		\midrule
		Tabular & 64 & 60 & 256 & $0.18$ & $4$ & $0.002$\\
		Fashion-constant & 6634 & 45 & 192 & $0.15$ & $1.75$ & $10^{-4}$\\
		Fashion-restart & 6634 & 45 & 192 & $0.02/0.075$ & $1.75$ & $10^{-4}$\\
		Nonlinear head & 86 & 120 & 384 & $1$ & $1$ & $10^{-4}$\\
		\bottomrule
	\end{tabular}
	\caption{Training configurations.  The restart uses $0.02$ for $t<25$
		and $0.075$ for $t\ge25$, with $t=0,\ldots,T-1$.}
	\label{tab:model-settings}
\end{table}

\subsubsection{Synthetic tabular population}
Tabular fits a linear logistic classifier to a population whose labels also depend on nonlinear feature interactions.  The public-design, private-training, and test samples contain $4500$, $4500$, and $1500$ records, respectively. Covariates $x\in\RR^{64}$ are centered Gaussian with covariance $\Sigma_{ij}=0.65^{|i-j|}+10^{-8}\mathbf1_{\{i=j\}}$. Using zero-based feature indices, the nonzero entries of $w$ are
\[
\begin{array}{c|rrrrrrrrrrrr}
	i&0&1&2&3&4&5&9&13&21&34&45&55\\\hline
	w_i&1.7&-1.4&1.1&-1&.9&-.8&.7&-.65&.6&-.55&.5&-.45
\end{array}
\]
Before random label flips, the probability of a positive label is
\[
\operatorname{sigmoid}\!\left(\frac{1.35}{2.43}
\left[x^\top w+0.85x_0x_1-0.65x_2x_3+0.45\sin x_4
+0.35\mathbf1_{\{x_5>0\}}\right]\right).
\]
Here $\operatorname{sigmoid}(z)=(1+e^{-z})^{-1}$. Each label is independently flipped with probability $0.05$. The fitted logistic loss is $\log(1+e^{-y x^\top\theta})$, with $y\in\{-1,1\}$, no intercept, and initialization $\theta_0=0$. Features are not standardized.  The public-design sample is generated independently of the $6000$ records split into private-training and test samples.

\subsubsection{Fashion-MNIST and the frozen-feature head}
For each seed, the official training split is randomly permuted. The first $8000$ records form the private training sample; the next
$12000$ form the public auxiliary sample.  The first $8000$ records within that auxiliary sample form the public-design sample used for
schedule optimization, public validation, and the public diagnostic evaluation in Figure~4(c) of the main article. The test sample comprises $3000$ seed-selected records from the official test split.  Pixel intensities are divided by $255$, averaged over nonoverlapping $2\times2$ blocks, and flattened to $196$ features. Each feature is centered by its mean over the $12000$ public  records and divided by its public standard deviation plus $10^{-6}$. The same transformation is applied to public, private, and test data.

The full-network classifier is $196$--$32$--$10$, with Softplus hidden activation $\log(1+e^z)$ and softmax cross-entropy data loss.
The weight matrices have independent centered Gaussian entries with variances $1/196$ and $1/32$, respectively; both bias vectors start at zero. Fashion-constant uses this random initialization. Fashion-restart first performs $300$ noiseless public stochastic gradient descent (SGD) steps on the $12000$ public auxiliary records, using learning rate $0.05$, batch size $192$, clipping $1.75$, and regularization $10^{-4}$.  The resulting weights initialize both public-design simulations and private training.  All $6634$ parameters are updated and perturbed in subsequent private training.

Nonlinear head uses the first layer obtained by the same public pretraining procedure as Fashion-restart, then freezes it. Its $32$-dimensional Softplus outputs are centered using the $8000$ public-design records; principal component analysis is computed by full singular-value decomposition on that sample.  We project each centered feature vector onto the first eight principal directions, then divide each projected coordinate by its standard deviation on the public-design sample plus $10^{-6}$. The sign of each principal direction is fixed by requiring its largest-magnitude loading to be positive.  The $8$--$4$--$10$ Softplus head has $86$ trainable parameters.  Its weight matrices have independent centered Gaussian entries with variances $1/8$ and $1/4$, respectively; both bias vectors start at zero.  The head has no separate pretraining stage. During training on private records, only the head is updated and
perturbed; the encoder pretrained on public auxiliary records and the feature transform computed on public-design records remain fixed.

Each iteration samples batch indices uniformly without replacement, independently of record values.  A new batch is sampled at the next
iteration, so records may reappear in later batches.  The complete batch sequence is fixed before training and shared across methods; identical indices are used for the equally sized public-design and private samples. For schedule optimization on public-design data, the \emph{No-noise reference} is the final model trained on the public-design sample with Gaussian noise removed. For tracking evaluation on private training data, it is recomputed on the private training sample. In each case, noisy training and its reference share the initialization, batches, clipping, regularization, and learning rates.  Only the Gaussian perturbation differs.  The reference trained on private records is an offline evaluation target and does not enter schedule optimization or public validation.

\subsubsection{Gaussian paths, random seeds, and evaluation metrics}\label{sec:evaluation}

We use the tracking objective (3.2) with loss (3.6) of the main article at each stage, but change the data or Gaussian paths according to that stage's purpose.  A \emph{Gaussian path} is one complete sequence of standard Gaussian vectors over the training horizon.  Table~\ref{tab:paths} gives the number of paths used in the following three stages:
\begin{enumerate}
	\item \emph{Schedule optimization.} We simulate noisy training on the public-design records.  The mean squared distance to the public 	No-noise reference is $F_M(q)$ in (4.11).  We keep the Gaussian paths fixed while optimizing $q$.
	\item \emph{Public validation.} We keep the public-design records and their No-noise reference unchanged, but evaluate schedules with independent Gaussian paths. When different solver parameters or saved iterates yield several schedules (Section~\ref{sec:utility-solvers}), we select the one with lowest validation tracking error. Optimization and validation thus
	differ in noise samples, not in records. 
	\item \emph{Evaluation on private training data.} After selection, we fix $q$ and simulate training on the separate private records with fresh Gaussian paths.  We compute tracking error against the No-noise reference trained on those private records.  This evaluates a different training sample and reference from those used in $F_M$.
\end{enumerate}

Within a stage, methods share standard Gaussian vectors scaled by their respective schedules.  Consequently, their tracking errors can be compared on paired training runs.
\begin{table}[htbp]
	\centering\small
	\begin{tabular}{lrrr}
		\toprule
		Task & Optimization & Public validation & Private evaluation\\
		\midrule
		Tabular & 256 & 512 & 512\\
		Fashion-constant/restart & 8 & 48 & 256\\
		Nonlinear head & 128 & 256 & 512\\
		\bottomrule
	\end{tabular}
	\caption{Gaussian path counts for allocation-quality studies.
		The public diagnostic in Figure~4(c) also uses $512$ independent paths.}
	\label{tab:paths}
\end{table}

Ten problem seeds specify the data, initialization, and batches shared across methods.  The two head budgets use independent groups of ten seeds.  The code repository records the seed identifiers and exact random-number generation settings.

For each model trained on private records, we compute its tracking error against the corresponding No-noise reference and its prediction scores on the held-out test sample.  Test accuracy is the fraction of correctly classified test records; unregularized cross-entropy is the mean negative log probability of the true test labels. We then average each metric over Gaussian paths.
The full-network prediction comparison in Figure~5 reuses the PB-AMD schedules computed in Section~\ref{sec:utility-solvers}
and uses $256$ noisy training runs on private records for each fixed schedule.

To quantify PB-AMD's temporal redistribution in Figure~3 of the main article, we compare its mean per-step charge in the last third of
training with that in the first third. Averaging this ratio over problem seeds gives $1.26$ to $1.36$ for Tabular across the three budgets and about $3.77$ for Fashion-restart at each budget.

Statistical comparisons use the ten problem seeds as the independent units.  For each seed, let $J_{\rm PB}$ and $J_{\rm base}$ denote
path-averaged tracking errors on private training data for PB-AMD and the comparator. The relative error reduction is $100(1-J_{\rm PB}/J_{\rm base})$, expressed as a percentage of the comparator's tracking error. We average these ten reductions, rather than taking a ratio of errors averaged across seeds, and compute pointwise $95\%$ confidence intervals with Student's $t$ distribution and nine degrees of freedom. Prediction means and their pointwise intervals likewise use the ten seed-level accuracy or cross-entropy scores, with the same Student critical value. 

Two-sided paired Student tests assess whether the mean tracking reduction, accuracy difference, or cross-entropy difference is zero.
Holm's correction adjusts the $p$-values to control the probability of any false declaration of significance within a group of comparisons at the $5\%$ level. We use adjusted $p<0.05$; the confidence intervals above are unadjusted. We apply the correction separately to two groups:
\begin{itemize}
	\item Eight tracking-error tests: PB-AMD versus learning-rate allocation (LR), Decay, Linearized, and Geometric at two budgets.
	\item Twelve prediction tests: PB-AMD versus Uniform for Fashion-constant and Fashion-restart, three budgets, and two metrics.
\end{itemize}

Figure~5 of the main article displays Fashion-restart only, but the prediction correction includes both tasks. The four allocation rules are defined in Section~5.1 of the main article and specified in Section~\ref{sec:head-solvers} below. All six head comparisons against LR, Decay, and Geometric remain significant, with Holm-adjusted $p<1.3\times10^{-5}$. The two Linearized comparisons have adjusted $p=0.144$ and $0.058$ at budgets $0.5$ and $0.25$, respectively, both above the $0.05$ threshold.

Diagnostic instances A and B in Figure~4 of the main article were selected after evaluation for the largest PB-AMD gains over Linearized at each budget. Their intervals measure Gaussian-path uncertainty within each fixed problem, rather than variation across  problem seeds. Let $P_m,B_m$ be PB-AMD's and the comparator's squared parameter tracking errors on the same Gaussian path $m$.  With $512$ paths, set $r=\bar P/\bar B$; bars denote path averages.  The relative reduction is $100(1-r)$, with delta-method standard error
$100\,\operatorname{sd}(P_m-rB_m)/(\sqrt{512}\,\bar B)$, using sample standard deviation.  To account for the selection of A and B after evaluation, simultaneous intervals use a Bonferroni correction over eighty comparisons (twenty instances, four rules). The critical value is $t_{511,\,1-0.05/(2\cdot80)}$, where $t_{\nu,p}$ denotes the $p$-quantile of Student's distribution with $\nu$ degrees of freedom. For diagnostic A (budget $0.5$) and B (budget $0.25$), the tracking reductions over Linearized on private training data reported in the main article have simultaneous intervals $[31.20,33.60]\%$ and $[8.20,10.20]\%$, respectively.

These intervals concern tracking on private training data. Figure~4(c) of the main article instead evaluates the fixed schedules of A on public-design records: the linearized objective is computed from (4.14), and the nonlinear error is averaged over $512$ independent Gaussian paths on the same public-design records. This diagnostic compares how the linearized objective and nonlinear dynamics rank the two fixed schedules.

\subsection{Solvers for the utility comparisons}\label{sec:utility-solvers}

This section specifies schedule construction for the allocation-quality comparisons in Sections~5.2--5.4 of the main article. Section~\ref{sec:matched} gives the separate convergence protocol for Figure~6. PB-AMD, PGD, and penalty optimization start from Uniform, $q_t^0=\rho_{\max}/T$, and use the public fixed-path tracking term $F_M$ defined in Section~\ref{sec:evaluation}.  PB-AMD and PGD minimize this term on $Q$; the penalty method minimizes it with the additional budget-excess term in Section~\ref{sec:penalty}.

For Tabular and full-network training, Table~\ref{tab:utility-steps} gives the PB-AMD stepsize, the PGD stepsize grid, and the update counts. Each stepsize is applied to the unscaled gradient $\nabla F_M$. The returned schedules are selected as follows:
\begin{itemize}
	\item \emph{PB-AMD} retains the iterate with lowest $F_M$, including $q^0$. After the listed minimum number of updates, it stops when that iterate's absolute Karush--Kuhn--Tucker (KKT) residual is at most $5\times10^{-5}$.  The limit is $2000$ updates; a run that fails this criterion is rejected.  Section~\ref{sec:stationarity} defines the residual.
	\item \emph{PGD} runs each of the four stepsizes for the listed fixed number of updates and retains the lowest-$F_M$ iterate from each run. Public validation on independent Gaussian paths selects the best of these four returned schedules by tracking error on the public-design records. 
	\item \emph{Penalty + repair} runs the six parameter pairs in Table~\ref{tab:penalty-grid} for the same fixed update count.
	Section~\ref{sec:penalty} specifies the raw optimizer and budget repair; public validation selects the repaired schedule with lowest tracking error on the public-design records.
\end{itemize}
Nonlinear head instead uses the Armijo and checkpoint-selection procedure in Section~\ref{sec:head-solvers}.

\begin{table}[htbp]
	\centering\small
	\begin{tabular}{lrrl}
		\toprule
		Task & PB-AMD step & Minimum/fixed updates & PGD stepsize grid\\
		\midrule
		Tabular & $0.005$ & $180$ & $\{3\cdot10^{-5},10^{-4},3\cdot10^{-4},10^{-3}\}$\\
		Fashion-constant & $0.0005$ & $100$ & $\{3\cdot10^{-6},10^{-5},3\cdot10^{-5},10^{-4}\}$\\
		Fashion-restart & $0.005$ & $100$ & $\{3\cdot10^{-6},10^{-5},3\cdot10^{-5},10^{-4}\}$\\
		\bottomrule
	\end{tabular}
	\caption{Utility solver settings. PB-AMD checks the KKT tolerance after
		the listed minimum update count; PGD and penalty use the listed fixed count.}
	\label{tab:utility-steps}
\end{table}

\subsubsection{Budget-preserving updates and stationarity measures}\label{sec:stationarity}

PB-AMD implements the floor-aware entropic step (4.2) in Algorithm~1 of the main article. Bisection in logarithmic coordinates solves the normalization equation (4.3), with at most $160$ iterations and a tolerance $10^{-13}\max\{1,\rho_{\max}\}$ on the difference between total charge and $\rho_{\max}$.

PGD projects $q-\alpha\nabla F_M(q)$ onto the same floored simplex by sorting coordinates and computing the Euclidean projection threshold. Both methods correct summation roundoff on the largest coordinate.

For $h=\nabla F_M(q)$, set $I=\{t:q_t>q_{\min}+10^{-9}\}$ and $\bar h_I=|I|^{-1}\sum_{t\in I}h_t$. The reported absolute KKT residual is
\[
\max\left\{\max_{t\in I}|h_t-\bar h_I|,
\max_{t\notin I}[\bar h_I-h_t]_+\right\},
\]
where $[a]_+=\max\{a,0\}$, and the second maximum is zero if all coordinates belong to $I$.  The first term tests equality of gradient
coordinates above the floor; the second tests the one-sided KKT inequality at the floor.  Table~\ref{tab:direct-numerical-checks} reports this absolute residual for the returned PB-AMD schedules. Budget feasibility is checked separately.

Figure~6 instead compares all three solvers with the variational gap (4.12) of the main article, applied to $F_M$:
\[
G_{F_M}(q)=h^\top q-q_{\min}\sum_t h_t
-(\rho_{\max}-Tq_{\min})\min_t h_t.
\]

We replace a negative roundoff value by zero. For Penalty + repair, we evaluate this gap at the repaired charges defined in Section~\ref{sec:penalty}. Writing $F_0=F_M(q^0)$ for the tracking error on public-design records at the Uniform schedule, we divide
the gap by $F_0$ to obtain the common convergence measure $G_{F_M}(q)/F_0$. The absolute KKT residual controls acceptance of
PB-AMD outputs in the utility comparisons; the normalized variational gap compares all three solvers in Figure~6 at a common accuracy threshold.

\subsubsection{Armijo updates and comparison schedules for the head}\label{sec:head-solvers}

The head comparison in Table~4 and Figure~4 tests PB-AMD against the four allocation rules defined in Section~5.1 of the main article.
PB-AMD takes exactly $20$ mirror updates and selects among iterates $0,10,20$ by public validation, using $256$ new Gaussian paths on the public-design records.  The initial step is $\alpha_0=0.3\rho_{\max}/F_0$. We form each trial $q^{\rm tr}$ using the entropic update (4.2) of the main article and accept it when 
\[
F_M(q^{\rm tr})\le F_M(q)
+0.1\,\ip{\nabla F_M(q)}{q^{\rm tr}-q}
+10^{-12}\max\{1,F_M(q)\}.
\]

Rejected trials halve the step, with at most $30$ trials per update. The next initial trial stepsize is the accepted step multiplied by $1.1$, capped at $2\rho_{\max}/F_0$.  All trials use the same fixed Gaussian paths.

The comparison rules assign positive weights $w_t$ and use $q_t=\max\{q_{\min},cw_t\}$ with $\sum_tq_t=\rho_{\max}$; $c$ is found
by $90$ bisection iterations.  Thus all rules satisfy the same floor and exact budget as PB-AMD, but use different information to choose the weights:
\begin{itemize}
	\item \emph{LR} uses $w_t=\eta_t$, retaining only the learning-rate scale.
	\item \emph{Decay} uses $w_t=\eta_t\prod_{s=t+1}^{T-1}|1-\eta_s\lambda_{\rm reg}|$, accounting for subsequent regularization shrinkage.
	\item \emph{Linearized} uses the Jacobians along the No-noise reference trajectory $(\bar\theta_s)$ on public-design records.  With
	$A_s=I_d-\eta_sDg_s(\bar\theta_s)$, set 
	\[
	a_t=\frac{\eta_t^2(2C_{\rm clip}/b)^2}{2}
	\norm{A_{T-1}\cdots A_{t+1}}_F^2,\qquad w_t=\sqrt{a_t}.
	\]
	Here $I_d$ is the identity on the $d$ trainable parameters, $g_s$ is the regularized clipped-gradient query for the fixed batch
	at step $s$, $Dg_s$ is its Jacobian, and $\|\cdot\|_F$ is the Frobenius norm.  This is the coefficient in (4.14) of the main article with identity terminal tracking weight. The empty matrix product is the identity. We evaluate the full matrix products rather than estimating their Frobenius norms with random directions. 
	\item \emph{Geometric} uses $w_t=\eta_t e^{\gamma t/(T-1)}$. To choose its shape parameter, public validation first searches
	$\gamma=-4,-3.5,\ldots,4$, then takes $18$ golden-section iterations on $[\max\{-4,\gamma_0-0.5\},\min\{4,\gamma_0+0.5\}]$, where $\gamma_0$ is the grid point with lowest public-validation error. The best grid schedule is retained if refinement
	does not lower its validation error.
\end{itemize}

\subsubsection{Penalty optimization and exact-budget repair}\label{sec:penalty}

The penalty method replaces exact normalization by the quadratic budget-excess term in (4.4) of the main article.  Its raw objective is
$F_M(q)+\frac{\beta}{2}[\sum_tq_t-\rho_{\max}]_+^2$ in log-variance coordinates $v_t=-\log(2q_t)=\log u_t$.  With
$h=\nabla F_M(q)$, the gradient with respect to $v_t$ is $-q_t h_t-\beta[\sum_sq_s-\rho_{\max}]_+q_t$. Adam uses moment coefficients $0.9,0.999$, bias correction, and denominator stabilizer $10^{-8}$.  The initial moments are zero. After each update $v_t$ is clipped to $[-\log(4\rho_{\max}),-\log(2\cdot10^{-8})]$, so that raw charges lie in $[10^{-8},2\rho_{\max}]$. We return the iterate with lowest penalized objective among all generated iterates, including the initial and final points.

\begin{table}[htbp]
	\centering\small
	\begin{tabular}{lrrrrrr}
		\toprule
		$\beta$ & 1 & 3 & 10 & 100 & 1000 & 10000\\
		\midrule
		Tabular Adam stepsize & $.04$ & $.04$ & $.035$ & $.025$ & $.015$ & $.008$\\
		Full-network/head Adam stepsize & $.035$ & $.032$ & $.025$ & $.017$ & $.01$ & $.006$\\
		\bottomrule
	\end{tabular}
	\caption{Penalty candidates.  Each column specifies one pair of
		penalty weight and Adam stepsize, not a Cartesian-product grid.}
	\label{tab:penalty-grid}
\end{table}

\begin{figure}[!htbp]
	\centering
	\includegraphics[width=\textwidth,trim=0 5.5pt 0 0,clip]{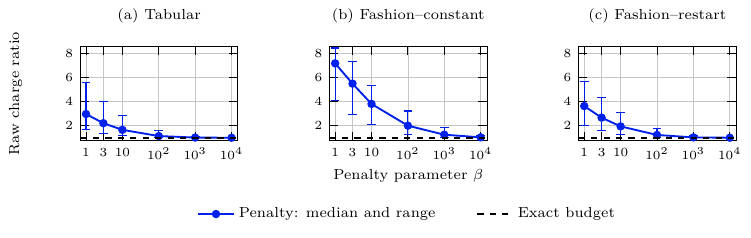}
	\caption{Raw penalty charge divided by the prescribed budget: medians and
		ranges over three budgets and ten seeds.  Ratios above the dashed line at
		one violate the prescribed budget.}
	\label{fig:penalty-charge}
\end{figure}

Figure~\ref{fig:penalty-charge} uses the raw charges to assess budget enforcement.  For fixed-budget tracking comparisons, Penalty + repair instead replaces the raw vector using $z_t=[q_t^{\rm raw}-q_{\min}]_+$ and $q_t=q_{\min}+(\rho_{\max}-Tq_{\min})z_t/\sum_sz_s$.
If $\sum_sz_s=0$, the repaired vector is uniform. Public validation selects the repaired candidate with lowest tracking error on the public-design records.

Across the $90$ Tabular and full-network task--budget instances, all $540$ raw candidates from the six penalty-weight/stepsize pairs violate the budget, with an overall median excess of $63\%$. Figure~\ref{fig:penalty-charge}(a), (b), and (c) show the charge-to-budget
ratios for Tabular, Fashion-constant, and Fashion-restart, respectively. Across the tested pairs, larger penalty weights reduce the excess, but none yields a feasible raw candidate. Repair supplies the exact-budget schedules used for utility and convergence comparisons. PB-AMD instead enforces the budget at each mirror update, so its output needs no feasibility repair.

\subsubsection{Public-data perturbations}\label{sec:public}

We test whether PB-AMD retains its Fashion-restart tracking benefit when the public-design sample changes but private training data do not (Section~5.2 of the main article). We redesign the Fashion-restart schedule at $\rho_{\max}=0.5$ after perturbing the public-design data, then evaluate it on the unchanged private training sample. Public pretraining, feature normalization, private records, and the No-noise reference trained on private records remain fixed.  We recompute the public No-noise reference on the perturbed design data.

Both perturbations are applied to the same public-design sample. Exactly $800$ labels are replaced by uniformly sampled different classes. Standardized design features are mapped to $0.9x+0.15\zeta$, where $\zeta$ has independent standard Gaussian entries.

PB-AMD uses $100$ fixed-step updates of size $0.005$ and returns the iterate with lowest $F_M$. No additional updates are performed to meet a KKT tolerance. The Gaussian path counts for optimization, public validation, and evaluation on private training data are $8,48,256$. The validation paths evaluate the returned schedule; they do not select an additional checkpoint.  The study uses the same ten full-network problem seeds as the allocation-quality comparisons.  We compare the redesigned schedule with Uniform using the same Gaussian paths for training on private records, then average the ten seed-level relative tracking reductions and compute the
pointwise Student interval defined in Section~\ref{sec:evaluation}.

The redesigned schedules retain a mean tracking-error reduction of $27.45\%$ over Uniform at $\rho_{\max}=0.5$, with a $95\%$ confidence interval of $[27.44,27.47]\%$. Thus a schedule redesigned from the perturbed public data still improves private-training tracking, without using private records for schedule construction.

\subsubsection{Stationarity, budget error, and the LQ optimizer}\label{sec:numerical-checks}

Table~\ref{tab:direct-numerical-checks} checks stationarity and exact-budget preservation for Tabular, Fashion-constant, and Fashion-restart, and checks objective accuracy against the analytical LQ optimum. The KKT entry is the largest residual among the $90$ returned PB-AMD schedules: three tasks, three budgets, and ten seeds per combination. Budget error is the largest
$|\sum_t q_t^{k+1}-\rho_{\max}|$ after a mirror update and summation-roundoff correction across these runs.

\begin{table}[!htbp]
	\centering
	\small
	\begin{tabular*}{\textwidth}{@{\extracolsep{\fill}}lcl@{}}
		\toprule
		Check & Value & Scope \\
		\midrule
		Maximum fixed-path KKT residual & $5.00\times10^{-5}$ & $90$ returned schedules \\
		Maximum budget error after mirror updates & $2.22\times10^{-16}$ & $90$ PB-AMD runs \\
		Relative objective gap to water-filling & $7.67\times10^{-16}$ & One LQ instance \\
		\bottomrule
	\end{tabular*}
	\vspace{6pt}
	
	\caption{Stationarity of the returned schedules for $F_M$, budget preservation, and agreement with the
		analytical LQ objective.}
	\label{tab:direct-numerical-checks}
	
\end{table}

In the Tabular and full-network runs, the budget errors in Table~\ref{tab:direct-numerical-checks} are at floating-point roundoff
level, and every returned PB-AMD schedule meets the stopping tolerance in Section~\ref{sec:stationarity}.

For the same runs, the seed-averaged private-training tracking errors of Penalty + repair differ from PB-AMD's by at most $0.102\%$ across the nine task--budget combinations; PGD's errors differ by less than $0.02\%$. Relative to Penalty + repair, PB-AMD attains comparable tracking quality without selecting a penalty weight. Section~\ref{sec:matched} assesses solver time independently at a common first-order tolerance.

The separate scalar LQ check minimizes $F(q)=\sum_t a_t/q_t$ with $T=80$, $\rho_{\max}=1$, and $q_{\min}=10^{-3}$.  For $t=0,\ldots,79$, the scalar transition and noise-injection coefficients are
\[
A_t=0.91+0.055\sin\!\left(\frac{2\pi(t+1)}{80}\right),\quad
B_t=0.035+0.02(t/79)^2,\quad
a_t=\frac12\left(B_t\prod_{s=t+1}^{79}A_s\right)^2.
\]

The empty product is one.  The analytical optimizer is the floored water-filling solution (4.15) of the main article, $q_t^\star=\max\{q_{\min},\sqrt{a_t/\lambda}\}$, with $\lambda$ chosen to satisfy the budget equality.  Initialization puts every
coordinate at the floor and the remaining charge on one coordinate. PB-AMD takes $200000$ updates with $\alpha=3\times10^{-4}$, below the relative-smoothness limit $3.305785\times10^{-4}$. Writing $q^{\rm final}$ for the last PB-AMD iterate, the reported relative objective gap is $(F(q^{\rm final})-F(q^\star))/F(q^\star)=7.671386\times10^{-16}$. The water-filling optimizer has $47$ coordinates at the floor. The agreement at numerical precision provides an implementation check against the analytical optimizer in Proposition~4.5.

\subsection{Matched time-to-stationarity protocol}\label{sec:matched}

Figure~6 of the main article compares the time taken by PB-AMD, PGD, and Penalty + repair to compute a schedule at the same relative
variational-gap tolerance for Tabular, Fashion-restart, and Nonlinear head. All three use the same tracking term $F_M$ on public-design
records and the same fixed Gaussian paths; the gap is evaluated on feasible schedules, including the repaired output of penalty optimization. One \emph{public-design instance} specifies these records, the model initialization, the minibatch sequence, and the fixed Gaussian paths for one problem seed.  This study uses different problem seeds from the allocation-quality comparisons in Section~\ref{sec:utility-solvers}. All runs use $\rho_{\max}=0.5$, Uniform initialization, and the models of Section~\ref{sec:models}.  The archived PB-AMD/PGD benchmark also includes Fashion-constant.  In the order Tabular, Fashion-constant,
Fashion-restart, Nonlinear head, the fixed Gaussian path counts are $256,8,8,128$.

The three \emph{calibration instances} per task are used to choose solver parameters.  Those parameters are then fixed on the ten
independent \emph{convergence instances} used to measure solver time and gap. Both sets concern schedule optimization on public records, not the held-out records used to measure classification accuracy. For Tabular and both full-network tasks, we run PB-AMD and PGD for $30$ updates on each of the three public calibration instances using each stepsize factor $\gamma\in\{0.03,0.1,0.3,1\}$.
With $F_0=F_M(q^0)$, the actual fixed stepsizes are $\alpha_{\rm PB}=\gamma\rho_{\max}/F_0$ and $\alpha_{\rm PGD}=\gamma\rho_{\max}^2/(TF_0)$. At Uniform, these scalings give the same first-order displacement for a small $\gamma$. We select each solver's $\gamma$ by minimizing the mean logarithm of its relative variational gap $G_{F_M}(q)/F_0$ across calibration instances.
Here the gap is measured after the $30$ calibration updates. Values below $10^{-16}$ are replaced by $10^{-16}$ before taking logarithms; ties favor smaller $\gamma$. The selected PB-AMD/PGD values are $(0.3,0.3)$ for Tabular, $(1,0.3)$ for Fashion-restart, and $(0.3,0.3)$ for Fashion-constant.

For Nonlinear head, both PB-AMD and PGD use Armijo backtracking rather than a calibrated fixed step.  Each is run on the same ten convergence instances. The dimensionless step $\gamma$ starts at $0.3$ and is capped at $2$; we convert it to $\alpha_{\rm PB}$ and $\alpha_{\rm PGD}$ using the same formulas as for the fixed-step runs above. The sufficient-decrease coefficient is $0.1$; rejected trials multiply the step by $0.5$, and accepted steps grow by $1.1$, with at most $30$ trials per update.  Both methods use the sufficient-decrease inequality and rounding allowance in Section~\ref{sec:head-solvers}, with their respective trial updates.

For each of the three tasks in Figure~6, penalty calibration uses the six pairs in Table~\ref{tab:penalty-grid}, the same three public calibration instances as PB-AMD and PGD, and $30$ raw updates per candidate. As for the other solvers, selection minimizes the mean logarithm of the relative gap, evaluated here at the repaired point, with the same $10^{-16}$ floor; ties favor smaller $\beta$.
The selected penalty-weight/Adam-stepsize pairs are $(1000,0.015)$ for Tabular, $(100,0.017)$ for Fashion-restart, and $(1000,0.01)$ for Nonlinear head. During each convergence run, we repair the iterate with lowest penalized objective found so far at each recorded iteration and compute the repaired point's variational gap.  If the repaired point is unchanged, its previous diagnostic is reused.

The common accuracy criterion is $G_{F_M}(q)/F_0$, defined in Section~\ref{sec:stationarity}.  Two thresholds serve different purposes:
\begin{itemize}
	\item \emph{Time comparison.} We record the first time at which the relative gap is at most $10^{-4}$, using the recorded diagnostics without interpolation.  This is the common target used for the time savings reported in Section~5.5 and Figure~6 of the main article.
	\item \emph{Curve termination.} A run continues until the relative gap is at most $10^{-6}$ or the update limit is reached.  The limits are $180,100,100,60$ for Tabular, Fashion-constant, Fashion-restart, and Nonlinear head, respectively.
\end{itemize}

For each instance where both solvers reach $10^{-4}$, let $t_{\rm PB}$ and $t_{\rm PGD}$ be their first recorded times to this
threshold. The reported time saving is $100[1-1/\operatorname{median}(t_{\rm PGD}/t_{\rm PB})]\%$, where the median is taken over these paired instances. In contrast, the curves use all ten convergence instances per task, showing the median and interquartile range of the latest relative gap available at each time.  Stopped runs retain their final gap at later plotted times, including runs that miss the target.

\subsection{Timing and implementation}\label{sec:timing}

The timings in Figure~6 use Python $3.12.14$, NumPy $2.3.5$, and double-precision arrays on a macOS $26.3$ ARM64 CPU platform.
NumPy implements the forward training recurrence and backward vector--Jacobian products.

We run solvers serially, alternating the PB-AMD/PGD order across instances. Neural objective and adjoint evaluations use four fixed worker threads, with one numerical-library thread per worker; Tabular evaluations are vectorized in one thread. All solvers share the backend within each task, and one adjoint evaluation covers all $M$ paths.

We time updates, required forward and adjoint evaluations, all line-search trials, repair, and final diagnostics. We exclude calibration, data/model construction, warm-up, file output, training on private records, prediction scoring, and diagnostics used only to draw the curves.

Penalty timings use separate runs on the same host and backend. We subtract intermediate adjoint time used only to plot gaps at repaired points, but retain one adjoint evaluation of the returned point unless the solver already computes it. Repair remains included. Raw timings and subtracted costs are archived. The code README provides experiment entry points, run commands, and exact replay settings. 

\noindent

\bibliographystyle{acm}
\bibliography{biblio}

\end{document}